\documentclass[10pt,twocolumn,letterpaper]{article}

 \usepackage{cvpr}              
\usepackage{multirow}
\usepackage[dvipsnames]{xcolor}

\definecolor{cvprblue}{rgb}{0.21,0.49,0.74}
\usepackage[pagebackref,breaklinks,colorlinks,citecolor=cvprblue]{hyperref}

\def\paperID{2939} 
\def\confName{CVPR}
\def\confYear{2024}

\title{DreamInpainter: Text-Guided Subject-Driven Image Inpainting with \\ Diffusion Models}

\author{Shaoan Xie*$^{1}$, Yang Zhao$^{2}$, Zhisheng Xiao$^{2}$, Kelvin C.K. Chan$^{2}$, 
Yandong Li$^{2}$, \\ Yanwu Xu$^{2,3}$, Kun Zhang$^{1,4}$,Tingbo Hou$^{2}$\\
\tt\small $^1$ Carnegie Mellon University \\
\tt\small $^2$ Google \\
\tt\small $^3$ Boston University \\
\tt\small $^4$ Mohamed bin Zayed University of Artificial Intelligence \\
\tt\small \{shaoan,kunz1\}@cmu.edu,\{yzhaoeric,zsxiao, kelvinckchan, yandongli, yanwuxu, tingbo\}@google.com \\
}

\begin{document}
\maketitle

\let\thefootnote\relax\footnotetext{* Work done as a student researcher of Google.}

\begin{abstract}
%

This study introduces \textbf{Text-Guided Subject-Driven Image Inpainting}, a novel task that combines text and exemplar images for image inpainting. While both text and exemplar images have been used independently in previous efforts, their combined utilization remains unexplored. Simultaneously accommodating both conditions poses a significant challenge due to the inherent balance required between editability and subject fidelity.
To tackle this challenge, we propose a two-step approach \textbf{DreamInpainter}. First, we compute dense subject features to ensure accurate subject replication. Then, we employ a discriminative token selection module to eliminate redundant subject details, preserving the subject's identity while allowing changes according to other conditions such as mask shape and text prompts. Additionally, we introduce a decoupling regularization technique to enhance text control in the presence of exemplar images.
Our extensive experiments demonstrate the superior performance of our method in terms of visual quality, identity preservation, and text control, showcasing its effectiveness in the context of text-guided subject-driven image inpainting.


\end{abstract}    
\section{Introduction}
\label{sec:intro}

Text-to-image generation models~\citep{rombach2022high,balaji2022ediffi,ramesh2022hierarchical,nichol2021glide,podell2023sdxl} revolutionize the field of visual content creation by serving as the cornerstone for a wide range of tasks ~\citep{ruiz2023dreambooth,zhang2023adding,hertz2022prompt,yang2023paint,tumanyan2023plug,gal2022image,blattmann2023align,poole2022dreamfusion}. Among them, image inpainting holds significant importance in various applications, such as image restoration, object removal, and augmented reality, where preserving visual coherence and maintaining the integrity of the overall scene are critical. 

\begin{figure}
    \centering
    \setlength{\tabcolsep}{2pt}
   \includegraphics[scale=0.5]{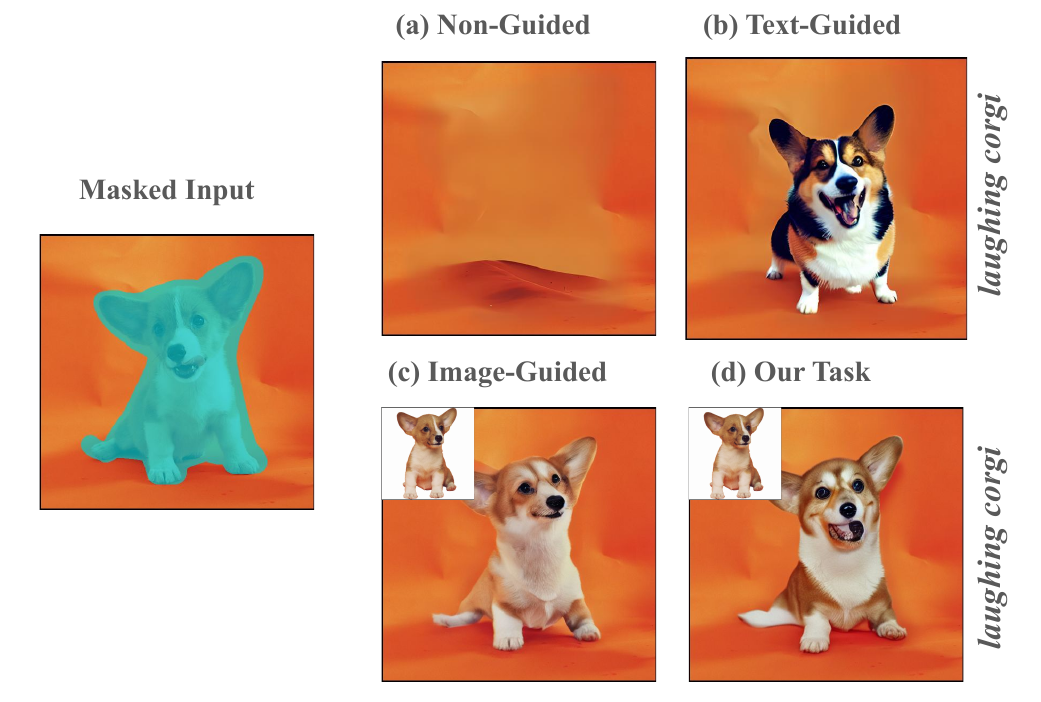}
    \caption{Different from previous works, which accepts only at most one condition for inpainting, we consider the task of \textbf{\textit{Text-Guided Subject-Driven Image Inpainting}}, the generalization of previous tasks with an objective to inpaint an image using both exemplar images and text description.}
    \label{fig:task_illus}
\end{figure}

While traditional image inpainting methods~\citep{lugmayr2022repaint} proficiently fill in missing areas based on visible pixels, they often lack fine-grained control over the inpainted content. In contrast, text-to-image diffusion models excel in generating content that adheres closely to user-provided guidance, making them a preferred choice for conditional inpainting tasks. Specifically, they serve as ideal candidates for \textit{text-conditioned inpainting}, where textual descriptions guide the generation of content within masked regions. Research has demonstrated the effectiveness of adapting text-to-image diffusion models into inpainting models through fine-tuning. This involves training the model to recover missing areas of an image using randomly generated masks, conditioned on corresponding image captions~\citep{nichol2021glide,rombach2022high}. Noteworthy advancements in this domain involve the incorporation of multimodal embeddings~\citep{avrahami2022blended} and considerations of mask shape~\citep{xie2023smartbrush}, which further enhance the precision of content control in the inpainting process.

While text-conditioned image inpainting suffices for basic applications like object removal, achieving stronger controllability over the inpainted content is desirable. An valuable form of conditioning is the use of reference images. For instance, imagine a scenario where a picturesque landscape photo prompts the desire to insert one's pet cat seamlessly into the scene. The task of generating images featuring specific objects in diverse contexts recently emerged as \textit{personalized generation}. Researchers have explored the fine-tuning of text-to-image diffusion models with regularization on reference images~\citep{ruiz2023dreambooth,gal2022image,kumari2023multi}. This approach allows the model to adapt to the appearance of a specific object while retaining the capability to generate diverse scenes. However, the fine-tuning process is often resource-intensive and time-consuming, posing limitations for real-time interactive applications. An alternative strategy for personalized generation involves extracting features from reference images using an auxiliary encoder ~\citep{wei2023elite,chen2023subject}. This encoder-conditioned text-to-image model can then be fine-tuned, eliminating the need for training on each unique instance. 

In this paper, we explore a new task at the crossroads of text-conditioned inpainting and personalized generation. Our objective is to inpaint a designated area within an image, guided not only by a textual description but also by a specific object represented in a reference image. The textual guidance serves as a creative prompt, directing the generation of the specified object within the image. To illustrate, envision the insertion of ``one's cat riding a skateboard" into a picturesque landscape. We name this distinctive task as \textit{\textbf{Text-Guided Subject-Driven Image Inpainting}}. See Fig. \ref{fig:task_illus} for an illustration of inpainting tasks with different types of guidance. The concurrent conditioning on both textual descriptions and reference subjects empowers users to benefit from the imaginative potential of text-based guidance while maintaining the precision offered by exemplar-based methodologies. Through the synergy of these dual conditions, our proposed task opens avenues for boundless creativity in the realm of image inpainting.

To the best of our knowledge, the task we address in this paper remains unexplored in existing literature. While some prior work has delved into exemplar-based inpainting, such as the notable contribution by ~\citep{yang2023paint,chen2023anydoor}, these approaches lack the nuanced capability of enabling textual control over the inpainted object. Despite the apparent simplicity of combining two well-studied tasks, our proposed task is highly non-trivial. On one hand, fine-tuning a personalized diffusion model for inpainting is challenging due to the limited availability of reference training images. On the other hand, a simplistic utilization of features from reference images to fine-tune a text-conditioned inpainting model induces the risk of copy-paste effect, where visual features overpower the generative process. 
As a result, we need sophisticated design considerations and the development of novel techniques for this unique task.

To tackle the task, we devise a bottom-up approach that first computes dense subject features ensure accurate subject replication, and applies filtering to remove excessive subject details, encouraging editability. Specifically, we compute dense features using the UNet encoder of the pre-trained diffusion model. We then rank each token and retain only tokens with high scores. Our token selection module preserves essential tokens for identity preservation while avoiding direct subject copying. Furthermore, to enhance text control, we propose a decoupling regularization during training. Instead of recovering only the masked region, we compute losses on the entire image, forcing network to leverage text captions for restoring details beyond the reference subject, preventing direct copying and significantly improving text-conditioning in the presence of reference subjects. Our experiments demonstrate that our method preserves identity more effectively and allows for more flexible text controls compared to existing state-of-the-art approaches. Our contributions are summarized as follows: 

\begin{itemize}
    \item We introduce the text-guided subject-driven image inpainting task, a novel task that require balance between identity preservation and editability by text prompt.
    \item We propose a discriminative token selection module and decoupling regularization to mitigate subject copying and enhance text-conditioning.
    \item  We conducted extensive experiments to demonstrate the effectiveness of our method. 
\end{itemize}
\section{Related Work}
\label{sec: related work}

\vspace{2mm} \noindent \textbf{Diffusion Models.}
Recently, diffusion probabilistic models is gaining wide attention due to its superior performance over GAN. Unlike GAN which requires adversarial training, DPMs adds noises on the images and train a UNet to recover the clean images and thus is known to be more stable. DPMs have been shown effective in different tasks, such as text-to-image generation \citep{ramesh2022hierarchical,saharia2022photorealistic,rombach2022high,ramesh2021zero}, video generation \citep{ho2022imagen,harvey2022flexible,mei2023vidm,esser2023structure,blattmann2023align,molad2023dreamix,wu2023tune,luo2023videofusion}, image editing \citep{kawar2023imagic,couairon2022diffedit,mokady2023null,nichol2021glide,hertz2022prompt}.

\vspace{2mm} \noindent \textbf{Diffusion Model for Image Inpainting.}
Image Inpainting aims to fill the missing region unconditionally or with given conditions, such as text and reference image. partial objects or inconsistent content in the background. GLIDE \citep{nichol2021glide} finetunes the text-to-image model with randomly generated masks and fill the missing region with text prompt during inference. Blended Diffusion \citep{avrahami2022blended} encourages the generated content to be aligned with text prompt by maximizing the CLIP similarity. SmartBrush \citep{xie2023smartbrush} proposes to utilize the existing segmentation dataset to finetune the text-to-image model rather than generating masks randomly. Inst-Inpaint \citep{yildirim2023inst} first generate a dataset which contains an original image and an image which removes the object. Then the model is able to remove the object by input text prompt without a user-specified object mask. DiffEDIT \citep{couairon2022diffedit} proposes to utilize the cross attention map to find the object and then replace the object with another text prompt. Paint-by-Example \citep{yang2023paint} proposes to use the CLIP image embedding to replace the CLIP text embedding and generates content which is similar to the object in the reference image.  However, the CLIP image embedding may fails to capture the detailed information. To address this problem, AnyDoor \citep{chen2023anydoor} proposes to use the high-frequence map of the reference object as additional information.

\vspace{2mm} \noindent \textbf{Token Pruning and Positive Sample Mining.}
Token selection has been an popular topic in improving the efficiency and performance of the transformer model. 
\citep{wang2022efficient, liu2022ts2} proposes to train a scorer network to select tokens with highest score to reduce the computation cost of video transformer. Many methods have been proposed to utilize the attention map in transformer layer to prune the unnecessary tokens \citep{goyal2020power, zhou2022token,fayyaz2022adaptive,kim2022learned}. By contrast, we are not provided with the attention map for the features. 
 Positive sample selection has also been an important topic in contrastive learning. \citep{hu2022qs} computes the entropy of different patches and select the most informative patches to serve as positive samples in image-to-image translation. \citep{jeon2021mining} proposes to use optimal transport to select tokens to improve temporal coherence.



\vspace{2mm} \noindent \textbf{Subject-Driven Image Generation and Editing.}
Subject-driven image generation \citep{ruiz2023dreambooth} aims to learn the concept from few images containing same subjects. Dreambooth \citep{ruiz2023dreambooth} finetunes the text-to-image model by denoising the given set of images with a special token and propose a prior regularization loss to avoid model forgetting. SuTI \citep{chen2023subject} proposes to generate millions of expert DreamBooth models and finetune the model with the dataset. Then it avoids the optimization for each new concept. DisenBooth \citep{chen2023disenbooth} proposes to disentangle the latents and learn a identity-embedding and an identity-irrelevant embedding and minimizes their similarity. ELITE \citep{wei2023elite} proposes to project multi-layer CLIP features of a given image in the textual word embedding space and inject the encoded patch features into cross attention layer. Unlike the above subject-driven image generation methods, our task is more challenging since we are only given one image during training and we need to generate the object that aligns with the input mask. Object composition is also strongly related to the task as it requires the object to be copied to another image in a realistic way.  \citep{song2023objectstitch} proposes to learn a meaningful image embedding which are in the same space as text embedding.

\section{Text-Guided Subject-Driven Inpainting}
\label{sec:method}
We first establish the objective of our proposed task: text-guided subject-driven image inpainting. Subsequently, we present our proposed approach designed to tackle the inherent challenges of this task.

\begin{figure*}[!t]
    \centering
    \includegraphics[scale=0.6]{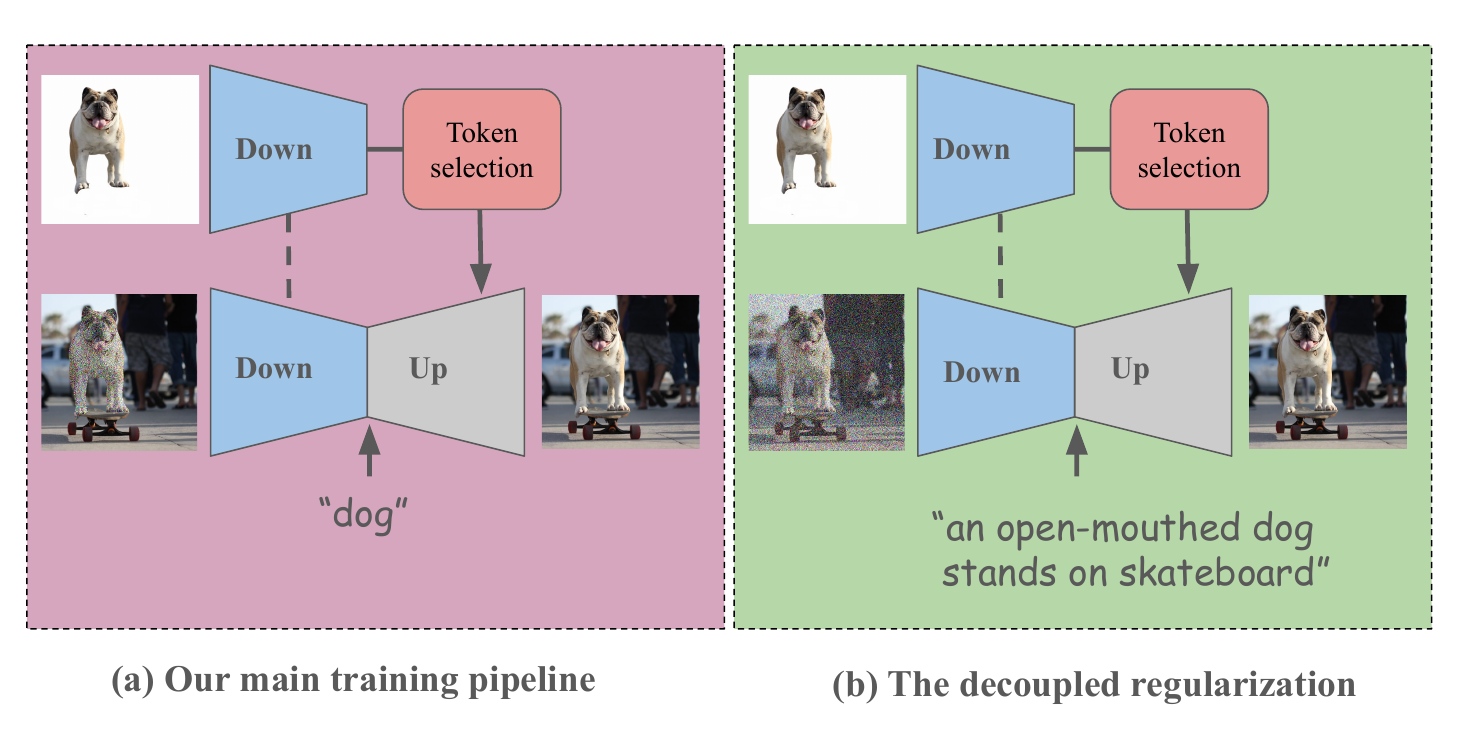}
    \caption{The training diagram of our method. On the left, we show the main training pipeline for inpainting and noise is added to the object only. On the right, we present the decoupling regularization and we add noise on the whole image. We first extract feature of reference object with the downstack of the Unet in diffusion model and perform token selection to avoid copy-paste.
    }
    \label{fig:diagraml}
\end{figure*}

\vspace{2mm} \noindent \textbf{Task Definition.} Given an image $x$ with a specified mask $m$, a text description $c$ of $x$, exemplar images $x_r$ and its corresponding class label $c_{x_r}$,
the primary objective is to generate an image $\tilde{x}$ that satisfies three crucial criteria:
\begin{itemize}
    \item \textbf{Preservation of subject identity.} The inpainted subject in the generated image $\tilde{x}$ is visually similar to $x_r$.
    \item \textbf{Alignment with texts.} The appearance of the inpainted subject aligns with the description specified in the text prompt $c$.
    \item \textbf{Background Unaltered.} $x$ and $\tilde{x}$ are identical except for the masked region.
\end{itemize}

\vspace{2mm} 
\noindent\textbf{Challenges.} 
We face two primary challenges in this context:
Firstly, acquiring an appropriate representation for $x_r$ to effectively capture the subject identity is crucial in the proposed tasks. However, our initial experiments reveal that a straightforward learning approach resulted in copy-paste artifacts, as illustrated in Fig.~\ref{fig:copy_demo}. Additionally, relying on pre-trained model features, such as the CLIP class token, lead to a loss of identity, as depicted in Fig.~\ref{fig:comp}. Secondly, exemplar images typically hold more influence than text descriptions. Consequently, when both conditions coexist, the network tends to prioritize the exemplar image over the text prompt. This prioritization introduces inconsistencies between the generated output and the intended text prompt, a discrepancy illustrated in Fig.~\ref{fig:decouplel_ablation}.

\vspace{2mm} 
\noindent \textbf{Our Approach.} 
In this work, we propose a diffusion-based approach to address the aforementioned challenges. Given the exemplar image $x_r$, we first compute its identity representation. We then apply our proposed token selection scheme to filter out unnecessary information to enhance editability. Furthermore, we introduce a decoupling regularization during training to avoid direct subject copying. The model is trained with the conventional loss used in diffusion models:
\begin{align}
    \mathcal{L} = \mathbb{E}_{x, c, c_{x_r}, t, \epsilon \sim \mathcal{N}(0, \mathbf{I})}\|\epsilon - \epsilon_{\theta}(x_t, t, c, c_{x_r})\|_2^2,
\end{align}
where $\epsilon_{\theta}$ is the learnable function parameterized by $\theta$ and $\epsilon$ is a schedule additive noise used to corrupt the data into a noisy version $x_t$. 




\subsection{Discriminative Subject Feature}
Existing methods typically adopt pretrained models such as CLIP~\citep{radford2021learning} to capture subject identity. However, models designed to capture high-level semantics could be insufficient for capturing low-level details of the subjects. 
%
\begin{figure}
    \centering
    \scalebox{0.8}{
   \begin{tabular}{ccc}
   Input & Reference & Output\\
        \includegraphics[scale=0.12]{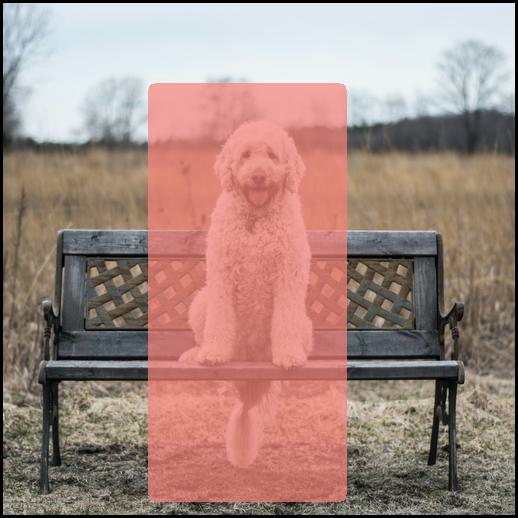}&
        \includegraphics[scale=0.12]{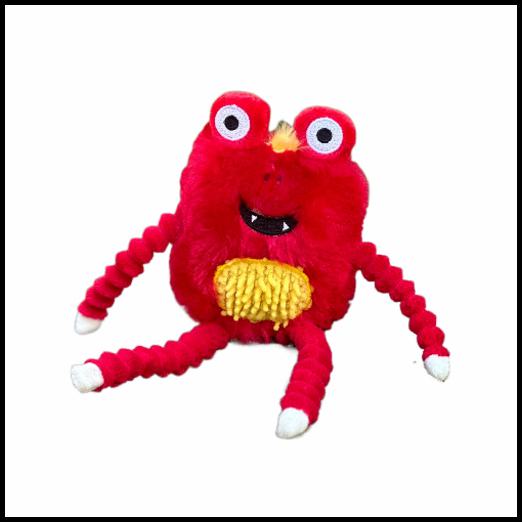} &
        \includegraphics[scale=0.12]{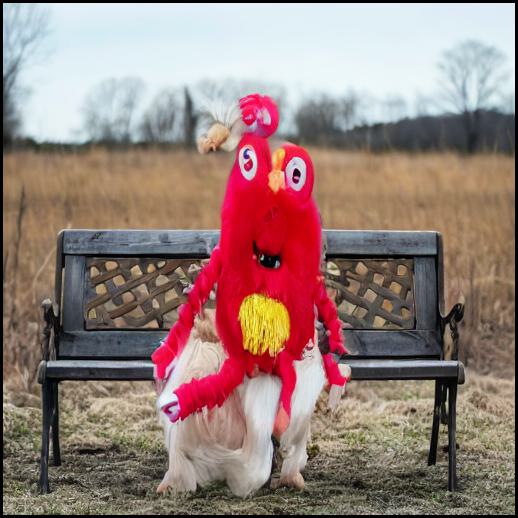}
        \\
          \includegraphics[scale=0.12]{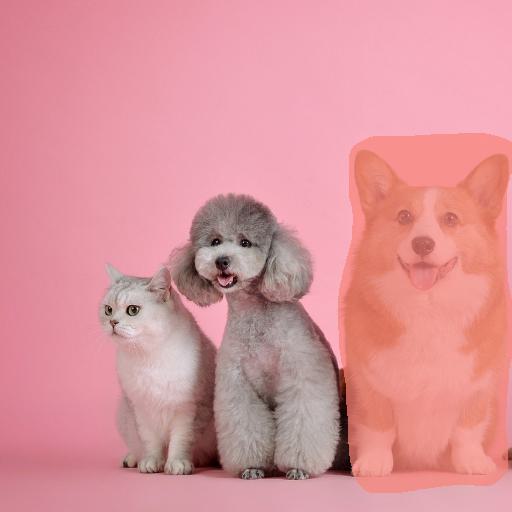}&
        \includegraphics[scale=0.12]{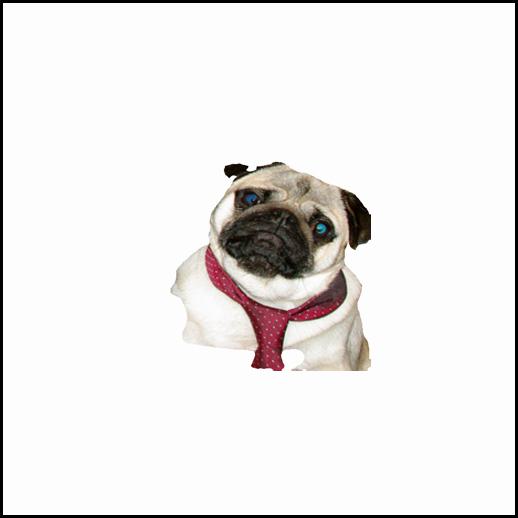} &
        \includegraphics[scale=0.12]{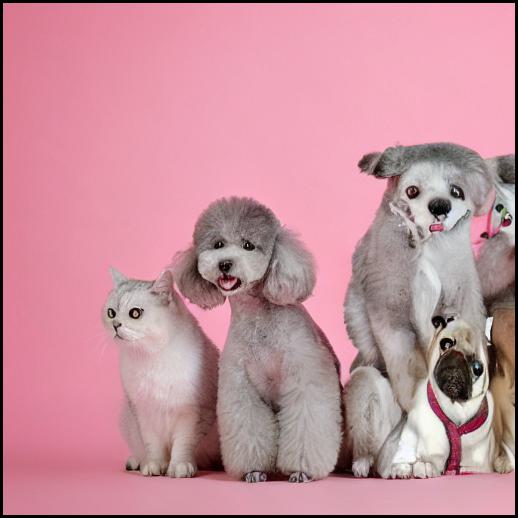}
        \\
   \end{tabular}
   }
    \caption{Copy-paste artifacts when using all tokens from UNet. The model learns the trivial mapping which copies the reference object to the masked region directly.}
    \label{fig:copy_demo}
\end{figure}
Therefore, we opt to leverage the information provided by the UNet of the diffusion model. The pretrained Stable Diffusion~\citep{rombach2022high} possesses the capability to reconstruct images and preserve low-level details in its intermediate layers, making it a valuable source of comprehensive information pertaining to the reference image. In this work, we propose to use the downsample stack of the UNet to obtain the final feature map as the representation of $x_r$:
\begin{align}
    f(x_r) := \text{UNetDownSample}(x_r, t, c).
\end{align}
Here, we feed the clean image $x_r$ into the downstack and set $t=0$ to obtain a clean representation for $x_r$. 

Similar to Stable Diffusion, our approach centers around a latent space of dimensions $64\times64\times8$, where the diffusion model is trained. The UNet incorporates a sequence of four downsampling blocks that progressively reduce the spatial dimensions of the input, from an initial size of $64\times64$ down to $8\times8$. Notably, we find that the output from the second downsampling block, with dimensions $32\times32\times768$, encapsulates rich and task-relevant information from the input feature. However, despite being informative, the excessive subject-relevant information leads to direct copying of the object, as shown in Fig.~\ref{fig:copy_demo}.

To address the issues, we propose to select the most discriminative tokens.
Given a flattened feature $f(x_r) \in \mathbb{R}^{N\times D}$ produced by the UNet downstack, we compute the pairwise distance $\mathbf{d}\in \mathbb{R}^{N\times N}$ between each element: 
\begin{align}
    d_{i, j} = \cos(f(x_r)_i, f(x_r)_j),
\end{align}
where $\cos(\cdot)$ is the cosine similarity. We then compute the score of each token as
\begin{align}
    \mathbf{p}_{i} = \text{softmax}([d_{i,1}, d_{i, 2}, ..., d_{i, N}])
\end{align}
Then we can obtain a score for each token as $s_i = \sum_{j=1}^N \mathbf{p}_{j,i}$, 
where $p_{j,i}$ is the $i$-th component of the vector $p_j$.
The score $s_i$ sums the contributions of other tokens from $j=1...N$ to the current token $i$. In such a way, tokens that are dissimilar to all other tokens would obtain a low score $s_i$. We can then rank the tokens by $s_i$ in ascending order and select the top $K$ tokens. The final representation for the reference image is $f(x_r) \in \mathbb{R}^{K\times D}$. We present examples of using different values for $K$ in Fig.~\ref{fig:num_token_vis}.



\subsection{Decoupling Regularization}
As aforementioned, the network tends to ignore the text description in the presence of exemplar image. For instance, suppose we have a reference image featuring a corgi and a text prompt \textit{``A smiling corgi''}. It is observed that the network tends to directly copy the subject without following the descrption \textit{``smiling''}.


To counteract this, we examine the training data and identify that the problem stems from the pairing of reference text and images. For example, in the case of \textit{``A dog photo''}, the training reference image is a cropped version of the dog, while the training text prompt corresponds to the class label of the segmentation, such as \textit{``dog''}. This setup introduces redundancy as the model often relies heavily on the information from the reference image, essentially ignoring the text prompt.

\begin{figure*}
    \centering
    \setlength{\tabcolsep}{2pt}
    \scalebox{0.85}{
    \begin{tabular}{cc|ccc}
    \toprule
         
         Input & Reference Subject & \multicolumn{3}{c}{Text-Guided Subject-Driven Inpainting} \\ \midrule
         
         && \textit{cat riding bike} & \textit{cat lying down} & \textit{cat on skateboard} \\ 
         \includegraphics[scale=0.17]{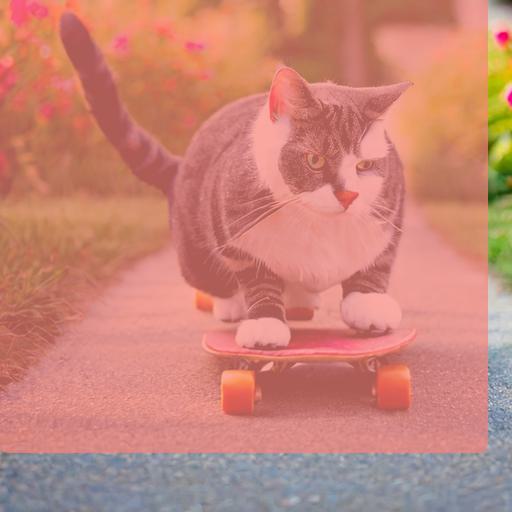}&
          \includegraphics[scale=0.17]{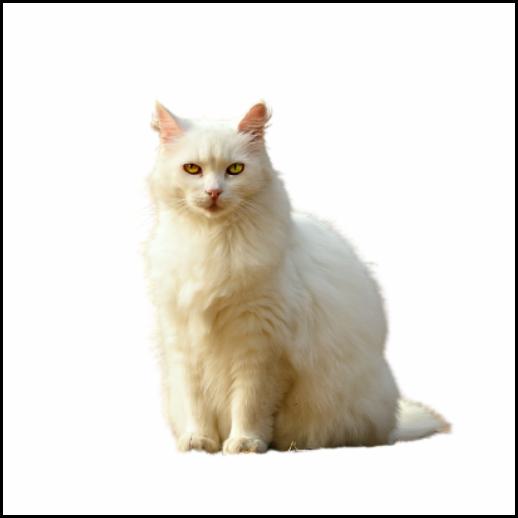}&
           \includegraphics[scale=0.17]{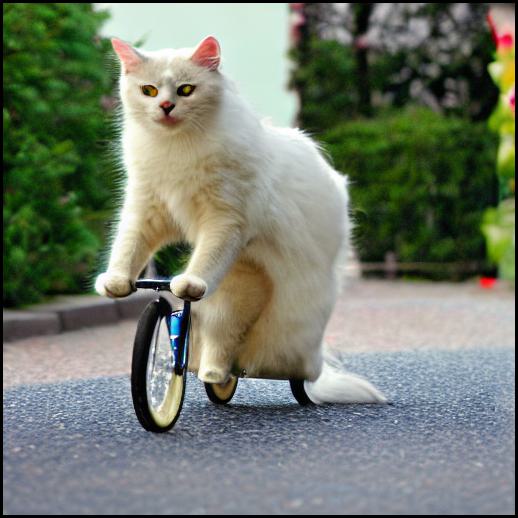}&
         \includegraphics[scale=0.17]{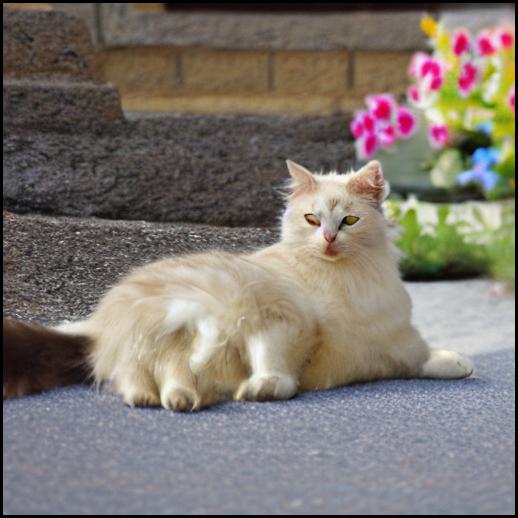} &
         \includegraphics[scale=0.17]{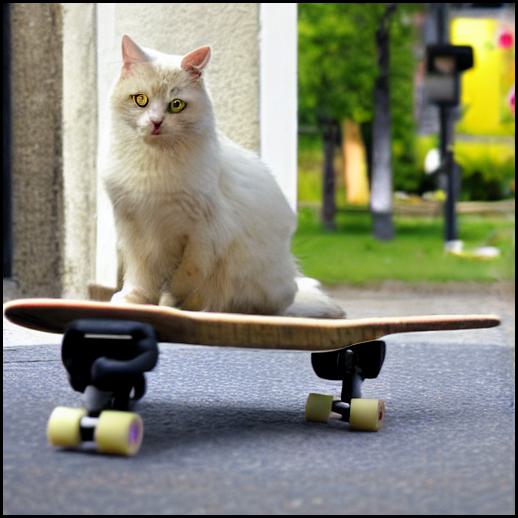}
         \\
        &  & \textit{duck toy} & \textit{swan toy} & \textit{chick toy} \\ 
         \includegraphics[scale=0.17]{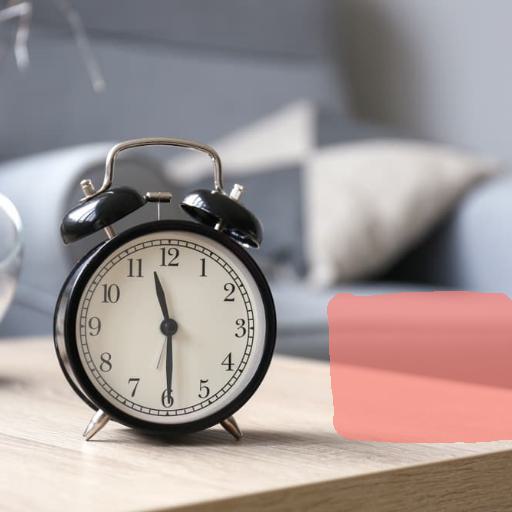}&
          \includegraphics[scale=0.17]{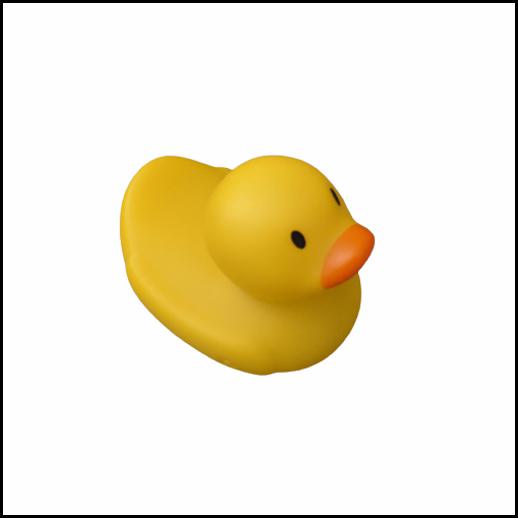}&
           \includegraphics[scale=0.17]{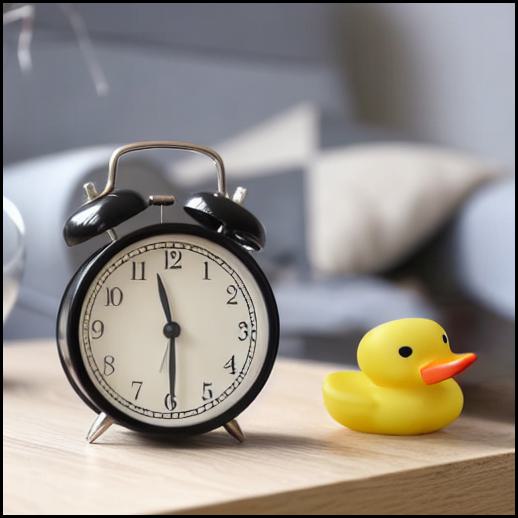}&
         \includegraphics[scale=0.17]{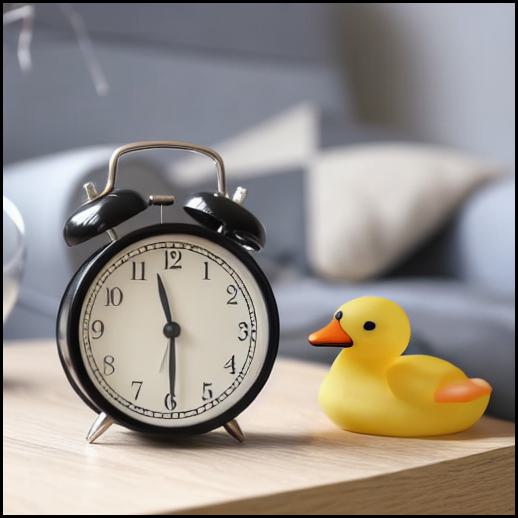} &
         \includegraphics[scale=0.17]{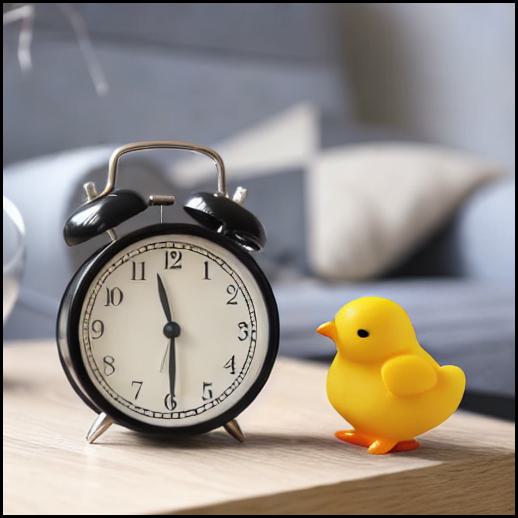}
         \\

          & & \textit{monkey} & \textit{cat} & \textit{bear} \\ 
         \includegraphics[scale=0.17]{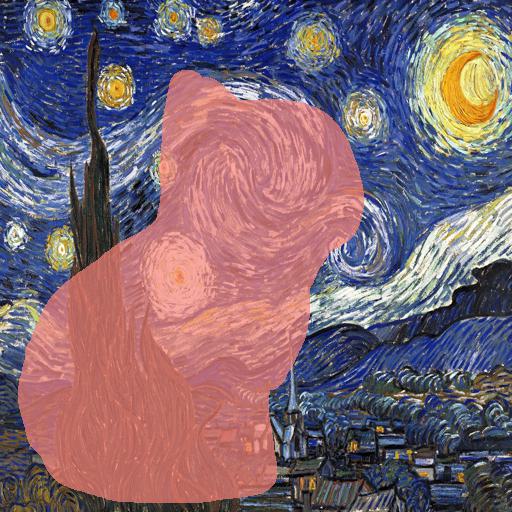}&
          \includegraphics[scale=0.17]{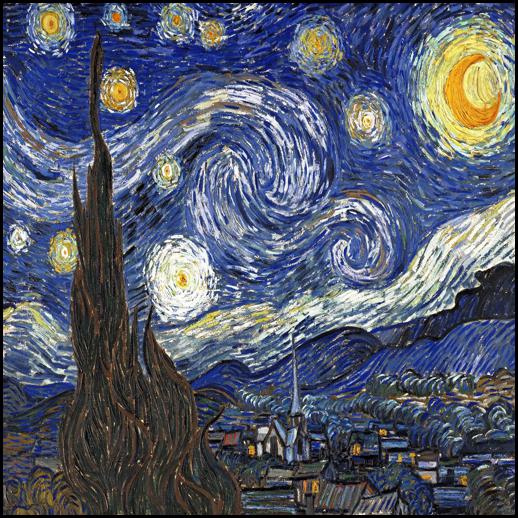}&
           \includegraphics[scale=0.17]{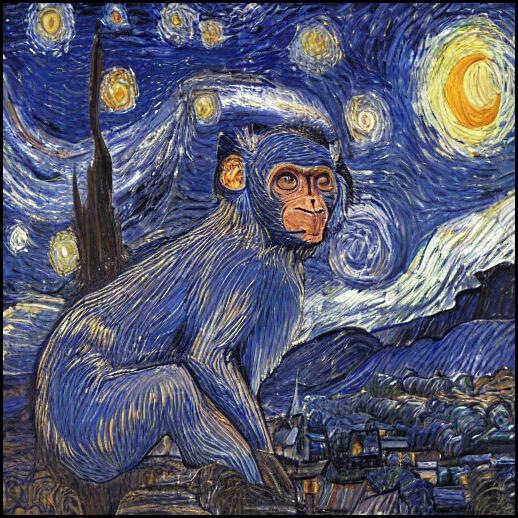}&
         \includegraphics[scale=0.17]{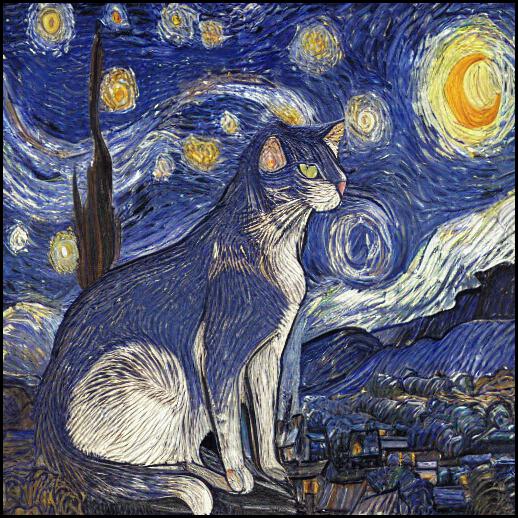} &
         \includegraphics[scale=0.17]{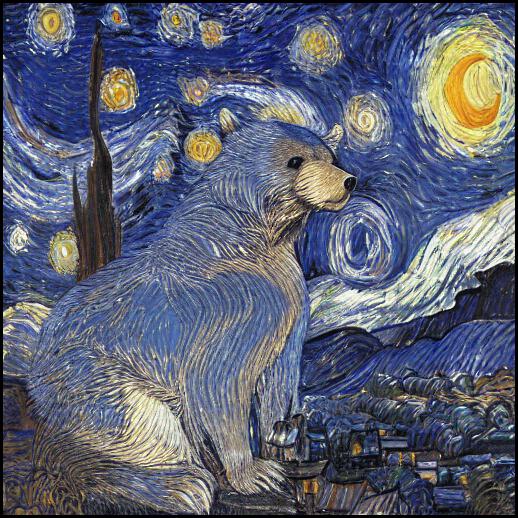}
         \\
         & & \textit{castle} & \textit{maple leaf} & \textit{boat on lake} \\ 
         \includegraphics[scale=0.17]{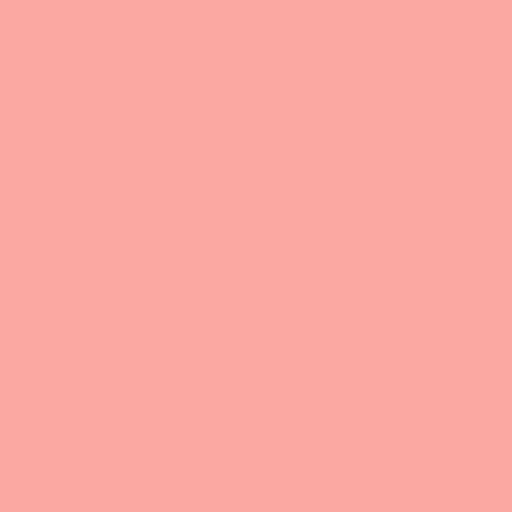}&
          \includegraphics[scale=0.17]{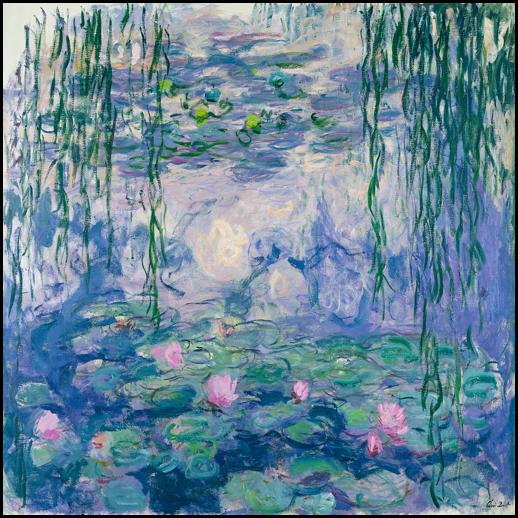}&
           \includegraphics[scale=0.17]{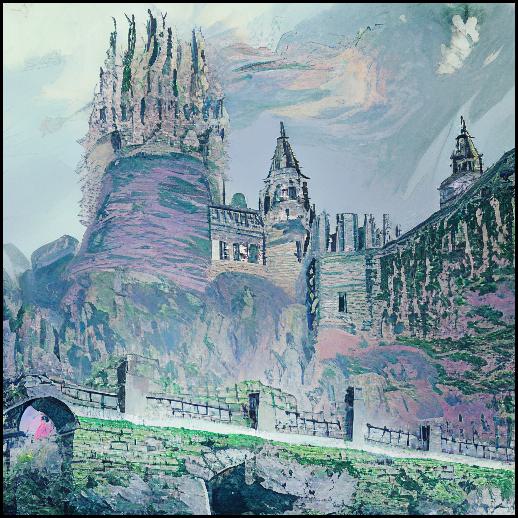}&
         \includegraphics[scale=0.17]{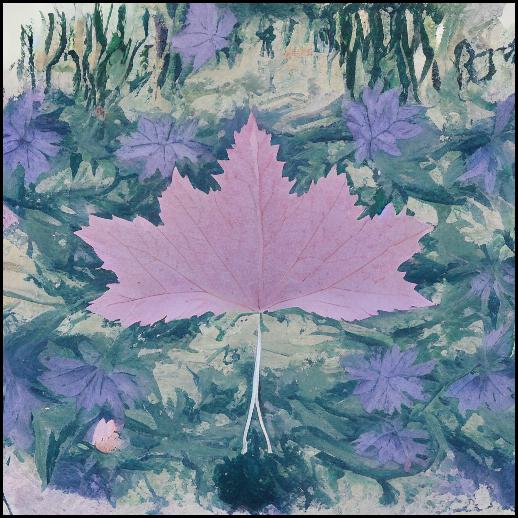} &
         \includegraphics[scale=0.17]{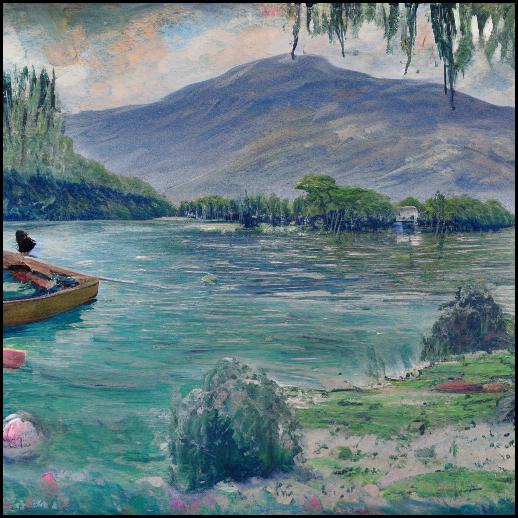}
         \\ \bottomrule
    \end{tabular}
    }
    \vspace{-2mm}
    \caption{Text-guided subject-driven image inpainting results. Note that the strong inpainting baselines~\citep{rombach2022high,yang2023paint} does not support both text and image guidance.}
    \label{fig:ext-example}
\end{figure*}

To disentangle the interdependencies within the training process, we introduce a \textbf{decoupling regularization} technique for multi-modal inpainting. Specifically, we inject noise across the entire image, rather than confining it solely to the segmented area. Furthermore, instead of using only class label as text prompts, we employ the image caption during training. This approach forces the model to acquire the skill of reconstructing the clean image from its noisy counterpart by harnessing both the text prompt and the reference image. For instance, in the case of an image depicting \textit{a dog on a skateboard}, we provide the precise caption in conjunction with the cropped segmentation of the dog. To faithfully restore the skateboard, the network must acquire information from the text prompt. As a result, our trained model also has the ability to introduce a skateboard into the image of the dog subject during the inference phase. We demonstrate such ability in Fig.~\ref{fig:ext-example}.

\subsection{Training and Inference}
\noindent \textbf{Training.}
As depicted in our primary training diagram, illustrated in Fig.~\ref{fig:diagraml}, our approach initiates with the feature extraction step using the UNet downsample stack. We empirically select the output feature from the second downsample block. Subsequently, we choose the top-$K$ tokens, with $K$ being set to 24. These selected tokens are then incorporated by introducing new cross-attention layers into the pretrained models.
In terms of fine-tuning, we fine-tune Stable Diffusion 1.4 on both the OpenImages~\citep{kuznetsova2020open} and MVImageNet dataset~\citep{yu2023mvimgnet}. For the purpose of implementing the decoupling regularization technique, we incorporate the MSCOCO dataset~\citep{lin2014microsoft} due to its inclusion of high-quality image captions. The data allocation ratios for training are set at 60\%, 20\%, and 20\% for each dataset.
During the training process, we utilize a batch size of 512 and conduct training for 40K steps, with learning rate set to $10^{-5}$.

\vspace{2mm} \noindent \textbf{Inference.}
During inference, we employ the DDIM sampler~\citep{song2020denoising} with 50 steps. In the context of jointly considering image and text conditions, we have observed that consistently conditioning on the image context yields significant benefits for subject-driven inpainting. Specifically, we calculate the predicted noise as follows:
\begin{equation}
    \epsilon_{\theta}(x_t, \varnothing,  c_{x_{r}})+w(\epsilon_{\theta}(x_t, c,  c_{x_{r}}) - \epsilon_{\theta}(x_t, \varnothing,  c_{x_{r}}))
\end{equation}
where $w$ is the guidance scale and we set it as 7.5 and $\varnothing$ is an empty string serving as the negative prompt.

\section{Experiments}
\label{sec:exp}

In this section, we first introduce the evaluation protocols, followed by the discussion of our performance on the proposed task. We then provide comparison with the state-of-the-art methods on the conventional subject-driven inpainting task. Finally, we verify the effectiveness of the proposed components through ablation studies.

\subsection{Experimental Setup}
\vspace{2mm} \noindent \textbf{Datasets.}
We perform evaluation on two datasets. COCOEE dataset~\citep{yang2023paint}, which selects 3500 similar images from MSCOCO validation dataset as reference image. But the reference and input image are not guaranteed to be the same identity.
To further verify our capability on identity preservation, we construct a dataset based on the subject-driven image generation dataset DreamBooth~\citep{ruiz2023dreambooth}. Specifically, for each subject, we have several images and we use one image as the input and the remaining images as the reference.
In total, we obtain 684 testing pairs and we denote the constructed dataset as DreamBoothEE. 

\vspace{2mm} \noindent \textbf{Baselines.}
We compare with two subject-driven inpainting methods, PBE~\citep{yang2023paint} and Blended Diffusion~\citep{avrahami2022blended}, and one text-guided inpainting method Stable Inpaint~\citep{rombach2022high}. BLIP~\citep{li2022blip} used to generate the image captions, which are then used as the text prompts for inpainting. 

\vspace{2mm} \noindent \textbf{Metrics.}
Our evaluation process begins with the calculation of the Fréchet Inception Distance (FID)~\citep{heusel2017gans} between the generated output images and their corresponding reference images. This metric, denoted as R-FID, allows us to assess the realism of the output images while considering their dependence on the reference images.
Additionally, we compute embeddings using two different methods, namely CLIP~\citep{radford2021learning} and DINO~\citep{caron2021emerging}, for both the generated output images and the background-masked reference images. This approach ensures that we capture the essential features of the subject while avoiding undesirable influences from the background in the reference images. These embeddings are denoted as F-CLIP and F-DINO, respectively.

\subsection{Text-Guided Subject-Driven Inpainting}
It's important to highlight that, to the best of our knowledge, there are no existing baselines capable of simultaneously supporting both text-guided and subject-driven inpainting. As shown in Figure~\ref{fig:ext-example}, the proposed task inherit multiple downstream applications, which has not been discussed in previous inpainting literature. 

\vspace{1mm}
\noindent\textbf{Creative Subject Inpainting.}
The first application is to insert the subject into the specified region. Our DreamInpainter is able to follow text decriptions, allowing us to modify the pose and appearence of the subject, as shown in the first row of Fig.~\ref{fig:ext-example}.

\vspace{1mm}
\noindent\textbf{Property Transformation.}
As shown in the second and third row, thanks to the strong text-understnding capability of DreamInpainter, we are able to alter the property of the subject without losing its identity. For example, we can generate a chick toy while retaining essential information from the reference duck toy. 

\vspace{1mm}
\noindent\textbf{Stylized Text-Guided Inpainting.}
Our model enables an interesting application of partially or completely inserting content to a stylized image under text guidance. In particular, as shown in the fourth row, we could coherently insert different subjects to the style image. We can also generate an image with the same style given texts through masking the entire image.

\begin{figure*}
    \centering
    \setlength{\tabcolsep}{2pt}
    \footnotesize
    \scalebox{0.85}{
   \begin{tabular}{ccccccc}
   Input &  Reference & Stable Inpaint~\citep{rombach2022high} &Blended~\citep{avrahami2022blended} & PBE~\citep{yang2023paint} & DreamInpainter (Ours) \\
       \includegraphics[height=2.6cm, width=2.6cm]{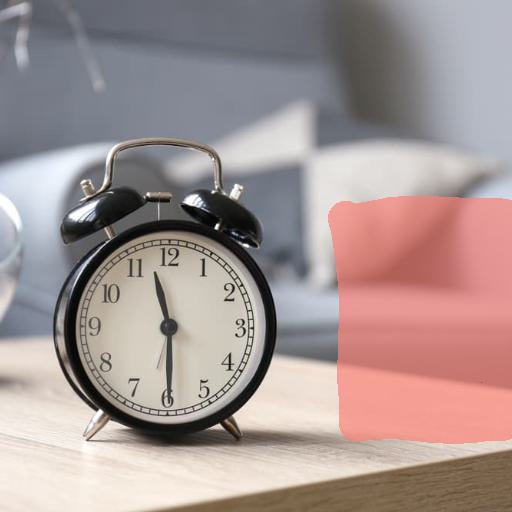} &
       \includegraphics[height=2.6cm, width=2.6cm]{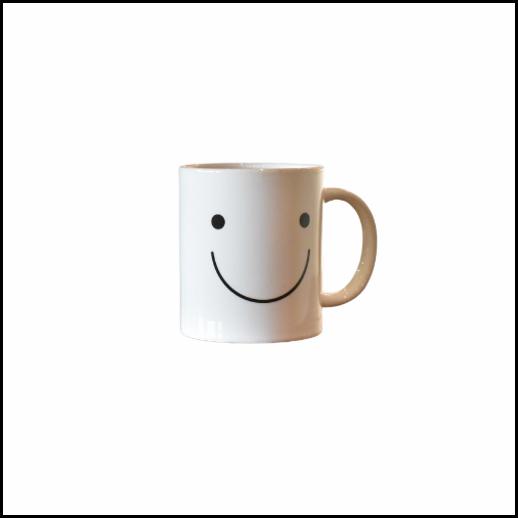} &
       \includegraphics[height=2.6cm, width=2.6cm]{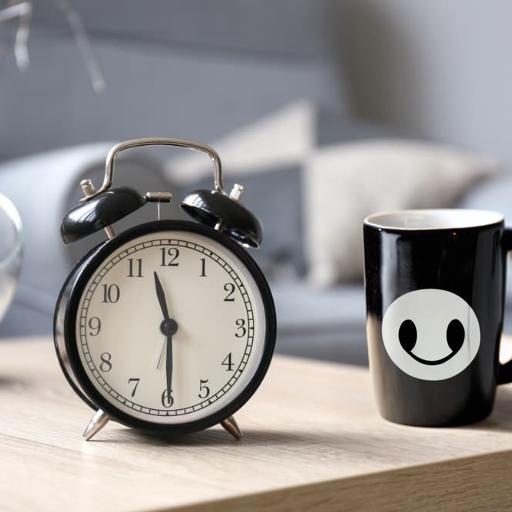} & 
        \includegraphics[height=2.6cm, width=2.6cm]{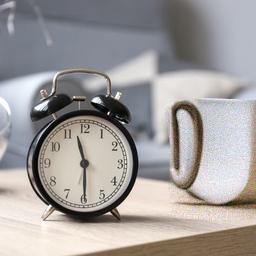} &
       \includegraphics[height=2.6cm, width=2.6cm]{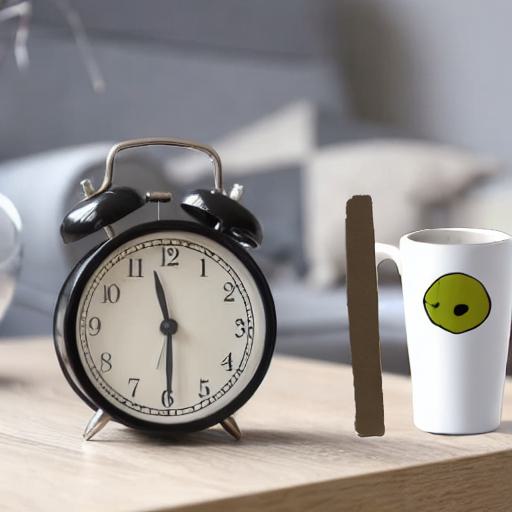} & 
       \includegraphics[height=2.6cm, width=2.6cm]{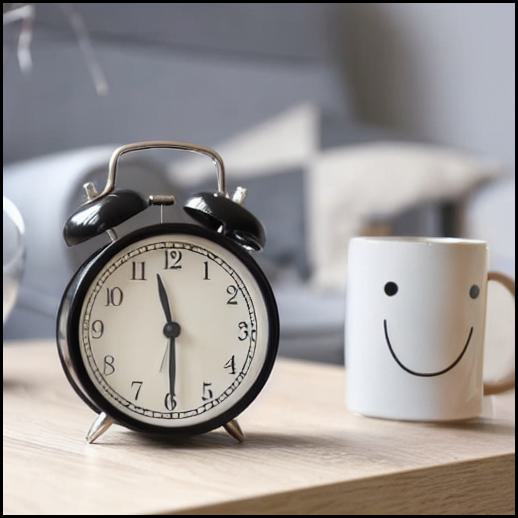} &
       \\ 
        \includegraphics[height=2.6cm, width=2.6cm]{figs/image_guided/input2.jpg} &
       \includegraphics[height=2.6cm, width=2.6cm]{figs/image_guided/ref2_1.jpg} & 
        \includegraphics[height=2.6cm, width=2.6cm]{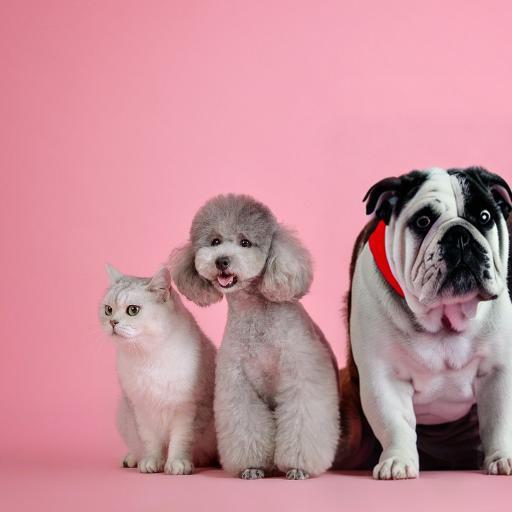} & 
        \includegraphics[height=2.6cm, width=2.6cm]{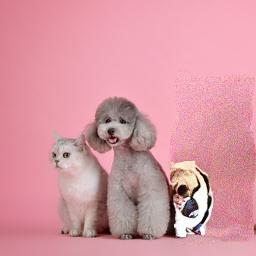} & 
       \includegraphics[height=2.6cm, width=2.6cm]{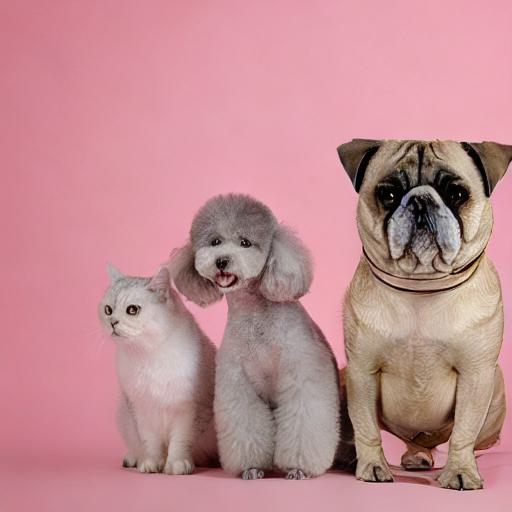} & 
       \includegraphics[height=2.6cm, width=2.6cm]{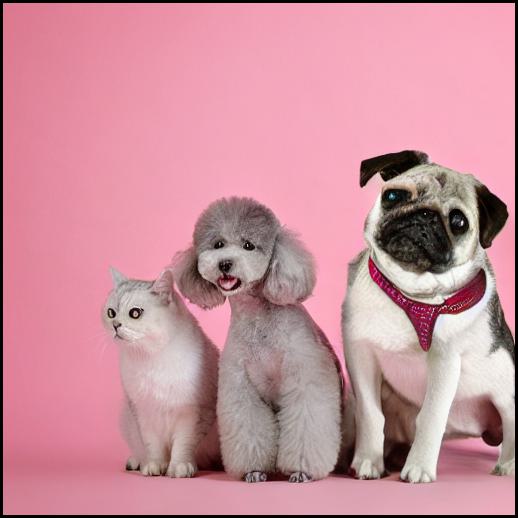} &
       \\ 
        \includegraphics[height=2.6cm, width=2.6cm]{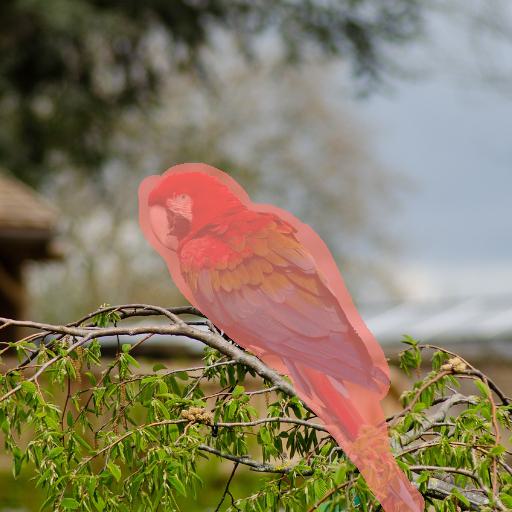} &
       \includegraphics[height=2.6cm, width=2.6cm]{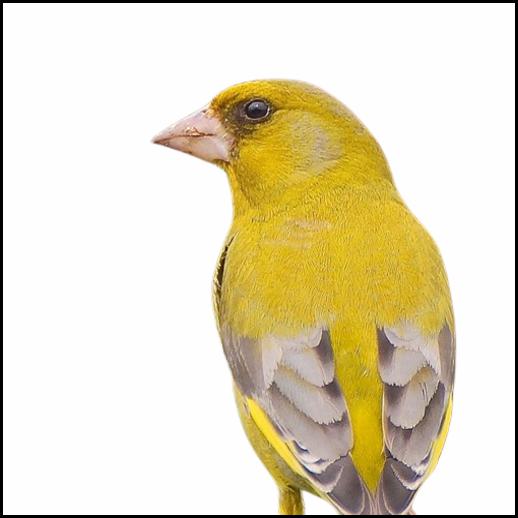} & 
         \includegraphics[height=2.6cm, width=2.6cm]{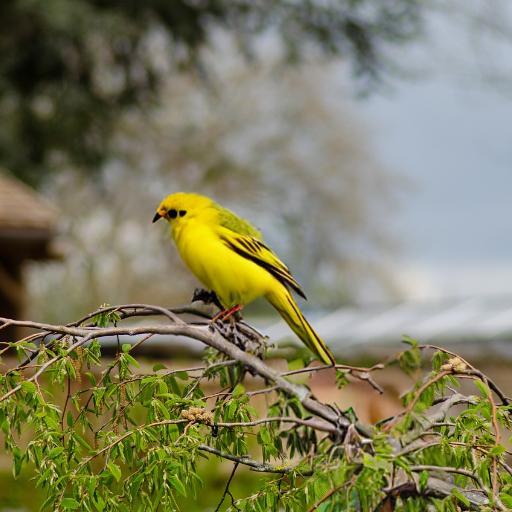} & 
        \includegraphics[height=2.6cm, width=2.6cm]{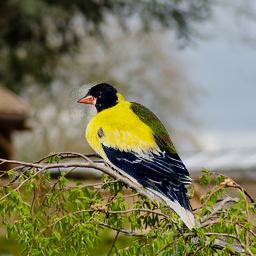} & 
       \includegraphics[height=2.6cm, width=2.6cm]{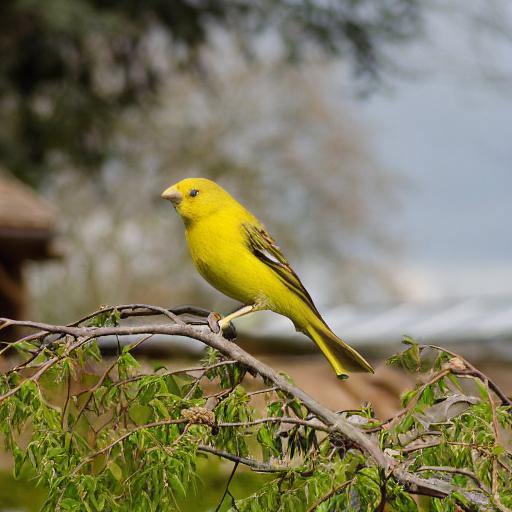} & 
       \includegraphics[height=2.6cm, width=2.6cm]{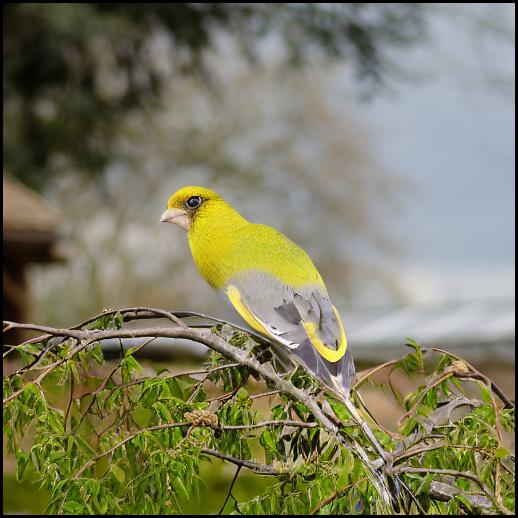} &
       \\ 
        \includegraphics[height=2.6cm, width=2.6cm]{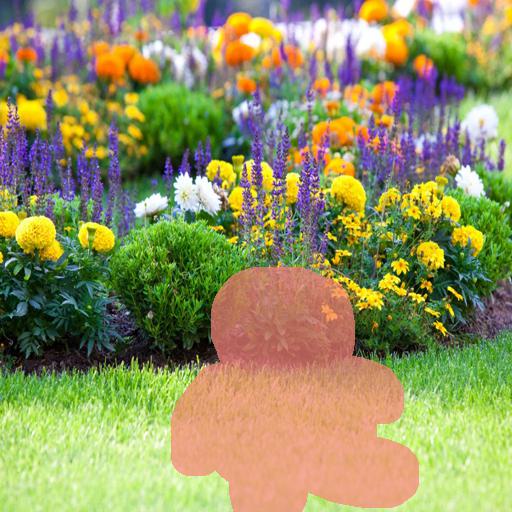} &
       \includegraphics[height=2.6cm, width=2.6cm]{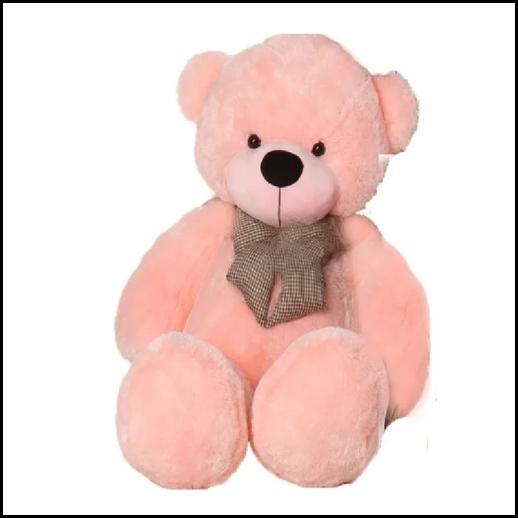} & 
        \includegraphics[height=2.6cm, width=2.6cm]{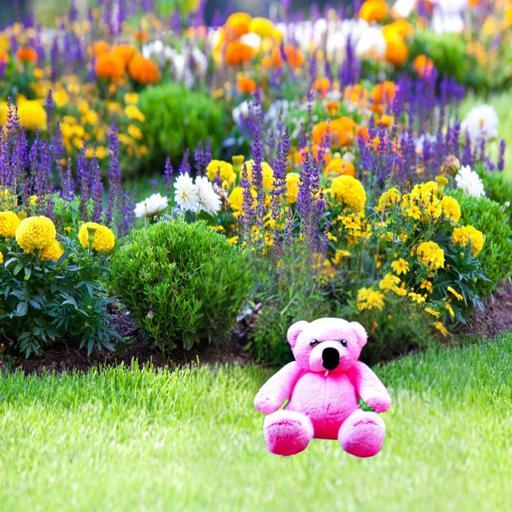} & 
        \includegraphics[height=2.6cm, width=2.6cm]{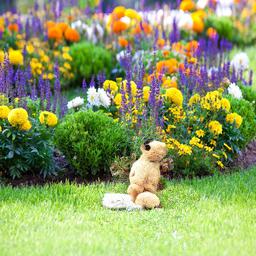} & 
       \includegraphics[height=2.6cm, width=2.6cm]{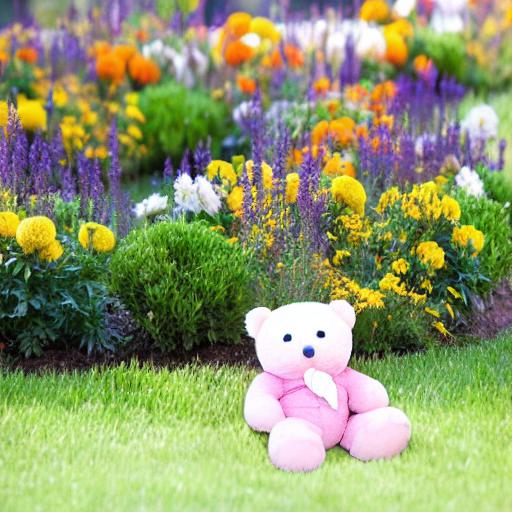} & 
       \includegraphics[height=2.6cm, width=2.6cm]{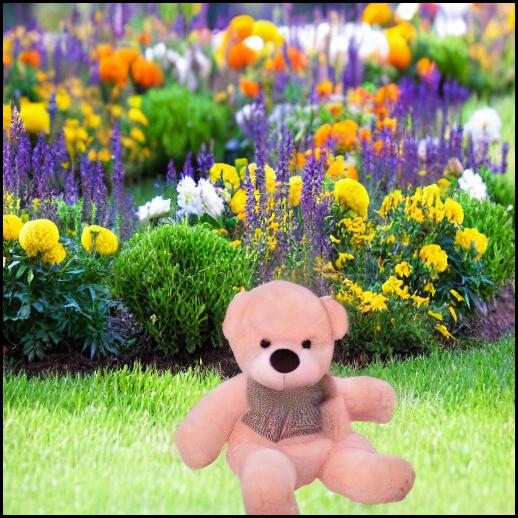} &
       \\ 
        \includegraphics[height=2.6cm, width=2.6cm]{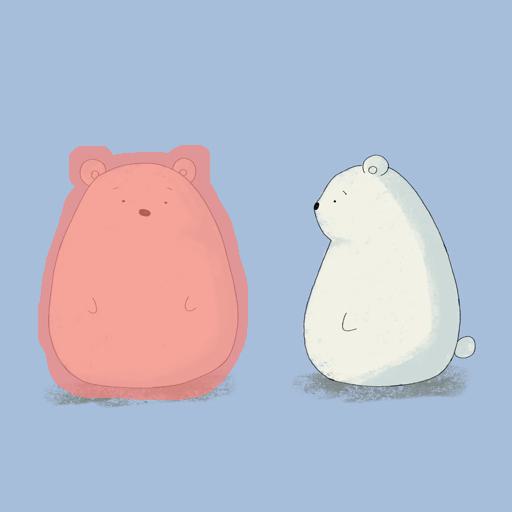} &
       \includegraphics[height=2.6cm, width=2.6cm]{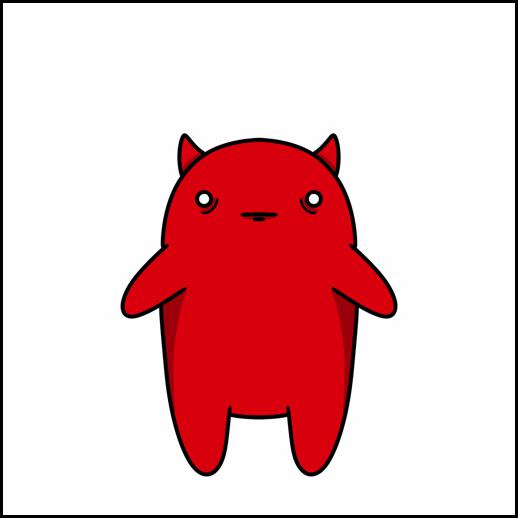} & 
       \includegraphics[height=2.6cm, width=2.6cm]{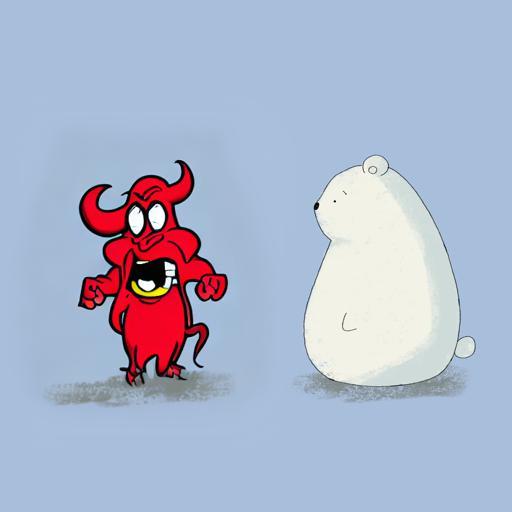} & 
        \includegraphics[height=2.6cm, width=2.6cm]{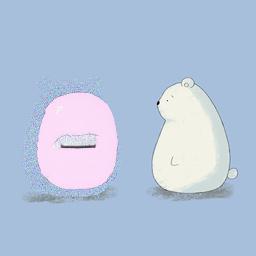} &
       \includegraphics[height=2.6cm, width=2.6cm]{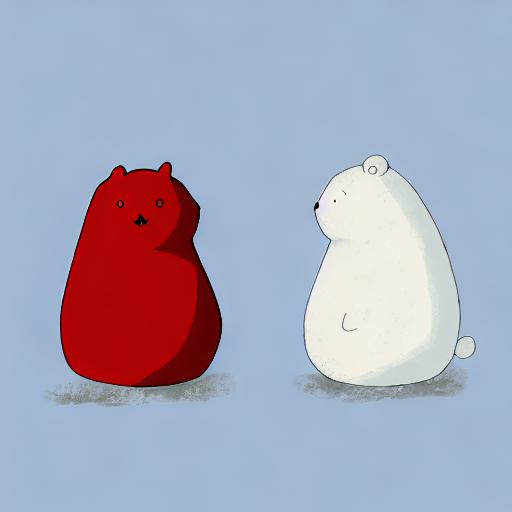} & 
       \includegraphics[height=2.6cm, width=2.6cm]{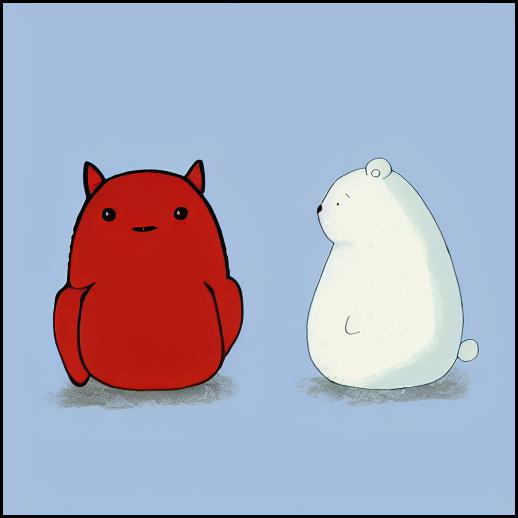} &
       \\ 

       \\ 
        \includegraphics[height=2.6cm, width=2.6cm]{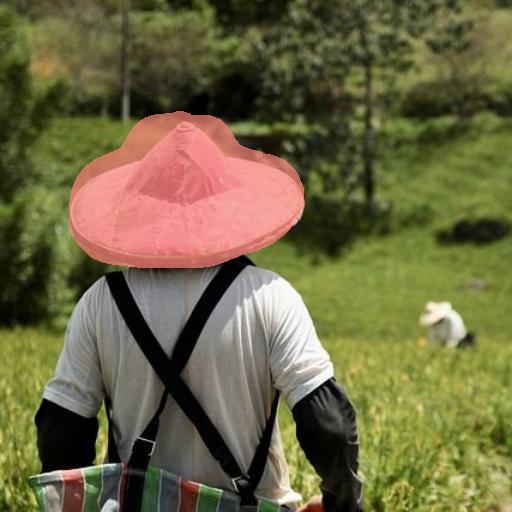} &
       \includegraphics[height=2.6cm, width=2.6cm]{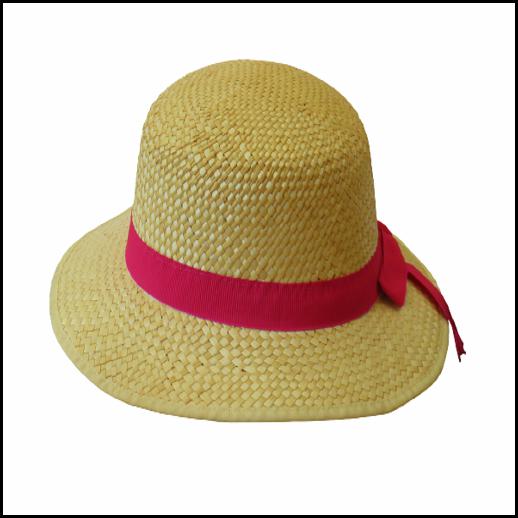} & 
       \includegraphics[height=2.6cm, width=2.6cm]{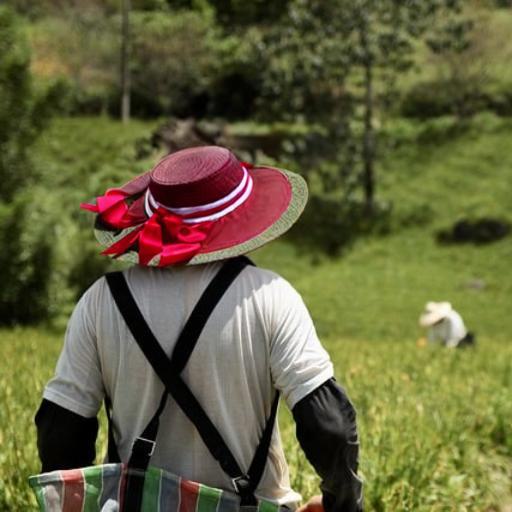} & 
        \includegraphics[height=2.6cm, width=2.6cm]{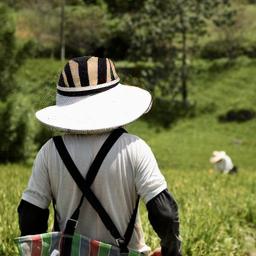} &
       \includegraphics[height=2.6cm, width=2.6cm]{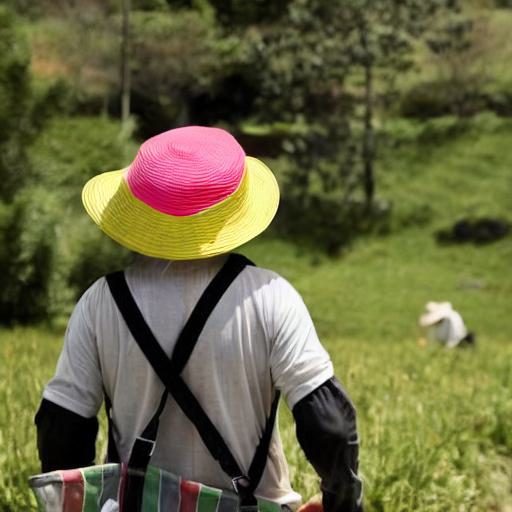} & 
       \includegraphics[height=2.6cm, width=2.6cm]{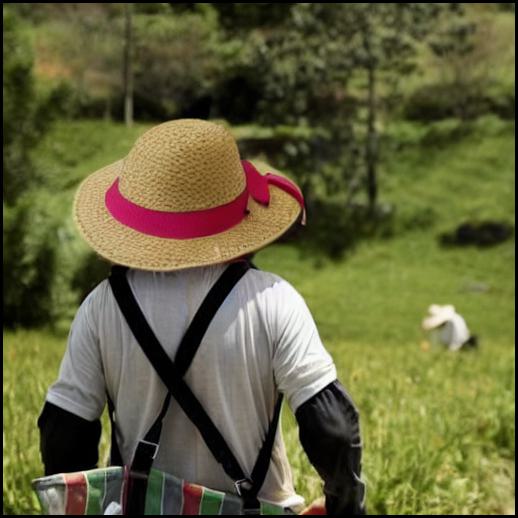} &
         \\ 

   \end{tabular}
       }
   \vspace{-2mm}
    \caption{Comparison to state-of-the-art inpainting methods. Stable Inpaint~\citep{rombach2022high} takes text as condition and Blended Diffusion~\citep{avrahami2022blended} and Paint-by-Example (PBE)~\citep{yang2023paint} take reference images as condition. By contrast, our method takes both text and image. We just use a short word as our text input, such as \textit{cup}, \textit{dog} and \textit{bird}. The image feature $c_{x_r}$ will inject the detailed information. } 

    \label{fig:comp}
\end{figure*}


\subsection{Comparison with Baselines}
We also compare our method with baselines on the well-established inpainting tasks with single guidance.
We visualize the generated samples in Fig.~\ref{fig:comp}. Our method better preserves identity of the input reference images compared to PBE~\citep{yang2023paint}. Importantly, instead of copying the object to the target image, our method is capable of generating content aligning with the mask shape. For example, in the fourth row, our method successfully generate a teddy bear that fits with the mask, which is different from its original shape, while retaining its identity. These successful examples demonstrate that our proposed token selection module effectively avoids the copy-paste artifacts without sacrificing identity. We also present the quantitative results in Tab.~\ref{tab:res_cocoee} and \ref{tab:res_dreambooth}. We are unable to compute FID on DreamboothEE dataset since it only contains 684 images. The large improvement on DINO~\citep{caron2021emerging} and CLIP~\citep{radford2021learning} score show that the output images generated by our method aligns much better with the reference subjects than the baseline methods.

\begin{table}[!ht]
    \centering
    \scalebox{0.85}{
    \begin{tabular}{l|ccc}
    \toprule
    Method & R-FID $\downarrow$ & F-CLIP $\uparrow$ & F-DINO $\uparrow$   \\  \midrule
    Stable-Inpaint~\citep{rombach2022high} & 11.38 & 70.47 &31.62\\ 
    PBE~\citep{yang2023paint} & 10.79 & 74.94 & 44.64\\ \midrule
    DreamInpainter (Ours) & \textbf{10.68} & \textbf{76.17} & \textbf{51.56} \\ \bottomrule
    \end{tabular}
    }
    \vspace{-2mm}
    \caption{Results on COCOEE dataset~\citep{yang2023paint}.}
    \label{tab:res_cocoee}
\end{table}

\begin{table}[!ht]
    \centering
    \scalebox{0.85}{
    \begin{tabular}{l|cc}
    \toprule
    Method & F-CLIP $\uparrow$   & F-DINO $\uparrow$\\  \midrule
    Stable-Inpaint~\citep{rombach2022high} &90.7 & 75.37 \\ 
    PBE~\citep{yang2023paint} & 91.7 & 79.77\\ \midrule
    DreamInpainter (Ours) &   \textbf{92.5}  & \textbf{82.22}\\ \bottomrule
    \end{tabular}
    }
    \vspace{-2mm}
    \caption{Results on DreamboothEE dataset. }
    \label{tab:res_dreambooth}
\end{table}

\begin{figure*}[!ht]
    \centering
    \scalebox{0.65}{
    \begin{tabular}{cc|cc|cc}
    \toprule
        \multirow{2}{*}{Input}  & \multirow{2}{*}{Reference} & \multicolumn{2}{c|}{With regularization} & \multicolumn{2}{c}{Without regularization} \\ 
        & & \textit{cat} & \textit{raccoon} & \textit{cat} & \textit{raccoon}\\ \midrule
        \includegraphics[width=0.2\linewidth]{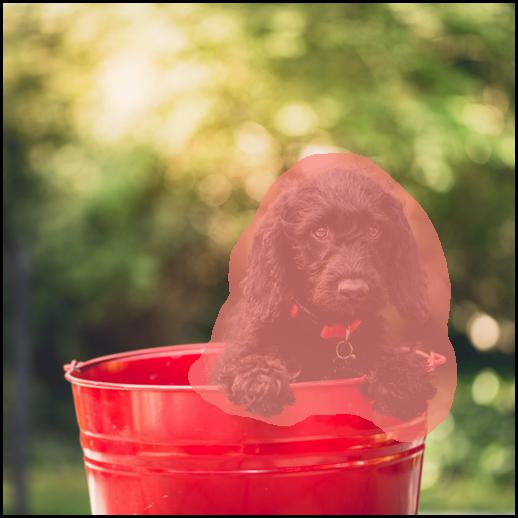} & 
         \includegraphics[width=0.2\linewidth]{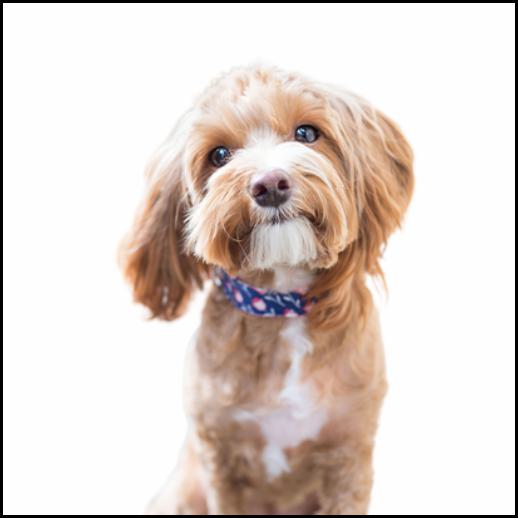} & 
           \includegraphics[width=0.2\linewidth]{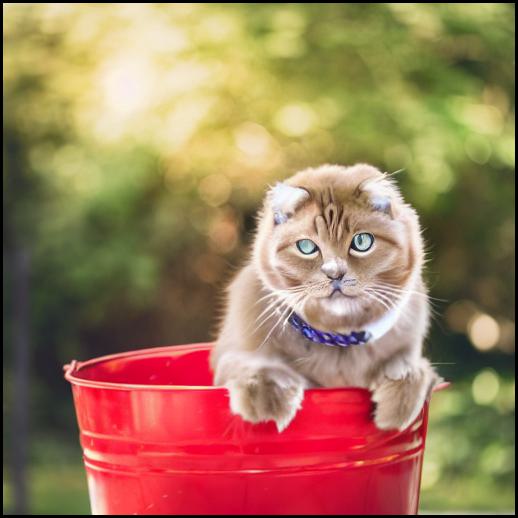} & 
           \includegraphics[width=0.2\linewidth]{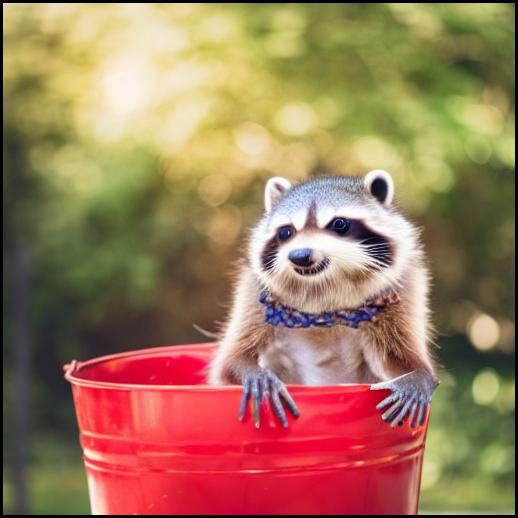} & 
            \includegraphics[width=0.2\linewidth]{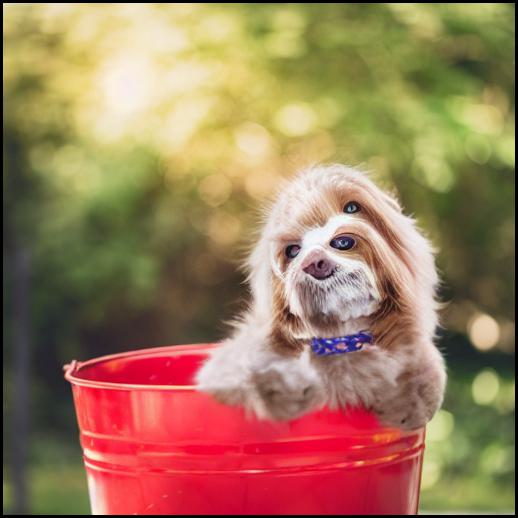} &
         \includegraphics[width=0.2\linewidth]{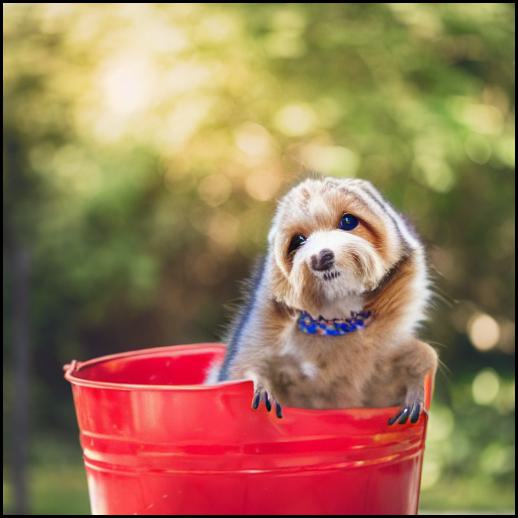}

         \\ \bottomrule

    \end{tabular}
    }
    \vspace{-2mm}
    \caption{Ablation of using decoupling regularization.}
    \label{fig:decouplel_ablation}
\end{figure*}
\begin{table}[ht]
    \centering
    \scalebox{0.85}{
    \begin{tabular}{l|cc}
    \toprule
    Method  & R-FID $\downarrow$  &F-CLIP $\uparrow$ \\  \midrule
    Baseline Unet Feature& 17.34 & 75.02 \\ 
    + Token Selection &\textbf{10.33}&74.62 \\
    + decoupling Regularization &10.68& \textbf{76.17}  \\ \bottomrule
    \end{tabular}
    }
    \caption{Ablation results on COCOEE dataset~\citep{yang2023paint}.}
    \label{tab:ablation}
\end{table}

\begin{figure}
    \centering
    \footnotesize
     \setlength{\tabcolsep}{2pt}
     \scalebox{0.85}{
    \begin{tabular}{cccc}
         Input & Reference & $K=1$ & $K=4$ \\  
         \includegraphics[scale=0.11]{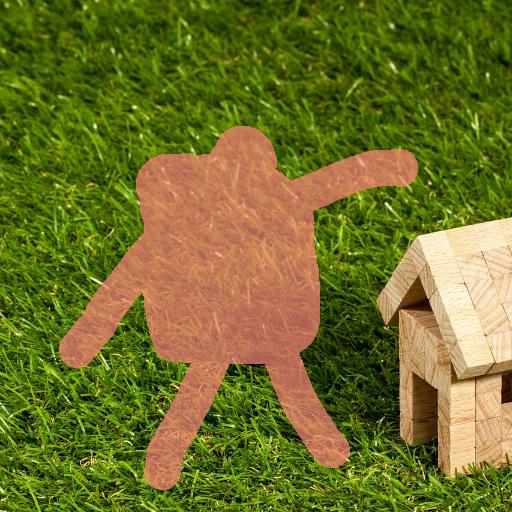}
         &  \includegraphics[scale=0.11]{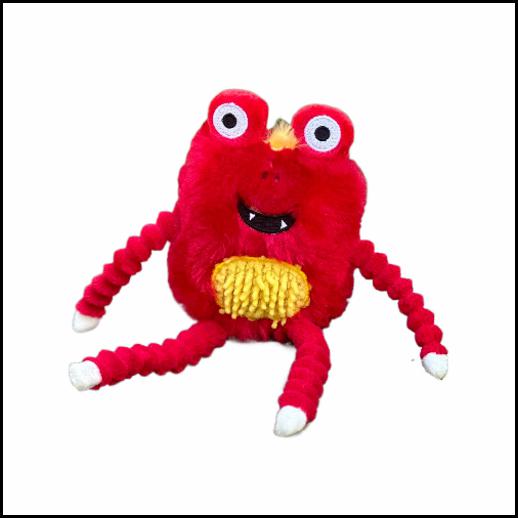}
         & \includegraphics[scale=0.11]{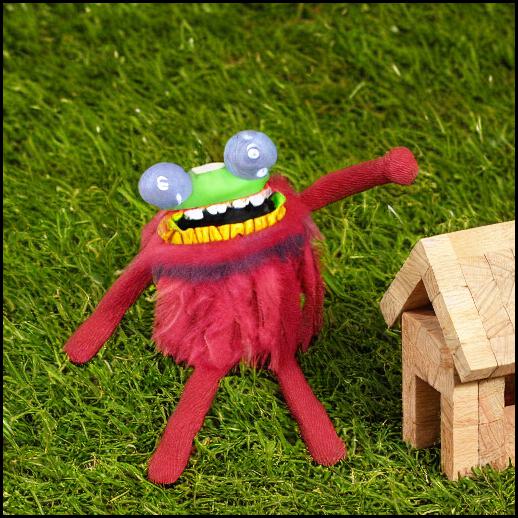} &
\includegraphics[scale=0.11]{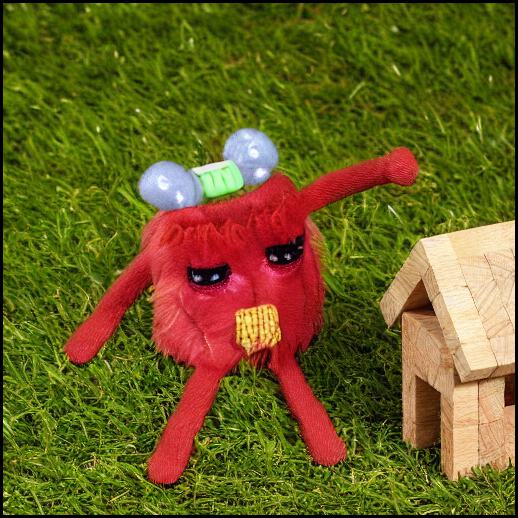}
        \\
         $K=8$ & $K=16$ &$K=24$ & $K=48$  \\
          \includegraphics[scale=0.11]{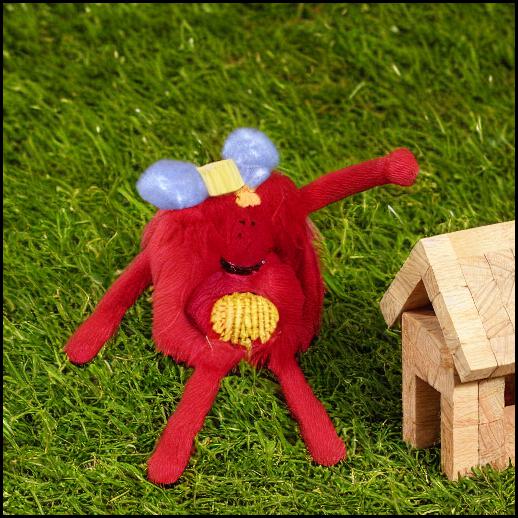}  
            & \includegraphics[scale=0.11]{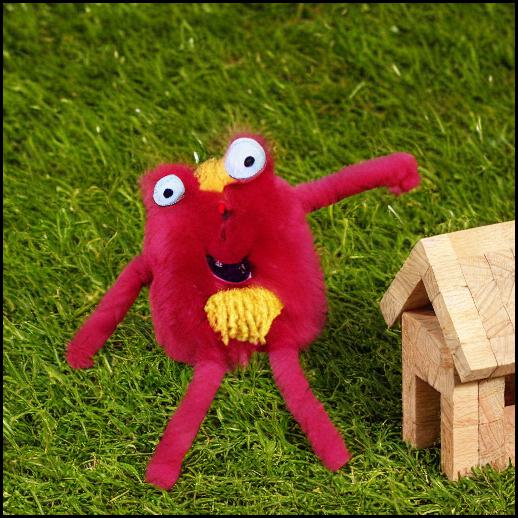}
         & \includegraphics[scale=0.11]{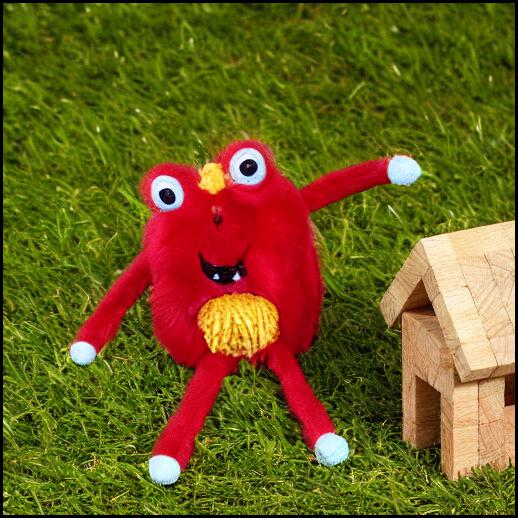} 
         & \includegraphics[scale=0.11]{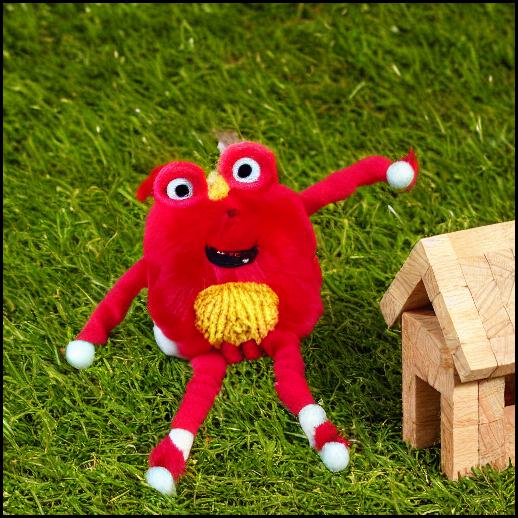} 
        \\
    \end{tabular}
    }
    \caption{The generation versus number of tokens $K$ selected.}
    \label{fig:num_token_vis}
\end{figure}

\subsection{Analysis and Ablation Study}
\begin{figure}[ht]
    \centering
    \setlength{\tabcolsep}{2pt}
    \scalebox{0.85}{
  \begin{tabular}{cccc}
        \includegraphics[scale=0.13]{figs/image_guided/ref4_1.jpg}&
       \includegraphics[scale=0.13]{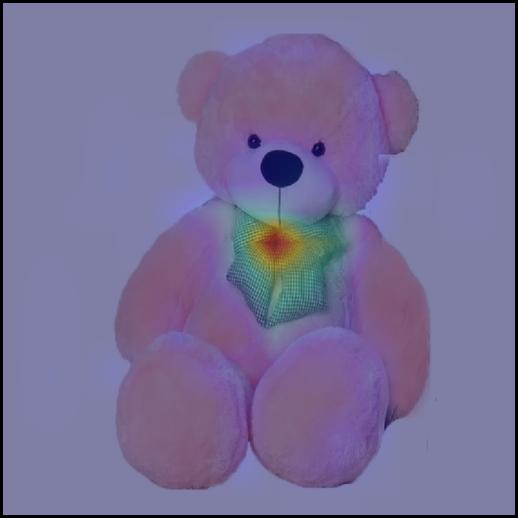} &
         \includegraphics[scale=0.13]{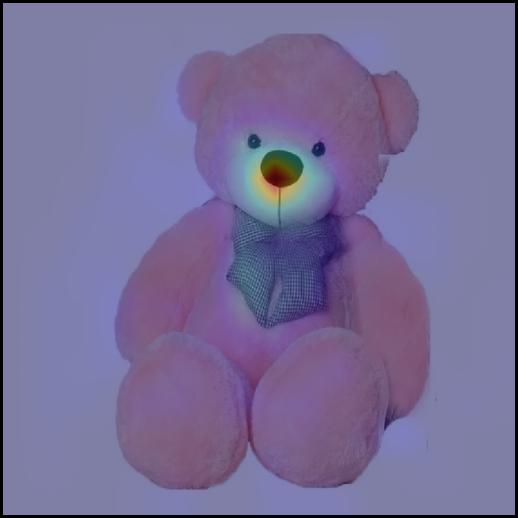}\\
        \includegraphics[scale=0.13]{figs/text_guide/ref2_1.jpg}&
       \includegraphics[scale=0.13]{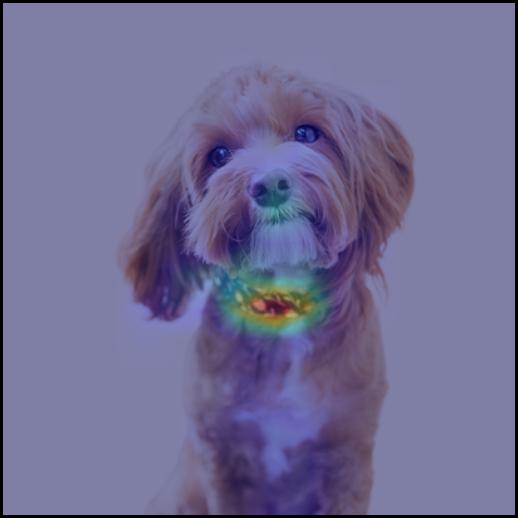} &
         \includegraphics[scale=0.13]{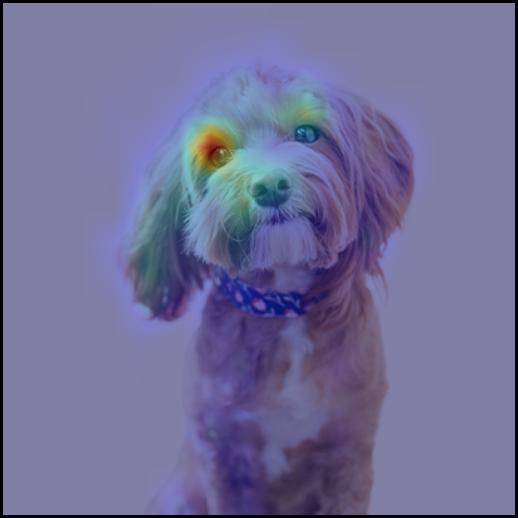}\\
  \end{tabular}
  }
    \caption{The visualization of the selected tokens. Our method learns to focus on special regions of the reference object and preserves identity while allowing certains changes for inpainting. \textbf{(Best viewed in color)}}
    \label{fig:vis_token}
\end{figure}

In this section, we analyze the importance of the proposed token selection module and decoupling regularization. We visualize the tokens and present two selected tokens in Fig.~\ref{fig:vis_token}. Our method is able to select the most discriminative regions of the whole image and thus preserve the identity of the input image. In addition, we also avoid the copy-paste problem since we only select the top $K$ tokens instead of using all high-dimensional features.

On the other hand, as shown in Table~\ref{tab:res_cocoee} and Table~\ref{tab:res_dreambooth}, the proposed decoupling regularization improves the result by a large margin in terms of DINO score. We also present examples without the regularization in Fig.~\ref{fig:decouplel_ablation}. The generations are almost identical regardless of different text prompts, showing that the model ignores the text prompt and relies on the reference image feature heavily. By contrast, the second rows shows the result with the regularization. The method learns to generates different animals while keeping the unique part of the reference dog. 

We also present the generated samples under different number of selected tokens $K$ in Fig.~\ref{fig:num_token_vis}. 
We notice that the network learns to recover the identity as we increase the value of $K$. With small $K$, such as 1 and 4, the model fail to capture the identity of the input toy. However, if we use excessive tokens, artifacts are observed and we conjecture that it is caused by the unnecessary subject information. We achieve best result when we use 24 tokens. This demonstrates the necessity of the proposed token selection module. 


\begin{figure}[!h]
    \centering
    \setlength{\tabcolsep}{2pt}
    \scalebox{0.85}{
    \begin{tabular}{cccc}
    Input & Reference & PBE \citep{yang2023paint} & Ours\\
         \includegraphics[scale=0.11]{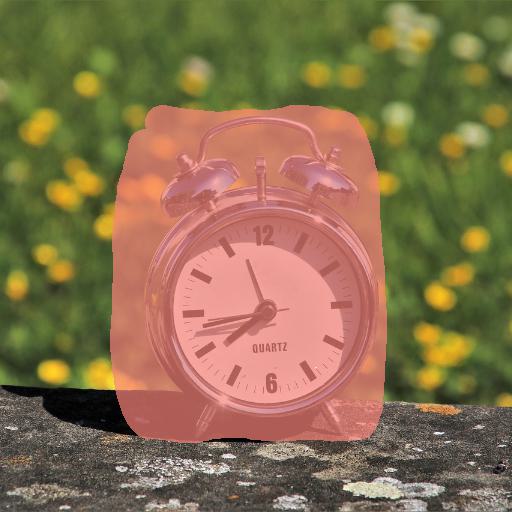}
         &  
          \includegraphics[scale=0.11]{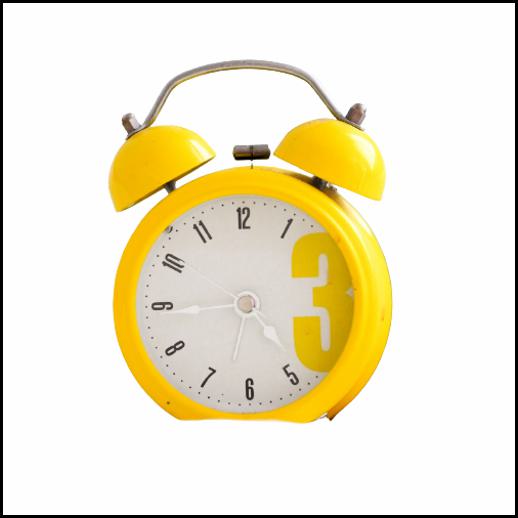}
          &
           \includegraphics[scale=0.11]{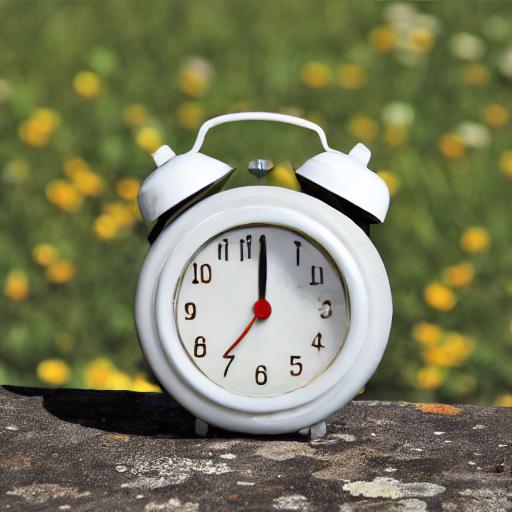}
           &
            \includegraphics[scale=0.11]{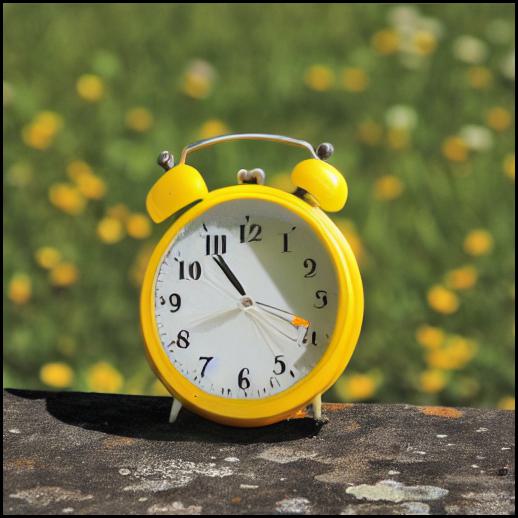}
         \\
          \includegraphics[scale=0.11]{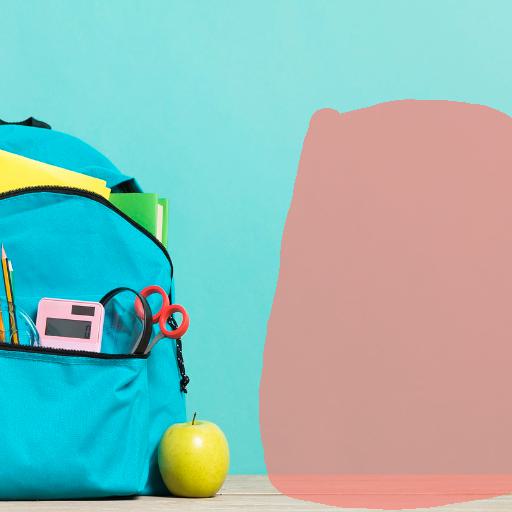}
         &  
          \includegraphics[scale=0.11]{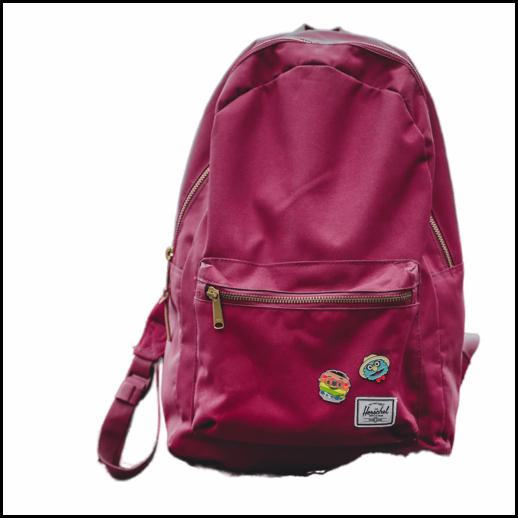}
          &
           \includegraphics[scale=0.11]{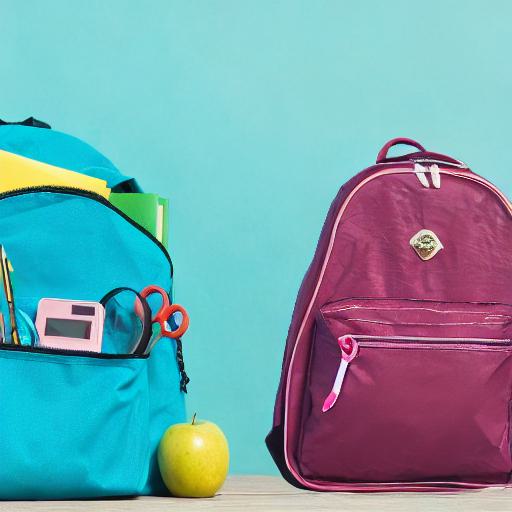}
           &
            \includegraphics[scale=0.11]{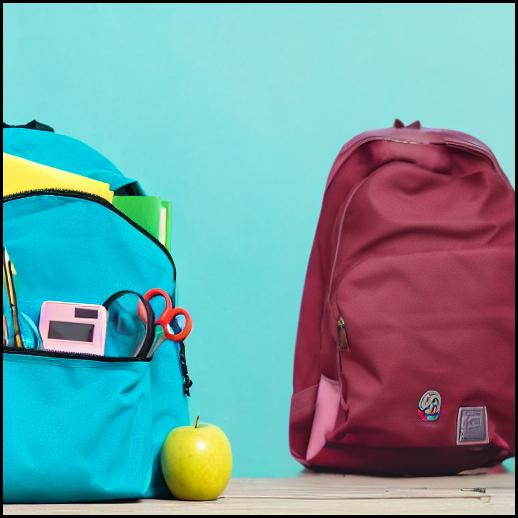}
         \\
    \end{tabular}
    }
    \caption{Limitation of our method. We may lose some wanted details for complex reference objects after fixed number of token selection.}
    \label{fig:limit}
\end{figure}
\section{Discussion and Conclusion}
\label{sec:conclusion}

While our method has demonstrated the ability to generate high-quality images through the use of token selection and decoupled regularization, it still encounters challenges when preserving intricate details of the reference object. As illustrated in Fig.~\ref{fig:limit}, our approach performs notably better than PBE~\citep{yang2023paint}, but we still struggle to fully recover the fine details of the reference clock and the logos on the backpack. This limitation may stem from the fact that the number of selected tokens is fixed and might not encompass all the intricate information about the object.
However, it's worth noting that using too many tokens could result in copy-paste artifacts, particularly for simpler objects. Currently, we adhere to a fixed number of tokens during training. In future work, it could be advantageous to explore the concept of adaptive token selection, where the number of selected tokens varies based on the complexity of different samples.

In summary, our approach introduces  the text-guided subject-driven image inpainting task with the goal of preserving the identity of reference images while adhering to other guidance signals, including user masks and additional text prompts. To tackle this challenge, we present DreamInpainter, which consists of two key components: a discriminative token selection module and a decoupling regularization technique. Our experimental results demonstrate that these modules effectively prevent trivial mappings and enable the model to generate high-quality content that aligns with both the reference image and the provided text prompt.


\clearpage
\newpage
{
    \small
    \bibliographystyle{ieeenat_fullname}
    \bibliography{main}

\begin{thebibliography}{52}
\providecommand{\natexlab}[1]{#1}
\providecommand{\url}[1]{\texttt{#1}}
\expandafter\ifx\csname urlstyle\endcsname\relax
  \providecommand{\doi}[1]{doi: #1}\else
  \providecommand{\doi}{doi: \begingroup \urlstyle{rm}\Url}\fi

\bibitem[Avrahami et~al.(2022)Avrahami, Lischinski, and Fried]{avrahami2022blended}
Omri Avrahami, Dani Lischinski, and Ohad Fried.
\newblock Blended diffusion for text-driven editing of natural images.
\newblock In \emph{Proceedings of the IEEE/CVF Conference on Computer Vision and Pattern Recognition}, pages 18208--18218, 2022.

\bibitem[Balaji et~al.(2022)Balaji, Nah, Huang, Vahdat, Song, Kreis, Aittala, Aila, Laine, Catanzaro, et~al.]{balaji2022ediffi}
Yogesh Balaji, Seungjun Nah, Xun Huang, Arash Vahdat, Jiaming Song, Karsten Kreis, Miika Aittala, Timo Aila, Samuli Laine, Bryan Catanzaro, et~al.
\newblock ediffi: Text-to-image diffusion models with an ensemble of expert denoisers.
\newblock \emph{arXiv preprint arXiv:2211.01324}, 2022.

\bibitem[Blattmann et~al.(2023)Blattmann, Rombach, Ling, Dockhorn, Kim, Fidler, and Kreis]{blattmann2023align}
Andreas Blattmann, Robin Rombach, Huan Ling, Tim Dockhorn, Seung~Wook Kim, Sanja Fidler, and Karsten Kreis.
\newblock Align your latents: High-resolution video synthesis with latent diffusion models.
\newblock In \emph{Proceedings of the IEEE/CVF Conference on Computer Vision and Pattern Recognition}, pages 22563--22575, 2023.

\bibitem[Caron et~al.(2021)Caron, Touvron, Misra, J{\'e}gou, Mairal, Bojanowski, and Joulin]{caron2021emerging}
Mathilde Caron, Hugo Touvron, Ishan Misra, Herv{\'e} J{\'e}gou, Julien Mairal, Piotr Bojanowski, and Armand Joulin.
\newblock Emerging properties in self-supervised vision transformers.
\newblock In \emph{Proceedings of the IEEE/CVF international conference on computer vision}, pages 9650--9660, 2021.

\bibitem[Chen et~al.(2023{\natexlab{a}})Chen, Zhang, Wang, Duan, Zhou, and Zhu]{chen2023disenbooth}
Hong Chen, Yipeng Zhang, Xin Wang, Xuguang Duan, Yuwei Zhou, and Wenwu Zhu.
\newblock Disenbooth: Disentangled parameter-efficient tuning for subject-driven text-to-image generation.
\newblock \emph{arXiv preprint arXiv:2305.03374}, 2023{\natexlab{a}}.

\bibitem[Chen et~al.(2023{\natexlab{b}})Chen, Hu, Li, Rui, Jia, Chang, and Cohen]{chen2023subject}
Wenhu Chen, Hexiang Hu, Yandong Li, Nataniel Rui, Xuhui Jia, Ming-Wei Chang, and William~W Cohen.
\newblock Subject-driven text-to-image generation via apprenticeship learning.
\newblock \emph{arXiv preprint arXiv:2304.00186}, 2023{\natexlab{b}}.

\bibitem[Chen et~al.(2023{\natexlab{c}})Chen, Huang, Liu, Shen, Zhao, and Zhao]{chen2023anydoor}
Xi Chen, Lianghua Huang, Yu Liu, Yujun Shen, Deli Zhao, and Hengshuang Zhao.
\newblock Anydoor: Zero-shot object-level image customization.
\newblock \emph{arXiv preprint arXiv:2307.09481}, 2023{\natexlab{c}}.

\bibitem[Couairon et~al.(2022)Couairon, Verbeek, Schwenk, and Cord]{couairon2022diffedit}
Guillaume Couairon, Jakob Verbeek, Holger Schwenk, and Matthieu Cord.
\newblock Diffedit: Diffusion-based semantic image editing with mask guidance.
\newblock \emph{arXiv preprint arXiv:2210.11427}, 2022.

\bibitem[Esser et~al.(2023)Esser, Chiu, Atighehchian, Granskog, and Germanidis]{esser2023structure}
Patrick Esser, Johnathan Chiu, Parmida Atighehchian, Jonathan Granskog, and Anastasis Germanidis.
\newblock Structure and content-guided video synthesis with diffusion models.
\newblock In \emph{Proceedings of the IEEE/CVF International Conference on Computer Vision}, pages 7346--7356, 2023.

\bibitem[Fayyaz et~al.(2022)Fayyaz, Koohpayegani, Jafari, Sengupta, Joze, Sommerlade, Pirsiavash, and Gall]{fayyaz2022adaptive}
Mohsen Fayyaz, Soroush~Abbasi Koohpayegani, Farnoush~Rezaei Jafari, Sunando Sengupta, Hamid Reza~Vaezi Joze, Eric Sommerlade, Hamed Pirsiavash, and J{\"u}rgen Gall.
\newblock Adaptive token sampling for efficient vision transformers.
\newblock In \emph{European Conference on Computer Vision}, pages 396--414. Springer, 2022.

\bibitem[Gal et~al.(2022)Gal, Alaluf, Atzmon, Patashnik, Bermano, Chechik, and Cohen-or]{gal2022image}
Rinon Gal, Yuval Alaluf, Yuval Atzmon, Or Patashnik, Amit~Haim Bermano, Gal Chechik, and Daniel Cohen-or.
\newblock An image is worth one word: Personalizing text-to-image generation using textual inversion.
\newblock In \emph{The Eleventh International Conference on Learning Representations}, 2022.

\bibitem[Goyal et~al.(2020)Goyal, Choudhury, Raje, Chakaravarthy, Sabharwal, and Verma]{goyal2020power}
Saurabh Goyal, Anamitra~Roy Choudhury, Saurabh Raje, Venkatesan Chakaravarthy, Yogish Sabharwal, and Ashish Verma.
\newblock Power-bert: Accelerating bert inference via progressive word-vector elimination.
\newblock In \emph{International Conference on Machine Learning}, pages 3690--3699. PMLR, 2020.

\bibitem[Harvey et~al.(2022)Harvey, Naderiparizi, Masrani, Weilbach, and Wood]{harvey2022flexible}
William Harvey, Saeid Naderiparizi, Vaden Masrani, Christian Weilbach, and Frank Wood.
\newblock Flexible diffusion modeling of long videos.
\newblock \emph{Advances in Neural Information Processing Systems}, 35:\penalty0 27953--27965, 2022.

\bibitem[Hertz et~al.(2022)Hertz, Mokady, Tenenbaum, Aberman, Pritch, and Cohen-Or]{hertz2022prompt}
Amir Hertz, Ron Mokady, Jay Tenenbaum, Kfir Aberman, Yael Pritch, and Daniel Cohen-Or.
\newblock Prompt-to-prompt image editing with cross attention control.
\newblock \emph{arXiv preprint arXiv:2208.01626}, 2022.

\bibitem[Heusel et~al.(2017)Heusel, Ramsauer, Unterthiner, Nessler, and Hochreiter]{heusel2017gans}
Martin Heusel, Hubert Ramsauer, Thomas Unterthiner, Bernhard Nessler, and Sepp Hochreiter.
\newblock Gans trained by a two time-scale update rule converge to a local nash equilibrium.
\newblock \emph{Advances in neural information processing systems}, 30, 2017.

\bibitem[Ho et~al.(2022)Ho, Chan, Saharia, Whang, Gao, Gritsenko, Kingma, Poole, Norouzi, Fleet, et~al.]{ho2022imagen}
Jonathan Ho, William Chan, Chitwan Saharia, Jay Whang, Ruiqi Gao, Alexey Gritsenko, Diederik~P Kingma, Ben Poole, Mohammad Norouzi, David~J Fleet, et~al.
\newblock Imagen video: High definition video generation with diffusion models.
\newblock \emph{arXiv preprint arXiv:2210.02303}, 2022.

\bibitem[Hu et~al.(2022)Hu, Zhou, Huang, Shi, Sun, and Li]{hu2022qs}
Xueqi Hu, Xinyue Zhou, Qiusheng Huang, Zhengyi Shi, Li Sun, and Qingli Li.
\newblock Qs-attn: Query-selected attention for contrastive learning in i2i translation.
\newblock In \emph{Proceedings of the IEEE/CVF Conference on Computer Vision and Pattern Recognition}, pages 18291--18300, 2022.

\bibitem[Jeon et~al.(2021)Jeon, Min, Kim, and Sohn]{jeon2021mining}
Sangryul Jeon, Dongbo Min, Seungryong Kim, and Kwanghoon Sohn.
\newblock Mining better samples for contrastive learning of temporal correspondence.
\newblock In \emph{Proceedings of the IEEE/CVF Conference on Computer Vision and Pattern Recognition}, pages 1034--1044, 2021.

\bibitem[Kawar et~al.(2023)Kawar, Zada, Lang, Tov, Chang, Dekel, Mosseri, and Irani]{kawar2023imagic}
Bahjat Kawar, Shiran Zada, Oran Lang, Omer Tov, Huiwen Chang, Tali Dekel, Inbar Mosseri, and Michal Irani.
\newblock Imagic: Text-based real image editing with diffusion models.
\newblock In \emph{Proceedings of the IEEE/CVF Conference on Computer Vision and Pattern Recognition}, pages 6007--6017, 2023.

\bibitem[Kim et~al.(2022)Kim, Shen, Thorsley, Gholami, Kwon, Hassoun, and Keutzer]{kim2022learned}
Sehoon Kim, Sheng Shen, David Thorsley, Amir Gholami, Woosuk Kwon, Joseph Hassoun, and Kurt Keutzer.
\newblock Learned token pruning for transformers.
\newblock In \emph{Proceedings of the 28th ACM SIGKDD Conference on Knowledge Discovery and Data Mining}, pages 784--794, 2022.

\bibitem[Kumari et~al.(2023)Kumari, Zhang, Zhang, Shechtman, and Zhu]{kumari2023multi}
Nupur Kumari, Bingliang Zhang, Richard Zhang, Eli Shechtman, and Jun-Yan Zhu.
\newblock Multi-concept customization of text-to-image diffusion.
\newblock In \emph{Proceedings of the IEEE/CVF Conference on Computer Vision and Pattern Recognition}, pages 1931--1941, 2023.

\bibitem[Kuznetsova et~al.(2020{\natexlab{a}})Kuznetsova, Rom, Alldrin, Uijlings, Krasin, Pont-Tuset, Kamali, Popov, Malloci, Kolesnikov, Duerig, and Ferrari]{OpenImages}
Alina Kuznetsova, Hassan Rom, Neil Alldrin, Jasper Uijlings, Ivan Krasin, Jordi Pont-Tuset, Shahab Kamali, Stefan Popov, Matteo Malloci, Alexander Kolesnikov, Tom Duerig, and Vittorio Ferrari.
\newblock The open images dataset v4: Unified image classification, object detection, and visual relationship detection at scale.
\newblock \emph{IJCV}, 2020{\natexlab{a}}.

\bibitem[Kuznetsova et~al.(2020{\natexlab{b}})Kuznetsova, Rom, Alldrin, Uijlings, Krasin, Pont-Tuset, Kamali, Popov, Malloci, Kolesnikov, et~al.]{kuznetsova2020open}
Alina Kuznetsova, Hassan Rom, Neil Alldrin, Jasper Uijlings, Ivan Krasin, Jordi Pont-Tuset, Shahab Kamali, Stefan Popov, Matteo Malloci, Alexander Kolesnikov, et~al.
\newblock The open images dataset v4: Unified image classification, object detection, and visual relationship detection at scale.
\newblock \emph{International Journal of Computer Vision}, 128\penalty0 (7):\penalty0 1956--1981, 2020{\natexlab{b}}.

\bibitem[Li et~al.(2022)Li, Li, Xiong, and Hoi]{li2022blip}
Junnan Li, Dongxu Li, Caiming Xiong, and Steven Hoi.
\newblock Blip: Bootstrapping language-image pre-training for unified vision-language understanding and generation.
\newblock In \emph{International Conference on Machine Learning}, pages 12888--12900. PMLR, 2022.

\bibitem[Lin et~al.(2014)Lin, Maire, Belongie, Hays, Perona, Ramanan, Doll{\'a}r, and Zitnick]{lin2014microsoft}
Tsung-Yi Lin, Michael Maire, Serge Belongie, James Hays, Pietro Perona, Deva Ramanan, Piotr Doll{\'a}r, and C~Lawrence Zitnick.
\newblock Microsoft coco: Common objects in context.
\newblock In \emph{Computer Vision--ECCV 2014: 13th European Conference, Zurich, Switzerland, September 6-12, 2014, Proceedings, Part V 13}, pages 740--755. Springer, 2014.

\bibitem[Liu et~al.(2022)Liu, Xiong, Xu, Cao, and Jin]{liu2022ts2}
Yuqi Liu, Pengfei Xiong, Luhui Xu, Shengming Cao, and Qin Jin.
\newblock Ts2-net: Token shift and selection transformer for text-video retrieval.
\newblock In \emph{European Conference on Computer Vision}, pages 319--335. Springer, 2022.

\bibitem[Lugmayr et~al.(2022)Lugmayr, Danelljan, Romero, Yu, Timofte, and Van~Gool]{lugmayr2022repaint}
Andreas Lugmayr, Martin Danelljan, Andres Romero, Fisher Yu, Radu Timofte, and Luc Van~Gool.
\newblock Repaint: Inpainting using denoising diffusion probabilistic models.
\newblock In \emph{Proceedings of the IEEE/CVF Conference on Computer Vision and Pattern Recognition}, pages 11461--11471, 2022.

\bibitem[Luo et~al.(2023)Luo, Chen, Zhang, Huang, Wang, Shen, Zhao, Zhou, and Tan]{luo2023videofusion}
Zhengxiong Luo, Dayou Chen, Yingya Zhang, Yan Huang, Liang Wang, Yujun Shen, Deli Zhao, Jingren Zhou, and Tieniu Tan.
\newblock Videofusion: Decomposed diffusion models for high-quality video generation.
\newblock In \emph{Proceedings of the IEEE/CVF Conference on Computer Vision and Pattern Recognition}, pages 10209--10218, 2023.

\bibitem[Mei and Patel(2023)]{mei2023vidm}
Kangfu Mei and Vishal Patel.
\newblock Vidm: Video implicit diffusion models.
\newblock In \emph{Proceedings of the AAAI Conference on Artificial Intelligence}, pages 9117--9125, 2023.

\bibitem[Mokady et~al.(2023)Mokady, Hertz, Aberman, Pritch, and Cohen-Or]{mokady2023null}
Ron Mokady, Amir Hertz, Kfir Aberman, Yael Pritch, and Daniel Cohen-Or.
\newblock Null-text inversion for editing real images using guided diffusion models.
\newblock In \emph{Proceedings of the IEEE/CVF Conference on Computer Vision and Pattern Recognition}, pages 6038--6047, 2023.

\bibitem[Molad et~al.(2023)Molad, Horwitz, Valevski, Acha, Matias, Pritch, Leviathan, and Hoshen]{molad2023dreamix}
Eyal Molad, Eliahu Horwitz, Dani Valevski, Alex~Rav Acha, Yossi Matias, Yael Pritch, Yaniv Leviathan, and Yedid Hoshen.
\newblock Dreamix: Video diffusion models are general video editors.
\newblock \emph{arXiv preprint arXiv:2302.01329}, 2023.

\bibitem[Nichol et~al.(2021)Nichol, Dhariwal, Ramesh, Shyam, Mishkin, McGrew, Sutskever, and Chen]{nichol2021glide}
Alex Nichol, Prafulla Dhariwal, Aditya Ramesh, Pranav Shyam, Pamela Mishkin, Bob McGrew, Ilya Sutskever, and Mark Chen.
\newblock Glide: Towards photorealistic image generation and editing with text-guided diffusion models.
\newblock \emph{arXiv preprint arXiv:2112.10741}, 2021.

\bibitem[Podell et~al.(2023)Podell, English, Lacey, Blattmann, Dockhorn, M{\"u}ller, Penna, and Rombach]{podell2023sdxl}
Dustin Podell, Zion English, Kyle Lacey, Andreas Blattmann, Tim Dockhorn, Jonas M{\"u}ller, Joe Penna, and Robin Rombach.
\newblock Sdxl: improving latent diffusion models for high-resolution image synthesis.
\newblock \emph{arXiv preprint arXiv:2307.01952}, 2023.

\bibitem[Poole et~al.(2023)Poole, Jain, Barron, and Mildenhall]{poole2022dreamfusion}
Ben Poole, Ajay Jain, Jonathan~T Barron, and Ben Mildenhall.
\newblock Dreamfusion: Text-to-3d using 2d diffusion.
\newblock In \emph{The Eleventh International Conference on Learning Representations}, 2023.

\bibitem[Radford et~al.(2021)Radford, Kim, Hallacy, Ramesh, Goh, Agarwal, Sastry, Askell, Mishkin, Clark, et~al.]{radford2021learning}
Alec Radford, Jong~Wook Kim, Chris Hallacy, Aditya Ramesh, Gabriel Goh, Sandhini Agarwal, Girish Sastry, Amanda Askell, Pamela Mishkin, Jack Clark, et~al.
\newblock Learning transferable visual models from natural language supervision.
\newblock In \emph{International conference on machine learning}, pages 8748--8763. PMLR, 2021.

\bibitem[Ramesh et~al.(2021)Ramesh, Pavlov, Goh, Gray, Voss, Radford, Chen, and Sutskever]{ramesh2021zero}
Aditya Ramesh, Mikhail Pavlov, Gabriel Goh, Scott Gray, Chelsea Voss, Alec Radford, Mark Chen, and Ilya Sutskever.
\newblock Zero-shot text-to-image generation.
\newblock In \emph{International Conference on Machine Learning}, pages 8821--8831. PMLR, 2021.

\bibitem[Ramesh et~al.(2022)Ramesh, Dhariwal, Nichol, Chu, and Chen]{ramesh2022hierarchical}
Aditya Ramesh, Prafulla Dhariwal, Alex Nichol, Casey Chu, and Mark Chen.
\newblock Hierarchical text-conditional image generation with clip latents.
\newblock \emph{arXiv preprint arXiv:2204.06125}, 1\penalty0 (2):\penalty0 3, 2022.

\bibitem[Rombach et~al.(2022)Rombach, Blattmann, Lorenz, Esser, and Ommer]{rombach2022high}
Robin Rombach, Andreas Blattmann, Dominik Lorenz, Patrick Esser, and Bj{\"o}rn Ommer.
\newblock High-resolution image synthesis with latent diffusion models.
\newblock In \emph{Proceedings of the IEEE/CVF conference on computer vision and pattern recognition}, pages 10684--10695, 2022.

\bibitem[Ruiz et~al.(2023)Ruiz, Li, Jampani, Pritch, Rubinstein, and Aberman]{ruiz2023dreambooth}
Nataniel Ruiz, Yuanzhen Li, Varun Jampani, Yael Pritch, Michael Rubinstein, and Kfir Aberman.
\newblock Dreambooth: Fine tuning text-to-image diffusion models for subject-driven generation.
\newblock In \emph{Proceedings of the IEEE/CVF Conference on Computer Vision and Pattern Recognition}, pages 22500--22510, 2023.

\bibitem[Saharia et~al.(2022)Saharia, Chan, Saxena, Li, Whang, Denton, Ghasemipour, Gontijo~Lopes, Karagol~Ayan, Salimans, et~al.]{saharia2022photorealistic}
Chitwan Saharia, William Chan, Saurabh Saxena, Lala Li, Jay Whang, Emily~L Denton, Kamyar Ghasemipour, Raphael Gontijo~Lopes, Burcu Karagol~Ayan, Tim Salimans, et~al.
\newblock Photorealistic text-to-image diffusion models with deep language understanding.
\newblock \emph{Advances in Neural Information Processing Systems}, 35:\penalty0 36479--36494, 2022.

\bibitem[Song et~al.(2020)Song, Meng, and Ermon]{song2020denoising}
Jiaming Song, Chenlin Meng, and Stefano Ermon.
\newblock Denoising diffusion implicit models.
\newblock \emph{arXiv preprint arXiv:2010.02502}, 2020.

\bibitem[Song et~al.(2023)Song, Zhang, Lin, Cohen, Price, Zhang, Kim, and Aliaga]{song2023objectstitch}
Yizhi Song, Zhifei Zhang, Zhe Lin, Scott Cohen, Brian Price, Jianming Zhang, Soo~Ye Kim, and Daniel Aliaga.
\newblock Objectstitch: Object compositing with diffusion model.
\newblock In \emph{Proceedings of the IEEE/CVF Conference on Computer Vision and Pattern Recognition}, pages 18310--18319, 2023.

\bibitem[Tumanyan et~al.(2023)Tumanyan, Geyer, Bagon, and Dekel]{tumanyan2023plug}
Narek Tumanyan, Michal Geyer, Shai Bagon, and Tali Dekel.
\newblock Plug-and-play diffusion features for text-driven image-to-image translation.
\newblock In \emph{Proceedings of the IEEE/CVF Conference on Computer Vision and Pattern Recognition}, pages 1921--1930, 2023.

\bibitem[Wang et~al.(2022)Wang, Yang, Li, Liu, Wu, and Jiang]{wang2022efficient}
Junke Wang, Xitong Yang, Hengduo Li, Li Liu, Zuxuan Wu, and Yu-Gang Jiang.
\newblock Efficient video transformers with spatial-temporal token selection.
\newblock In \emph{European Conference on Computer Vision}, pages 69--86. Springer, 2022.

\bibitem[Wei et~al.(2023)Wei, Zhang, Ji, Bai, Zhang, and Zuo]{wei2023elite}
Yuxiang Wei, Yabo Zhang, Zhilong Ji, Jinfeng Bai, Lei Zhang, and Wangmeng Zuo.
\newblock Elite: Encoding visual concepts into textual embeddings for customized text-to-image generation.
\newblock \emph{arXiv preprint arXiv:2302.13848}, 2023.

\bibitem[Wu et~al.(2023)Wu, Ge, Wang, Lei, Gu, Shi, Hsu, Shan, Qie, and Shou]{wu2023tune}
Jay~Zhangjie Wu, Yixiao Ge, Xintao Wang, Stan~Weixian Lei, Yuchao Gu, Yufei Shi, Wynne Hsu, Ying Shan, Xiaohu Qie, and Mike~Zheng Shou.
\newblock Tune-a-video: One-shot tuning of image diffusion models for text-to-video generation.
\newblock In \emph{Proceedings of the IEEE/CVF International Conference on Computer Vision}, pages 7623--7633, 2023.

\bibitem[Xie et~al.(2023)Xie, Zhang, Lin, Hinz, and Zhang]{xie2023smartbrush}
Shaoan Xie, Zhifei Zhang, Zhe Lin, Tobias Hinz, and Kun Zhang.
\newblock Smartbrush: Text and shape guided object inpainting with diffusion model.
\newblock In \emph{Proceedings of the IEEE/CVF Conference on Computer Vision and Pattern Recognition}, pages 22428--22437, 2023.

\bibitem[Yang et~al.(2023)Yang, Gu, Zhang, Zhang, Chen, Sun, Chen, and Wen]{yang2023paint}
Binxin Yang, Shuyang Gu, Bo Zhang, Ting Zhang, Xuejin Chen, Xiaoyan Sun, Dong Chen, and Fang Wen.
\newblock Paint by example: Exemplar-based image editing with diffusion models.
\newblock In \emph{Proceedings of the IEEE/CVF Conference on Computer Vision and Pattern Recognition}, pages 18381--18391, 2023.

\bibitem[Yildirim et~al.(2023)Yildirim, Baday, Erdem, Erdem, and Dundar]{yildirim2023inst}
Ahmet~Burak Yildirim, Vedat Baday, Erkut Erdem, Aykut Erdem, and Aysegul Dundar.
\newblock Inst-inpaint: Instructing to remove objects with diffusion models.
\newblock \emph{arXiv preprint arXiv:2304.03246}, 2023.

\bibitem[Yu et~al.(2023)Yu, Xu, Zhang, Liu, Ye, Wu, Yan, Zhu, Xiong, Liang, et~al.]{yu2023mvimgnet}
Xianggang Yu, Mutian Xu, Yidan Zhang, Haolin Liu, Chongjie Ye, Yushuang Wu, Zizheng Yan, Chenming Zhu, Zhangyang Xiong, Tianyou Liang, et~al.
\newblock Mvimgnet: A large-scale dataset of multi-view images.
\newblock In \emph{Proceedings of the IEEE/CVF Conference on Computer Vision and Pattern Recognition}, pages 9150--9161, 2023.

\bibitem[Zhang et~al.(2023)Zhang, Rao, and Agrawala]{zhang2023adding}
Lvmin Zhang, Anyi Rao, and Maneesh Agrawala.
\newblock Adding conditional control to text-to-image diffusion models.
\newblock In \emph{Proceedings of the IEEE/CVF International Conference on Computer Vision}, pages 3836--3847, 2023.

\bibitem[Zhou et~al.(2022)Zhou, Hou, Yang, Jin, and Feng]{zhou2022token}
Daquan Zhou, Qibin Hou, Linjie Yang, Xiaojie Jin, and Jiashi Feng.
\newblock Token selection is a simple booster for vision transformers.
\newblock \emph{IEEE Transactions on Pattern Analysis and Machine Intelligence}, 2022.

\end{thebibliography}
}

\clearpage
\setcounter{page}{1}
\maketitlesupplementary

\section*{Overview of the Supplementary}
In Section \ref{sec:supp_impl}, we delve into the implementation details, while Section \ref{sec:supp_text} offers additional text-guided samples. Section \ref{sec:supp_image} showcases image-guided samples. Following that, Section \ref{sec:supp_ablation} presents a qualitative ablation study on both the proposed token selection module and the decoupling regularization. To conclude, Section \ref{sec:supp_limit} highlights further examples of our limitations and initiates a discussion on future work.

\section{Implementation}
\label{sec:supp_impl}
Within this section, we elucidate the implementation specifics of our approach. The Unet component within the stable diffusion architecture is comprised of 4 downsample blocks, 1 middle block, and 4 upsampling blocks. The Unet operates on input data with dimensions $\mathbb{R}^{4\times 64 \times 64}$. Our selection of the second downsample layer is deliberate, as it encapsulates substantial information pertaining to the object, while concurrently mitigating the inclusion of extraneous details, such as the overarching structural characteristics of the object. This choice aligns with the inherent necessity of accommodating shape alterations for the purpose of inpainting.

The shape of the second block is $\mathbb{R}^{C\times H\times W}$, and we reshape it into $\mathbb{R}^{HW \times C}$, where $C=1280, H=32, W=32$ (please note that there is a typo in the main paper, where 768 should be corrected to 1280). Consequently, we have 1280 tokens, each with a dimension of 1280. It's important to highlight that employing only 1024 tokens would lead the model to effectively replicate the object but struggle to generate realistic images, as illustrated in the main paper. Hence, we incorporate our proposed token selection module to identify the most discriminative tokens. As previously mentioned in the main paper, we opt for 24 tokens following the selection process.
Post-token selection, we initially feed the chosen tokens, denoted as $\mathbb{R}^{24\times 1280}$, through a series of five transformer blocks to enhance their representational capacity, drawing inspiration from PBE \citep{yang2023paint}. Subsequently, we integrate these transformed features into the Unet by introducing new Cross-Attention layers after the feedforward layers within each attention block in both the middle and upsampling blocks. Notably, these newly introduced parameters are trained alongside the pre-trained stable diffusion parameters.
Since the downsampling block necessitates text prompt input for the reference image, we consistently employ the same text prompt for inpainting. For instance, if \textit{dog} is used as the text prompt, we extract features of the reference dog and subsequently employ \textit{dog} for recovering the pristine version of the noisy dog.
Then we finetune the model on the OpenImages \citep{OpenImages}, MvImageNet \citep{yu2023mvimgnet} and COCO images \citep{lin2014microsoft}with ratios 60\%, 20\%,20\%. For the openimages dataset, we add noise to the segmentation region and use the cropped object as reference image. For the MvimageNet, we apply image segmentation model to the dataset and obtain segmentation mask and text class label for each segmentation. Then we add noise to the segmentation region and uses the object in different view as reference image. We find empirically that using MvImageNet helps alleviate but still fail to address the copy-paste issue when we use redundant Unet features as demonstrated in the main paper. As for the COCO images, we only use it for the proposed decoupling regularization as it contains high-quality image captions. As shown in the main paper, we add noises to the whole image and recover the clean version with the image caption and same cropped object.

During the training of the model, we employ the Adam optimizer with a learning rate of $10^{-5}$ and a batch size of 512. The training process encompasses 40,000 steps. For evaluation purposes, we utilize checkpoint models with exponential moving averages.
In the context of subject-driven image inpainting, we typically employ simple text labels as text inputs for reference feature extraction and image denoising. For example, we use labels such as "beer," "candle," and "sloth plushie" to guide the inpainting process.
On the other hand, for text-driven subject inpainting, we extract the reference features using text labels but utilize more specific text prompts for inpainting. For instance, we might use "dog" to extract the reference feature but generate an image of a "dog wearing sunglasses" using that reference.
It's noteworthy that during the inpainting process, we introduce noise exclusively to the masked region and employ DDIM \citep{song2020denoising} for image sampling. In each step, we replace the background with a clean background, following the methodology presented in \citep{xie2023smartbrush}. This is done to ensure that the background remains consistent and is not altered during the inpainting process.

\section{Text-guided Subject Driven Image Inpainting}
\label{sec:supp_text}
\begin{figure*}
    \centering
    \setlength{\tabcolsep}{2pt}
    \scalebox{0.85}{
    \begin{tabular}{cc|ccc}
    \toprule

         Input & Reference Subject & \multicolumn{3}{c}{Text-Guided Subject-Driven Inpainting} \\ \midrule

         & & \textit{corgi} & \textit{corgi laughing} & \textit{corgi wearing bowtie} \\ 
         \includegraphics[scale=0.19]{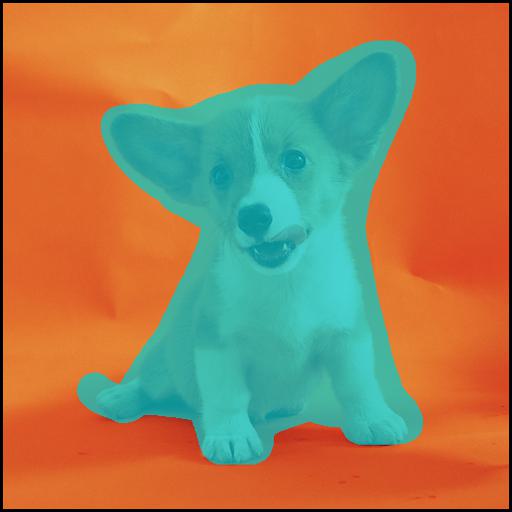}&
          \includegraphics[scale=0.19]{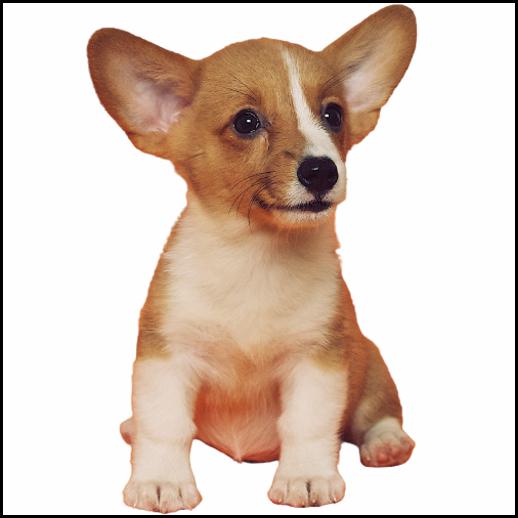}&
           \includegraphics[scale=0.19]{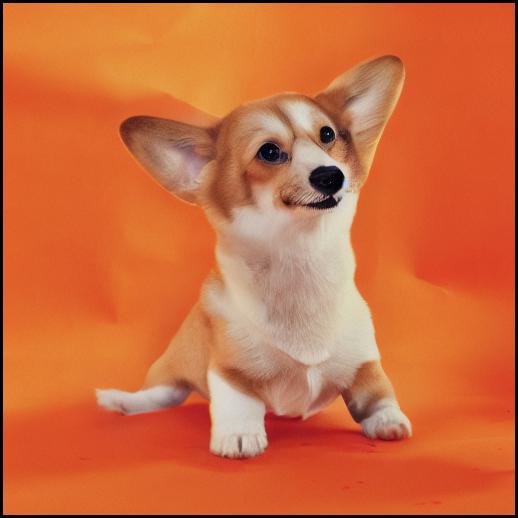}&
         \includegraphics[scale=0.19]{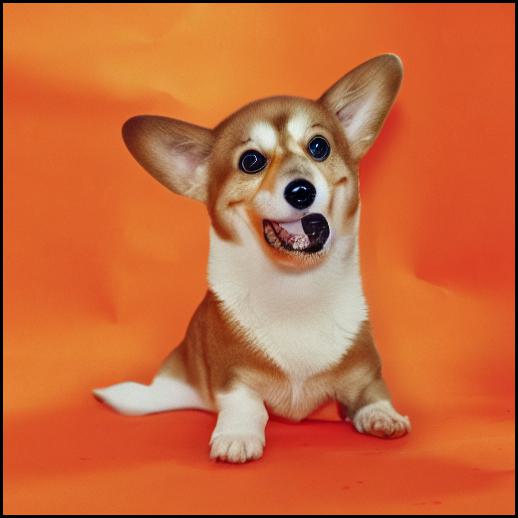} &
         \includegraphics[scale=0.19]{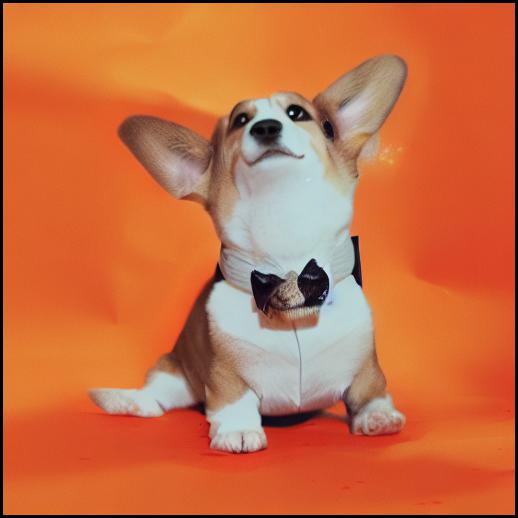}
         \\
           & & \textit{dog} & \textit{dog wearing sunglasses} & \textit{dog wearing hat} \\ 
           \includegraphics[scale=0.19]{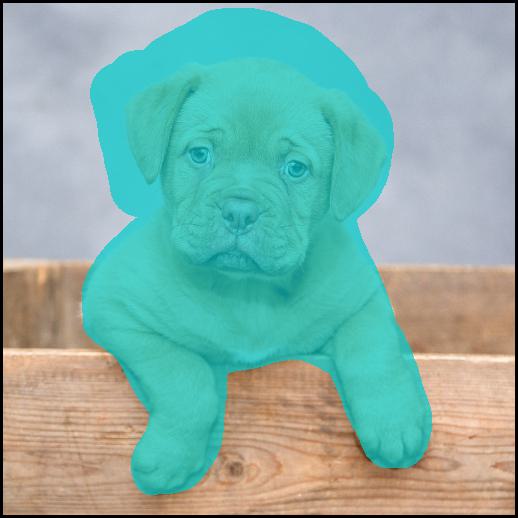}&
          \includegraphics[scale=0.19]{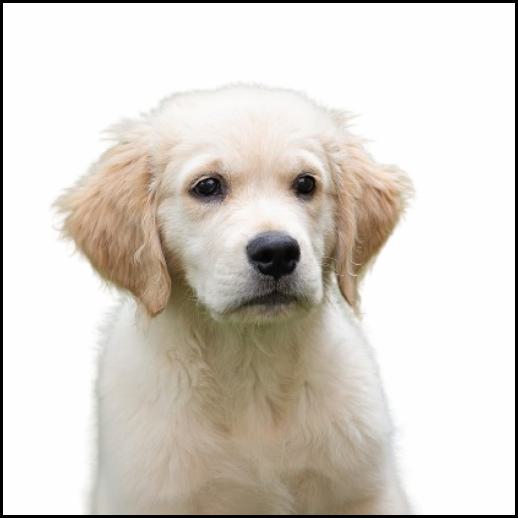}&
           \includegraphics[scale=0.19]{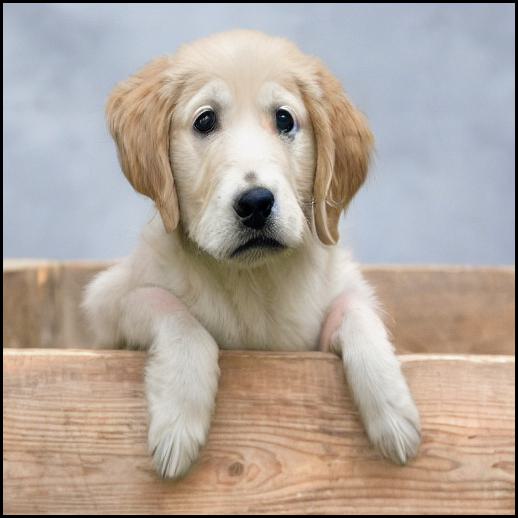}&
         \includegraphics[scale=0.19]{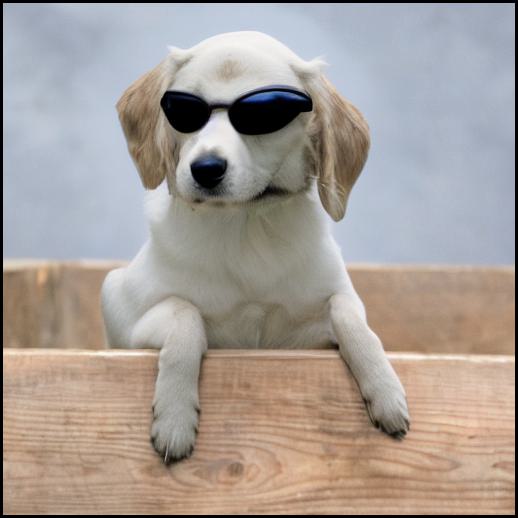} &
         \includegraphics[scale=0.19]{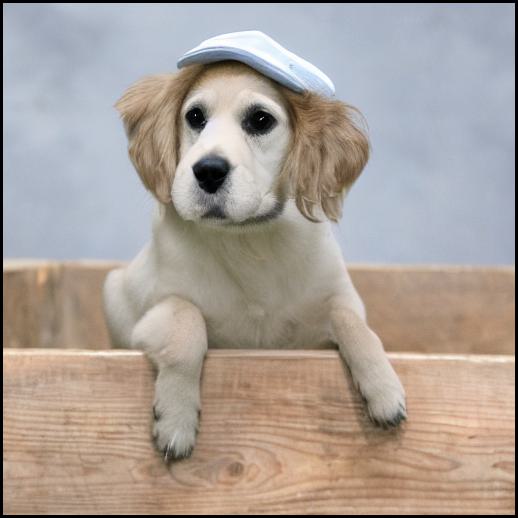}
         \\ 
          & & \textit{cat} & \textit{open-mouthed cat} & \textit{cat wearing blue scarf} \\ 
           \includegraphics[scale=0.19]{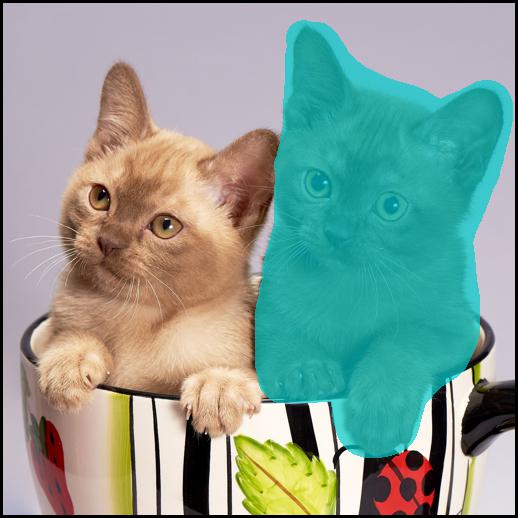}&
          \includegraphics[scale=0.19]{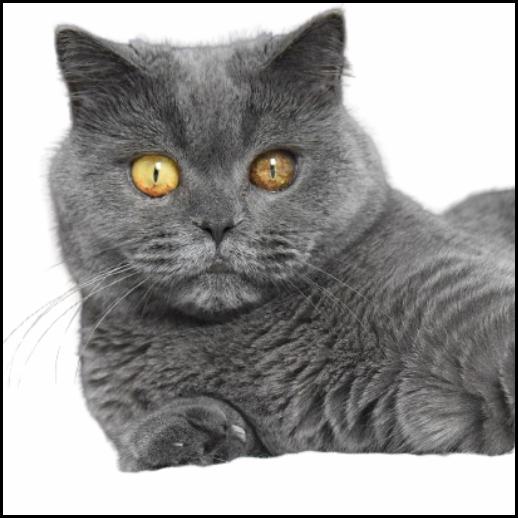}&
           \includegraphics[scale=0.19]{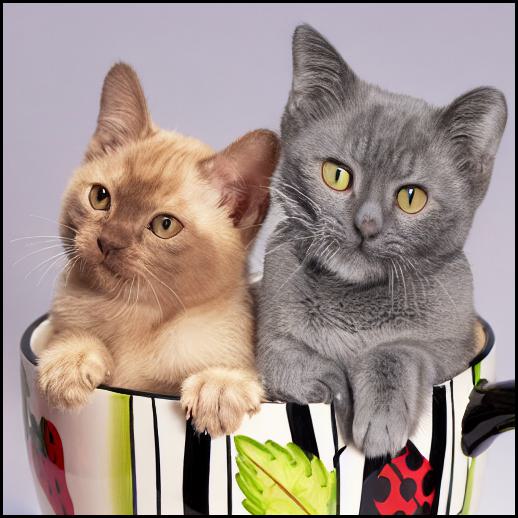}&
         \includegraphics[scale=0.19]{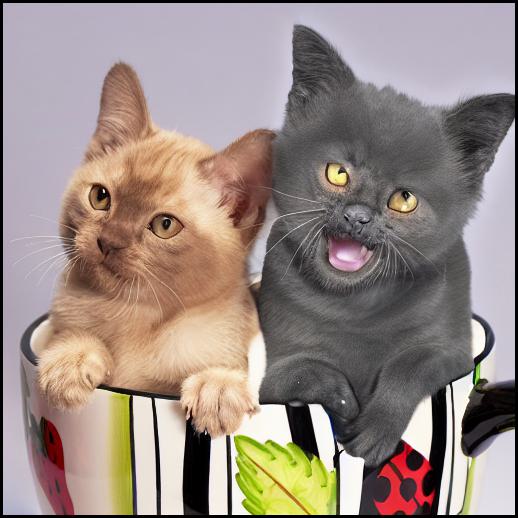} &
         \includegraphics[scale=0.19]{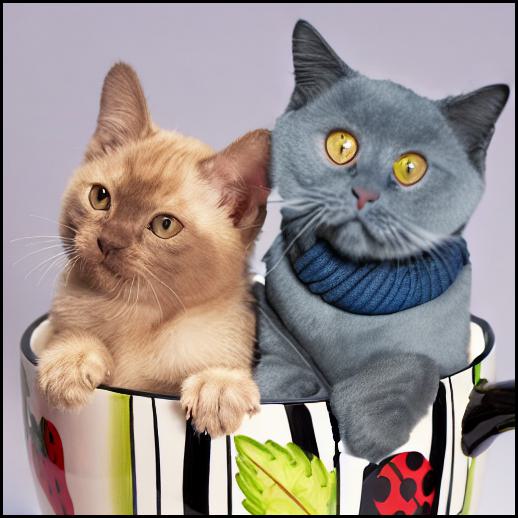}
         \\ 
          & & \textit{watermelon} & \textit{cube watermelon} & \textit{pumpkin-shape watermelon} \\ 
           \includegraphics[scale=0.19]{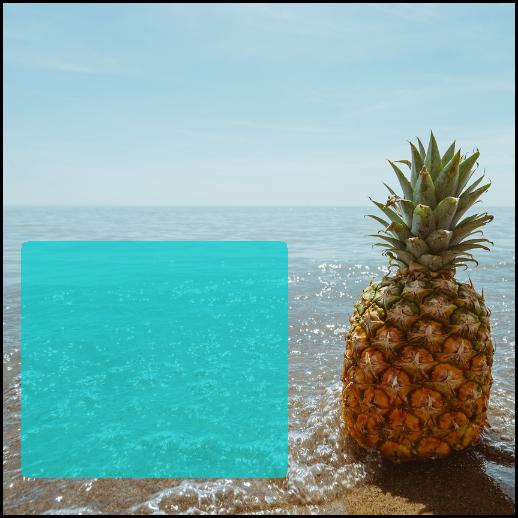}&
          \includegraphics[scale=0.19]{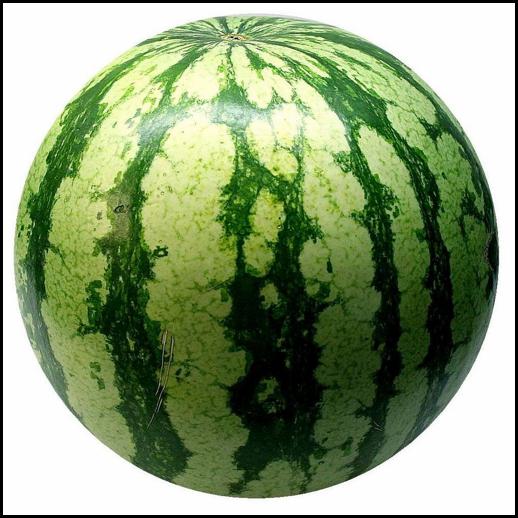}&
           \includegraphics[scale=0.19]{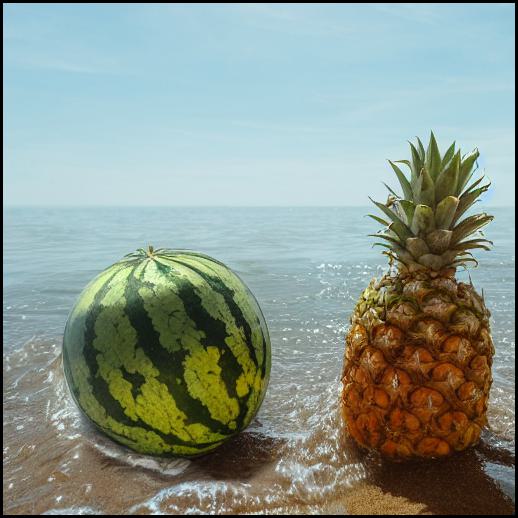}&
         \includegraphics[scale=0.19]{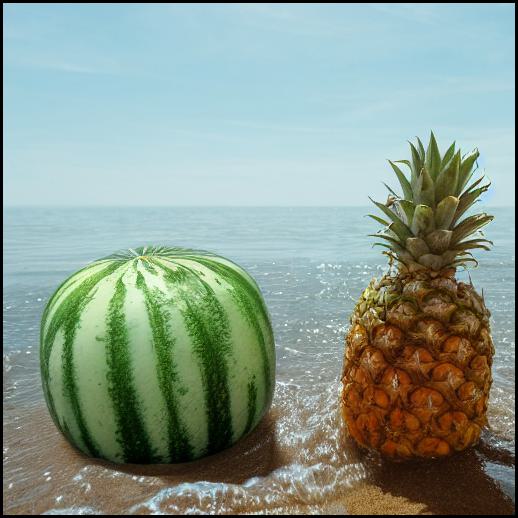} &
         \includegraphics[scale=0.19]{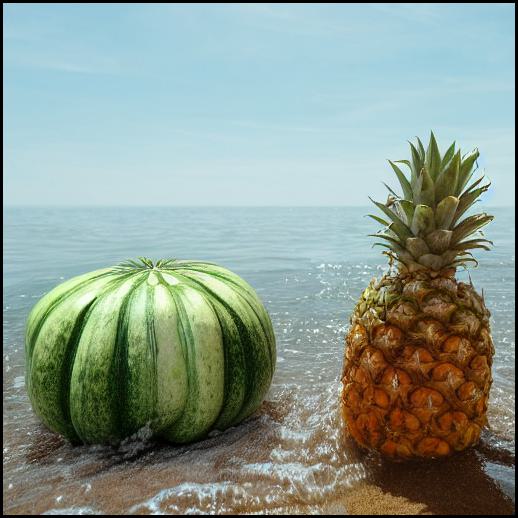}
         \\ 
           & & \textit{ballon} & \textit{heart-shaped ballon} & \textit{cube ballon} \\ 
           \includegraphics[scale=0.19]{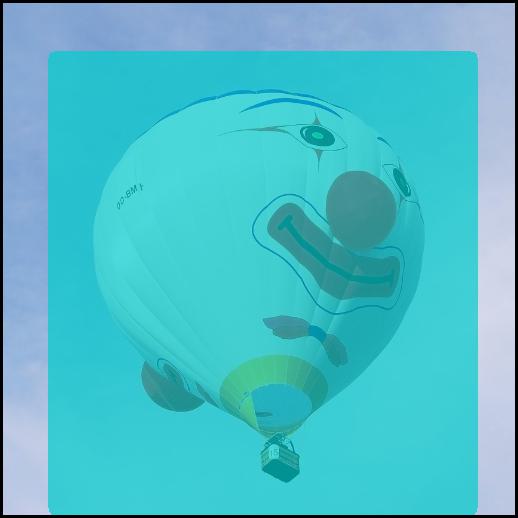}&
          \includegraphics[scale=0.19]{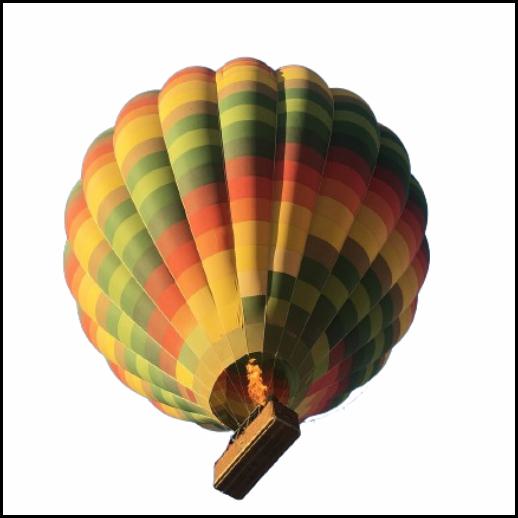}&
           \includegraphics[scale=0.19]{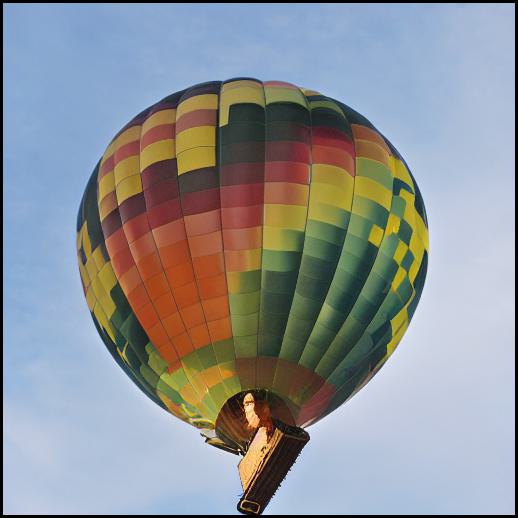}&
         \includegraphics[scale=0.19]{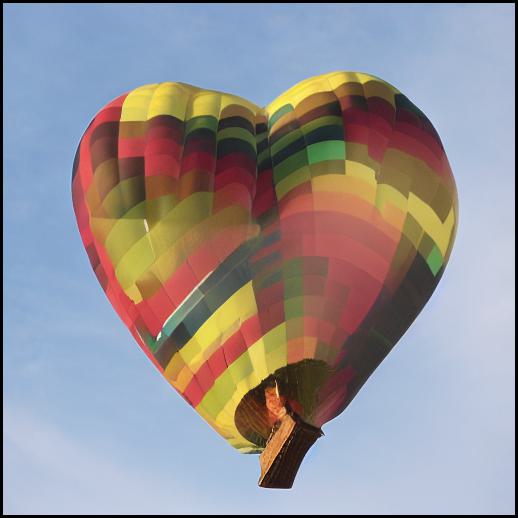} &
         \includegraphics[scale=0.19]{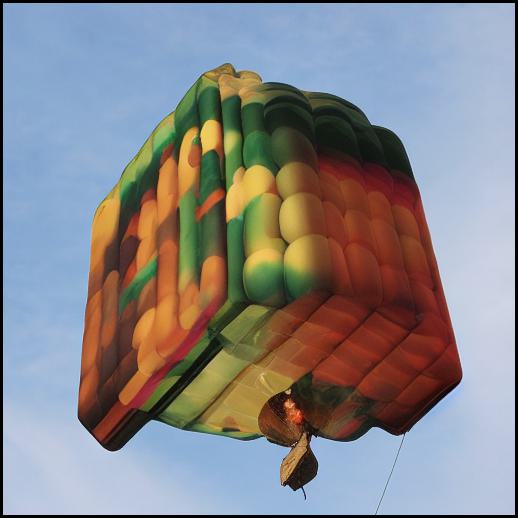}
         \\ 
         
            & & \textit{shoe} & \textit{boots} & \textit{vans} \\ 
           \includegraphics[scale=0.19]{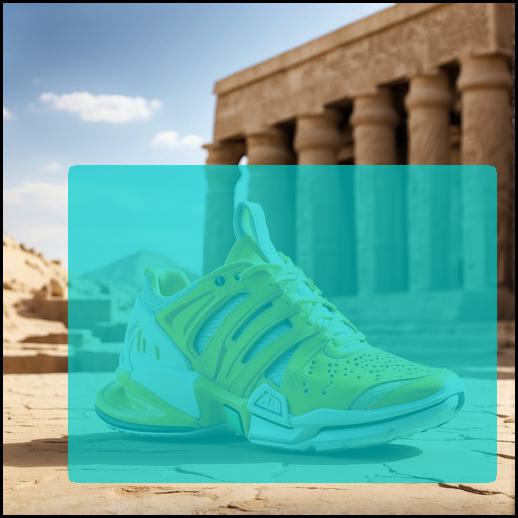}&
          \includegraphics[scale=0.19]{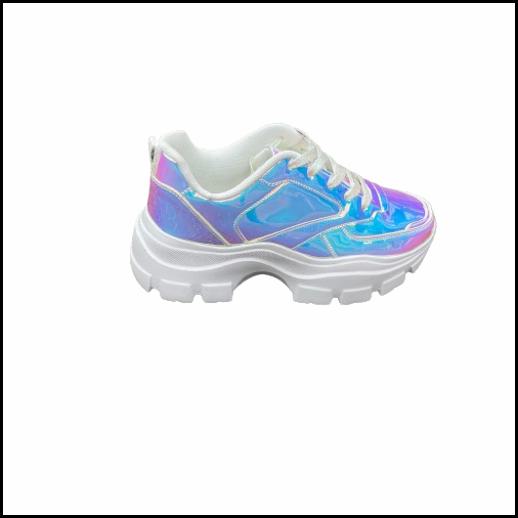}&
           \includegraphics[scale=0.19]{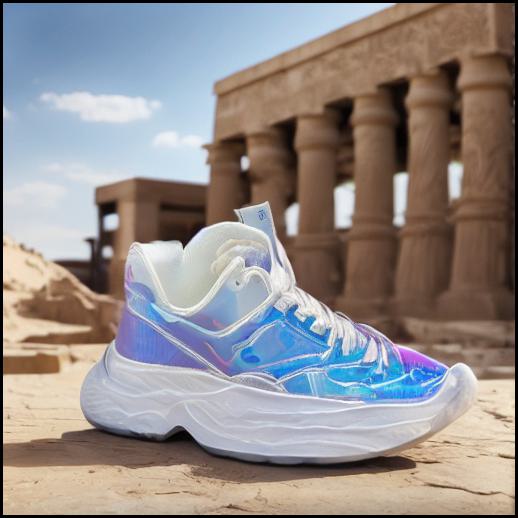}&
         \includegraphics[scale=0.19]{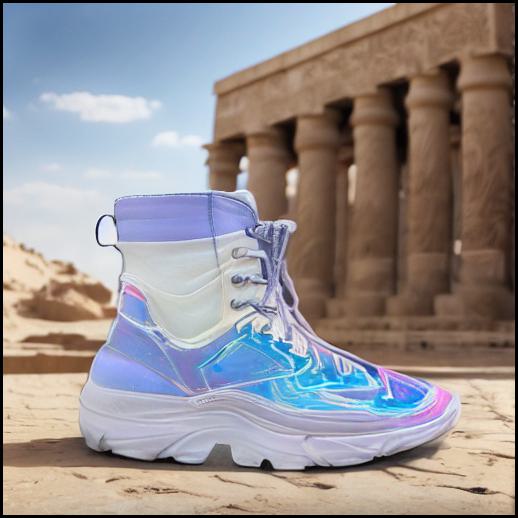} &
         \includegraphics[scale=0.19]{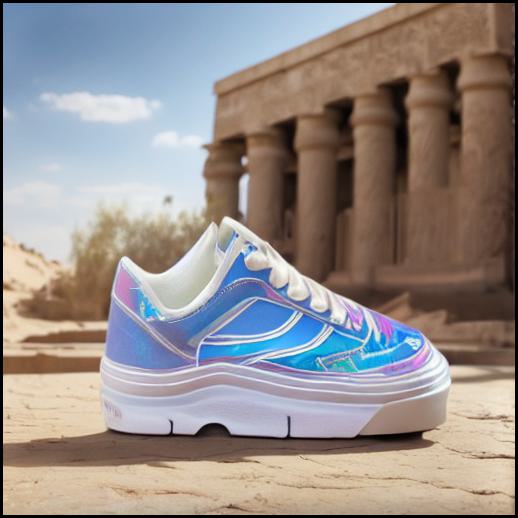}
         \\ 
          
         \bottomrule
    \end{tabular}
    }
    \vspace{-2mm}
    \caption{Text-guided subject-driven image inpainting results. Note that the strong inpainting baselines~\citep{rombach2022high,yang2023paint} does not support both text and image guidance.}
    \label{fig:supp_text}
\end{figure*}

\begin{figure*}
    \centering
    \setlength{\tabcolsep}{2pt}
    \scalebox{0.85}{
    \begin{tabular}{cc|ccc}
    \toprule

         Input & Reference Subject & \multicolumn{3}{c}{Text-Guided Subject-Driven Inpainting} \\ \midrule

         & & \textit{dog} & \textit{lion} & \textit{dinosour} \\ 
         \includegraphics[scale=0.19]{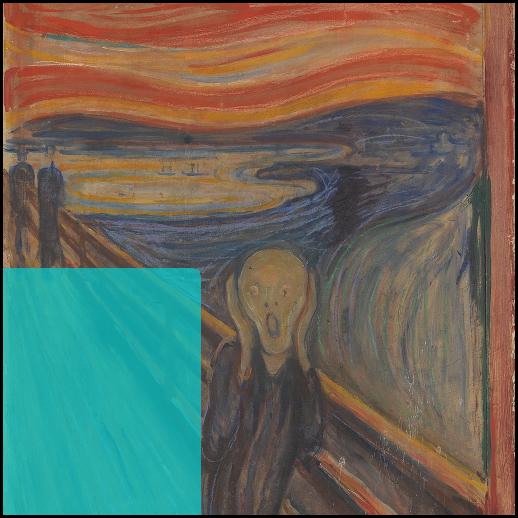}&
          \includegraphics[scale=0.19]{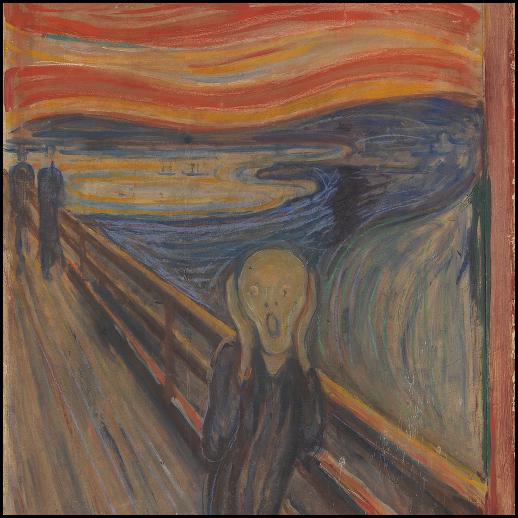}&
           \includegraphics[scale=0.19]{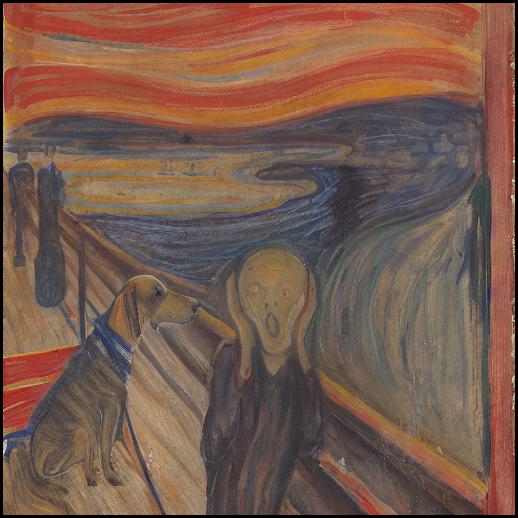}&
         \includegraphics[scale=0.19]{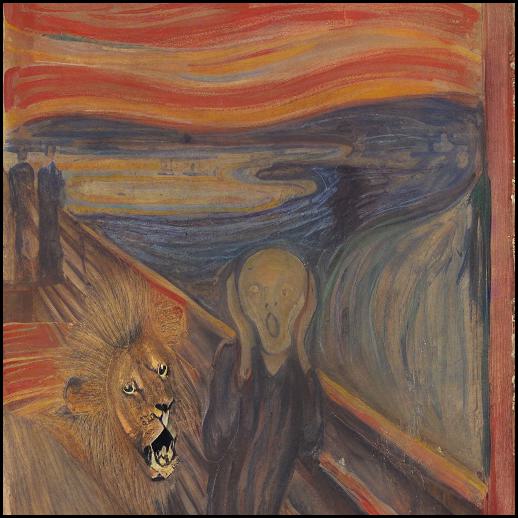} &
         \includegraphics[scale=0.19]{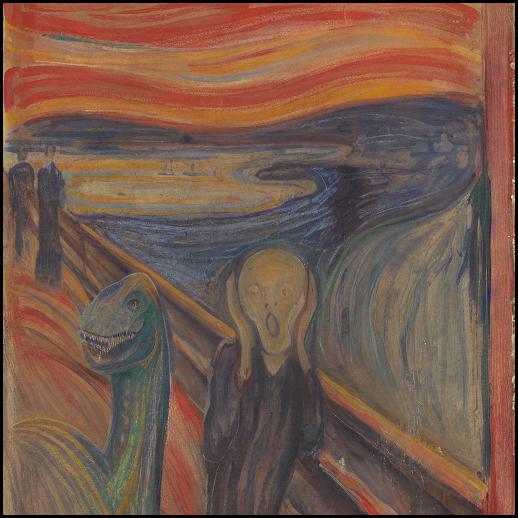}\\
           & & \textit{teddy bear} & \textit{coffee mug} & \textit{pikachu} \\ 
         \includegraphics[scale=0.19]{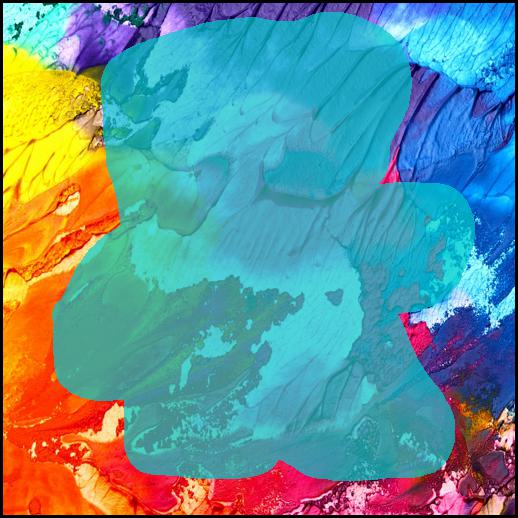}&
          \includegraphics[scale=0.19]{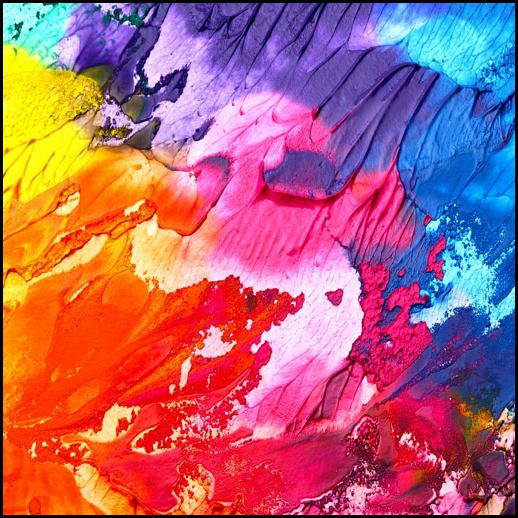}&
           \includegraphics[scale=0.19]{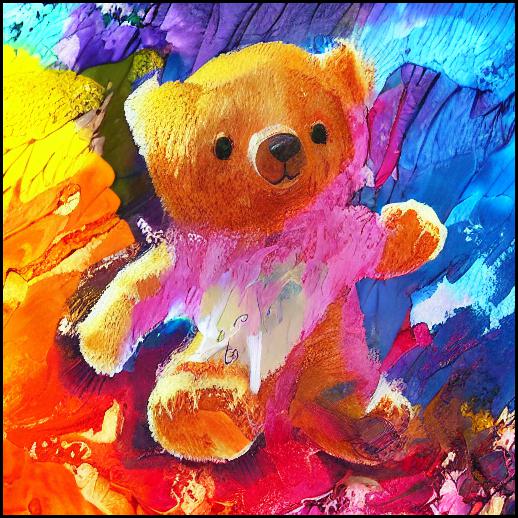}&
         \includegraphics[scale=0.19]{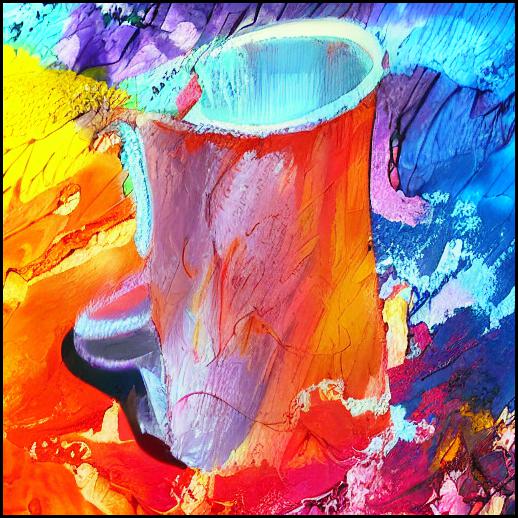} &
         \includegraphics[scale=0.19]{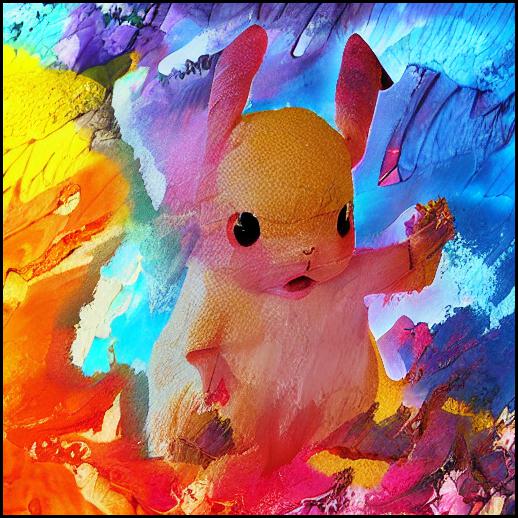}\\
           & & \textit{chair} & \textit{table} & \textit{microwave} \\ 
         \includegraphics[scale=0.19]{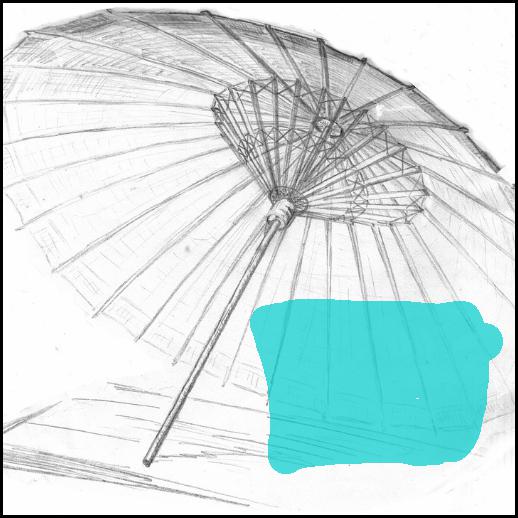}&
          \includegraphics[scale=0.19]{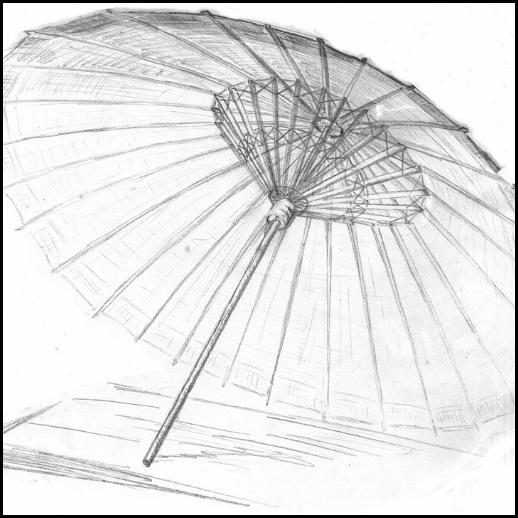}&
           \includegraphics[scale=0.19]{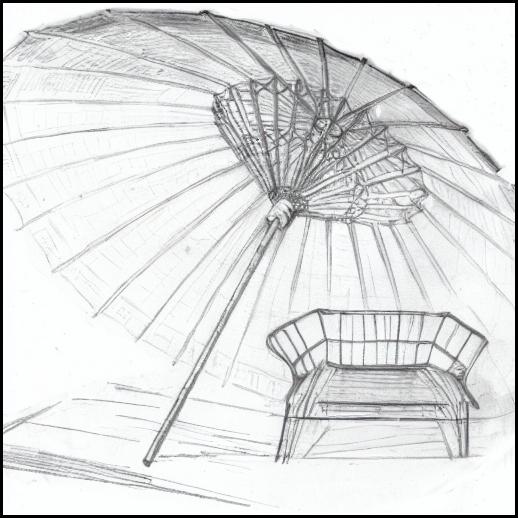}&
         \includegraphics[scale=0.19]{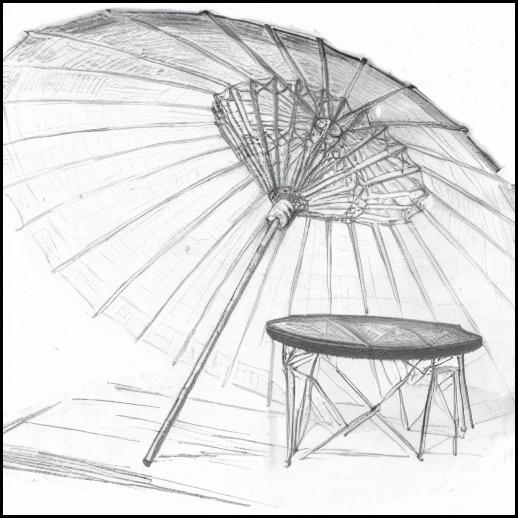} &
         \includegraphics[scale=0.19]{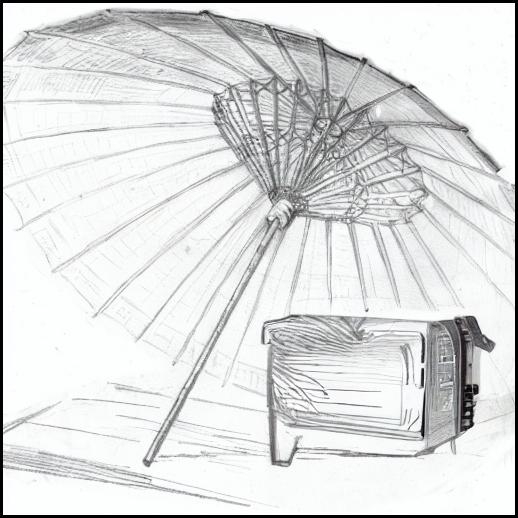}\\
           & & \textit{mushroom} & \textit{umbrella} & \textit{tree} \\ 
         \includegraphics[scale=0.19]{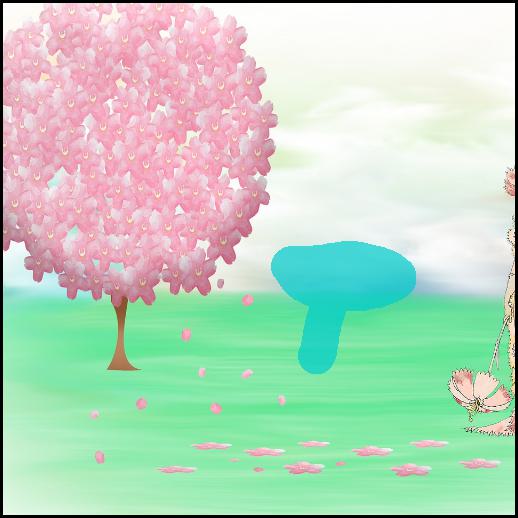}&
          \includegraphics[scale=0.19]{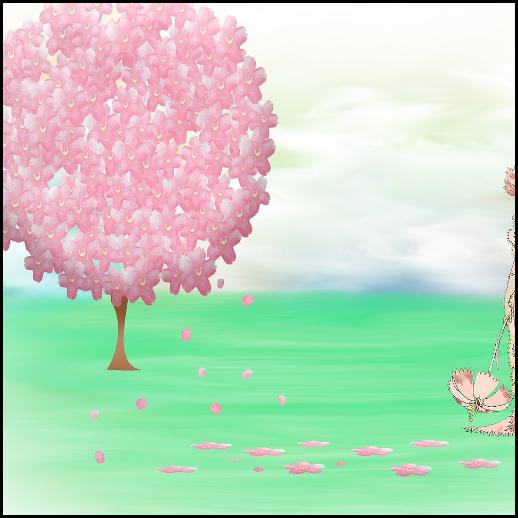}&
           \includegraphics[scale=0.19]{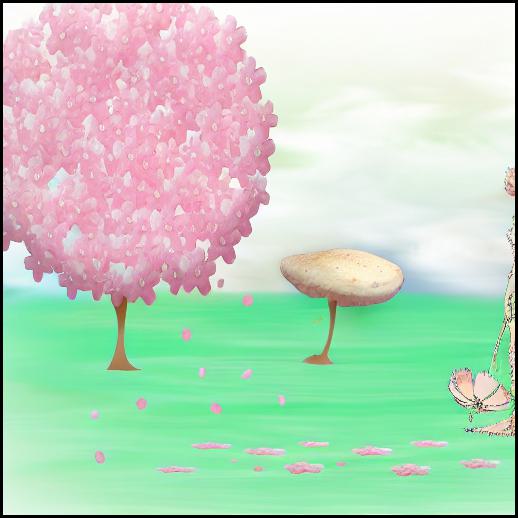}&
         \includegraphics[scale=0.19]{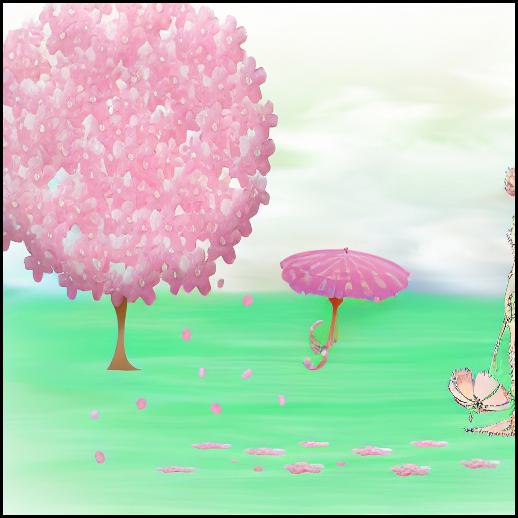} &
         \includegraphics[scale=0.19]{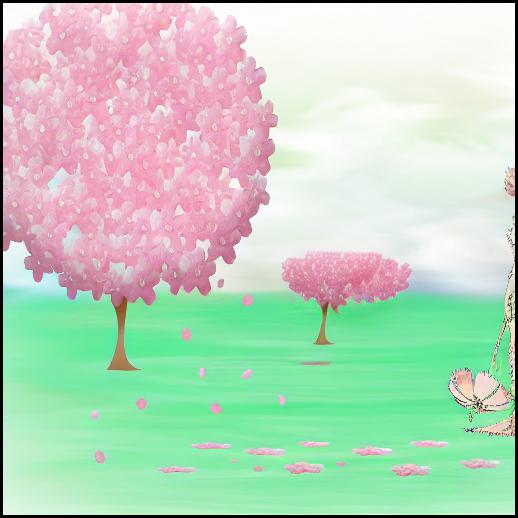}\\
           & & \textit{house} & \textit{icecreama} & \textit{fries} \\ 
         \includegraphics[scale=0.19]{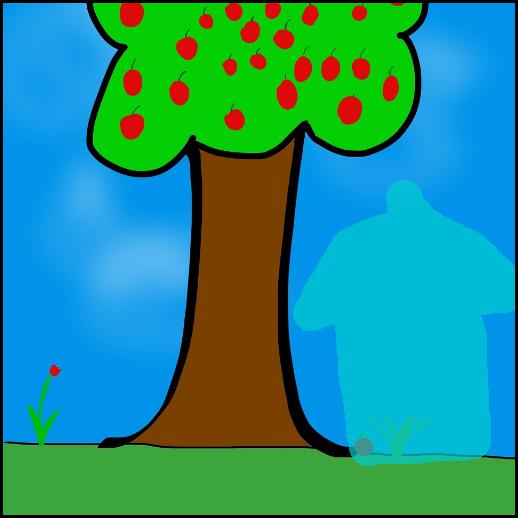}&
          \includegraphics[scale=0.19]{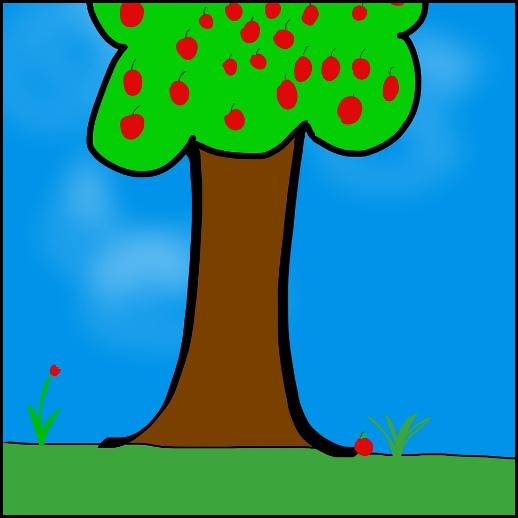}&
           \includegraphics[scale=0.19]{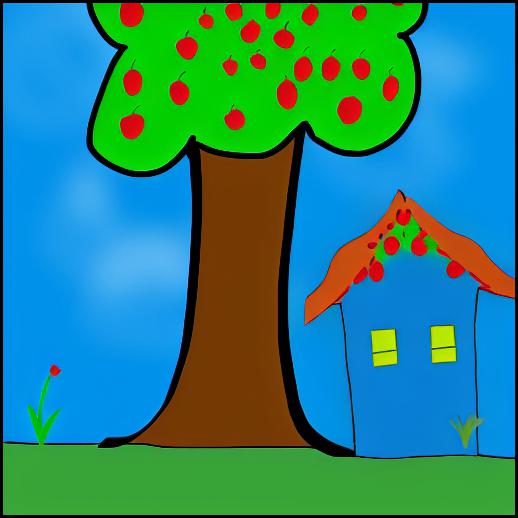}&
         \includegraphics[scale=0.19]{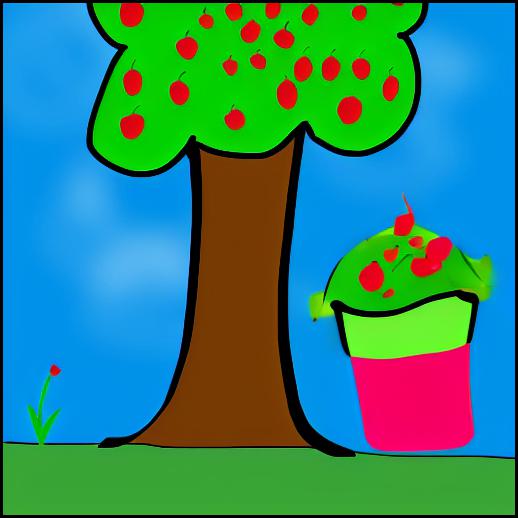} &
         \includegraphics[scale=0.19]{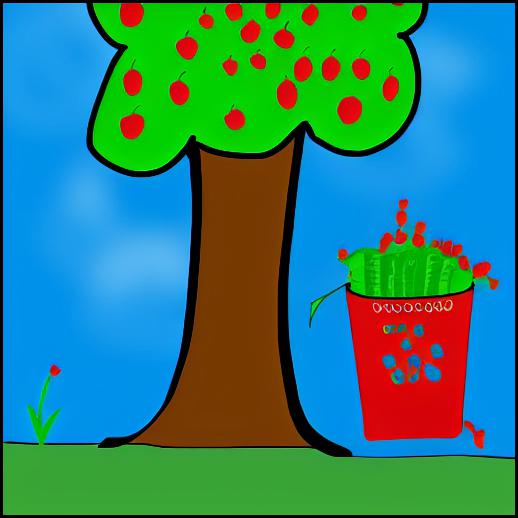}\\
            & & \textit{Taj Mahal} & \textit{church} & \textit{temple} \\ 
         \includegraphics[scale=0.19]{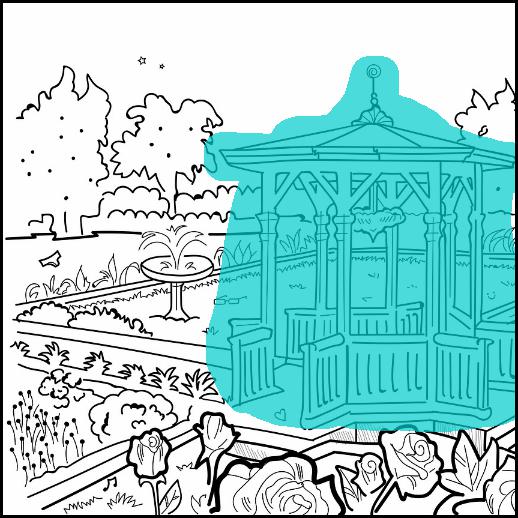}&
          \includegraphics[scale=0.19]{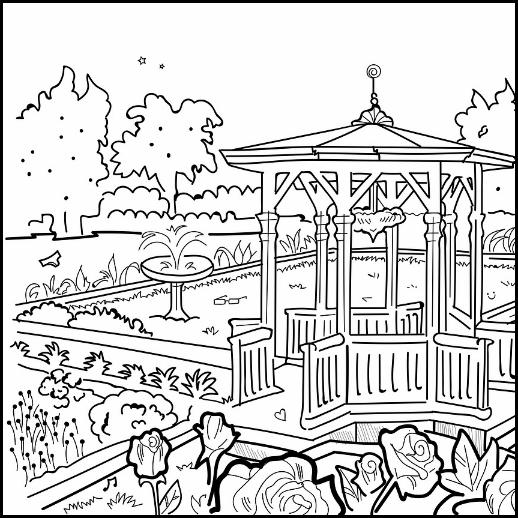}&
           \includegraphics[scale=0.19]{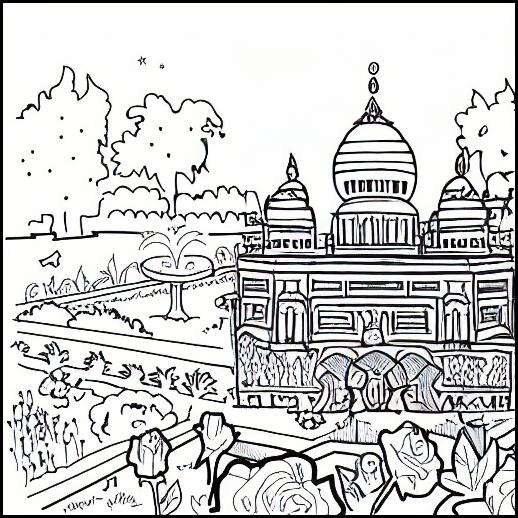}&
         \includegraphics[scale=0.19]{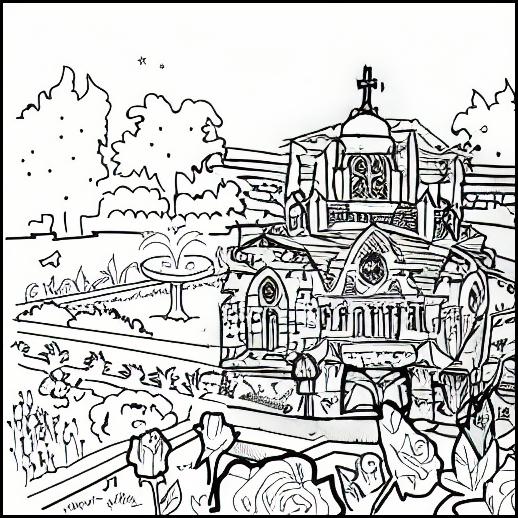} &
         \includegraphics[scale=0.19]{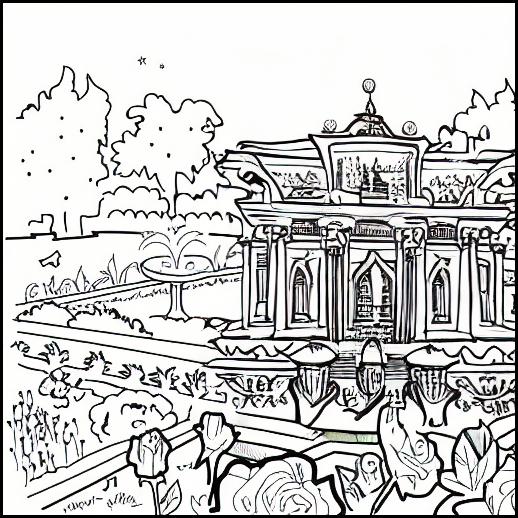}\\
          \bottomrule
    \end{tabular}
    }
    \vspace{-2mm}
    \caption{Text-guided subject-driven image inpainting results for style. Note that the strong inpainting baselines~\citep{rombach2022high,yang2023paint} does not support both text and image guidance.}
    \label{fig:supp_text_style1}
\end{figure*}

\begin{figure*}
    \centering
    \setlength{\tabcolsep}{2pt}
    \scalebox{0.85}{
    \begin{tabular}{cc|ccc}
    \toprule

         Input & Reference Subject & \multicolumn{3}{c}{Text-Guided Subject-Driven Inpainting} \\ \midrule

         & & \textit{golden gate bridge} & \textit{eiffel tower} & \textit{toronton skyline} \\ 
         \includegraphics[scale=0.19]{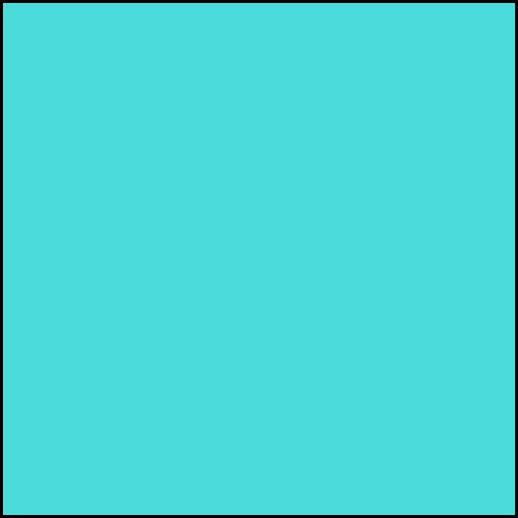}&
          \includegraphics[scale=0.19]{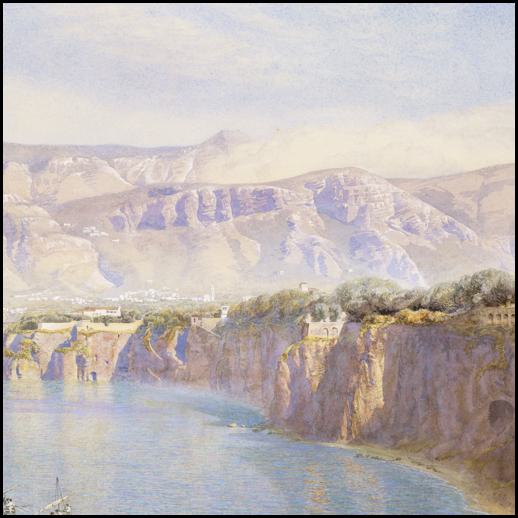}&
           \includegraphics[scale=0.19]{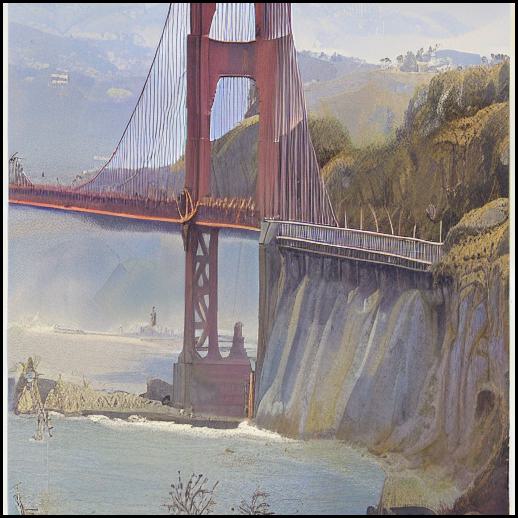}&
         \includegraphics[scale=0.19]{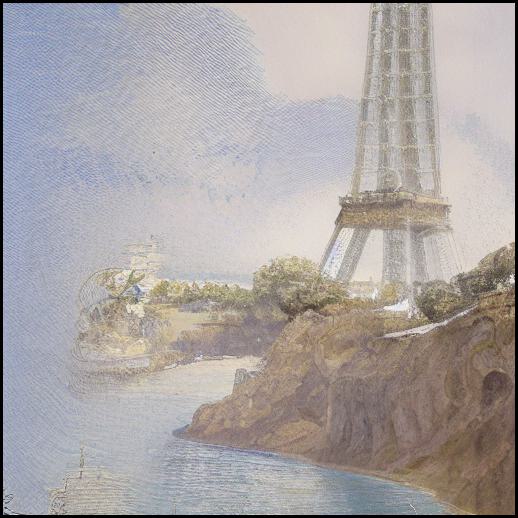} &
         \includegraphics[scale=0.19]{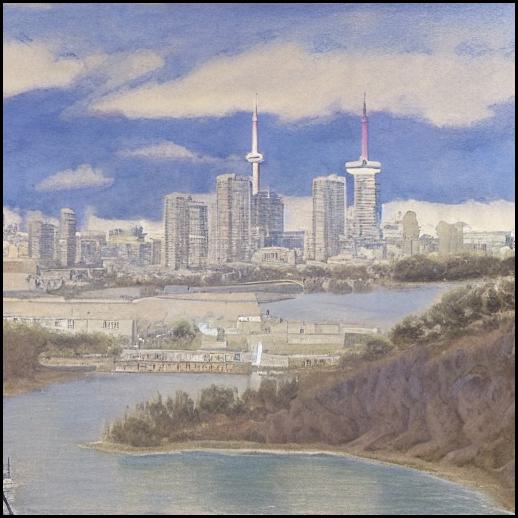}
         \\
           & & \textit{arc de triomphe} & \textit{statue of liberty} & \textit{big ben} \\ 
         \includegraphics[scale=0.19]{figs/supp_text_style/input1.jpg}&
          \includegraphics[scale=0.19]{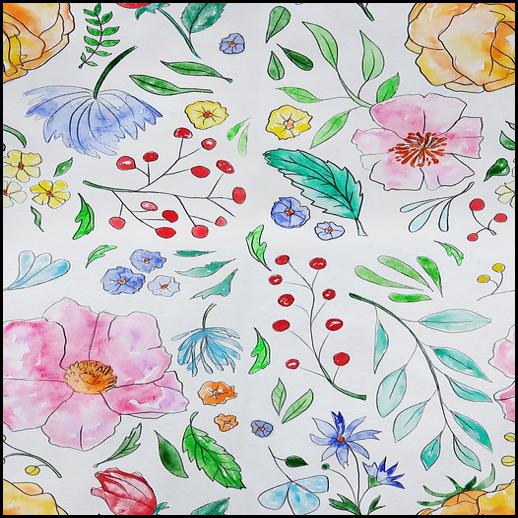}&
           \includegraphics[scale=0.19]{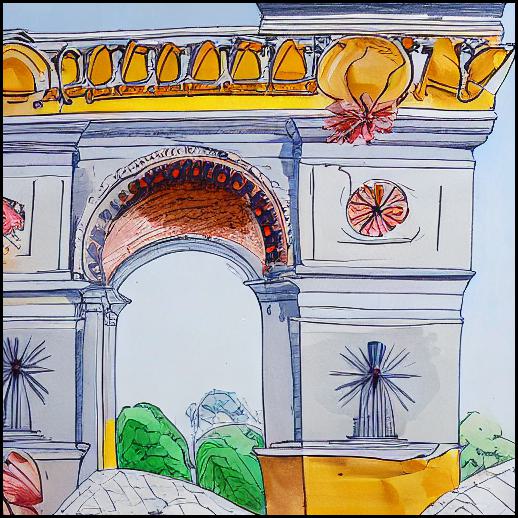}&
         \includegraphics[scale=0.19]{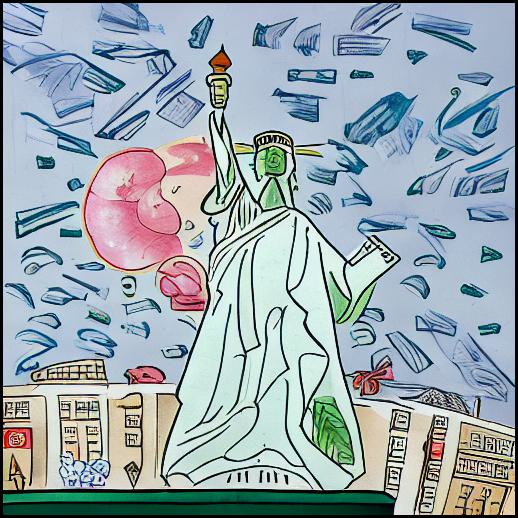} &
         \includegraphics[scale=0.19]{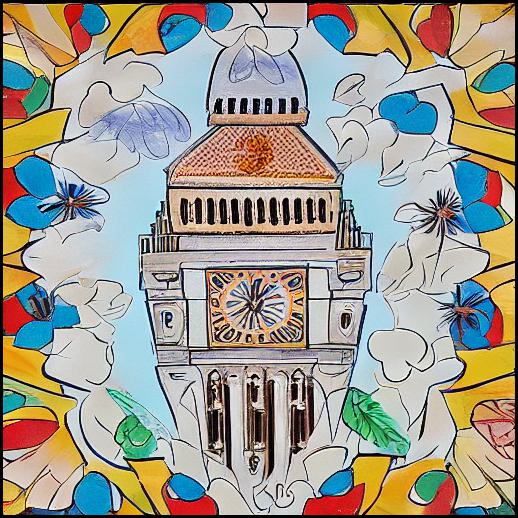}
         \\
           & & \textit{dog} & \textit{panda} & \textit{unicorn} \\ 
         \includegraphics[scale=0.19]{figs/supp_text_style/input1.jpg}&
          \includegraphics[scale=0.19]{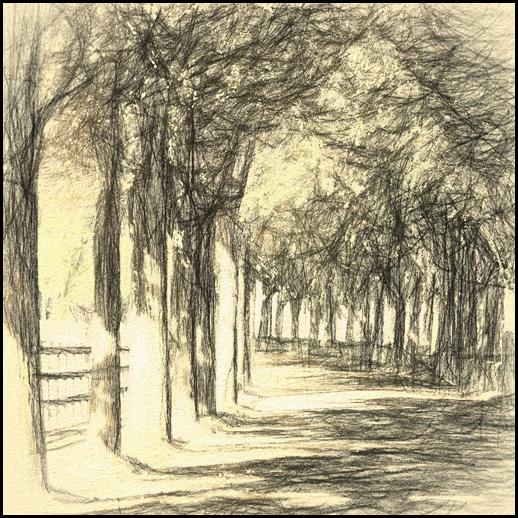}&
           \includegraphics[scale=0.19]{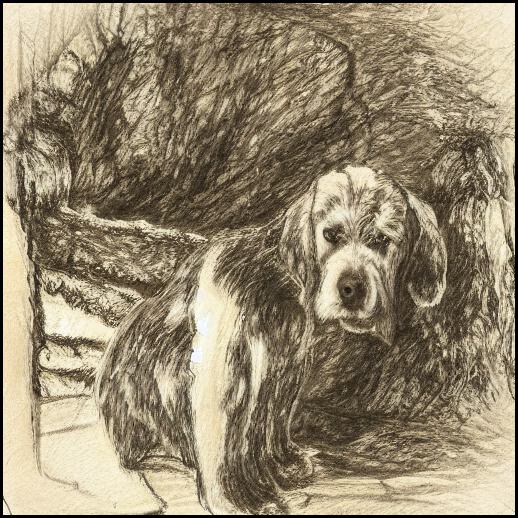}&
         \includegraphics[scale=0.19]{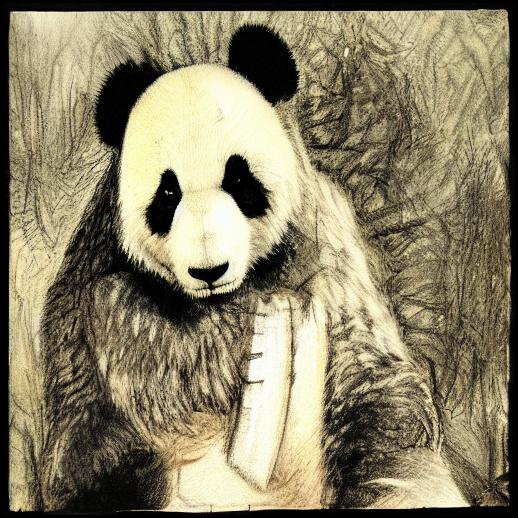} &
         \includegraphics[scale=0.19]{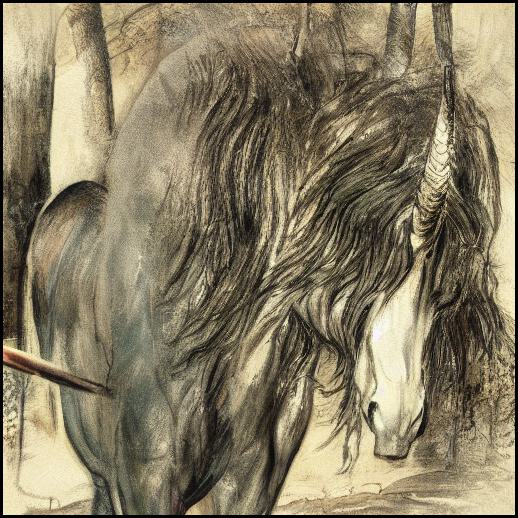}
         \\
         & & \textit{bedroom} & \textit{desk} & \textit{castle} \\ 
         \includegraphics[scale=0.19]{figs/supp_text_style/input1.jpg}&
          \includegraphics[scale=0.19]{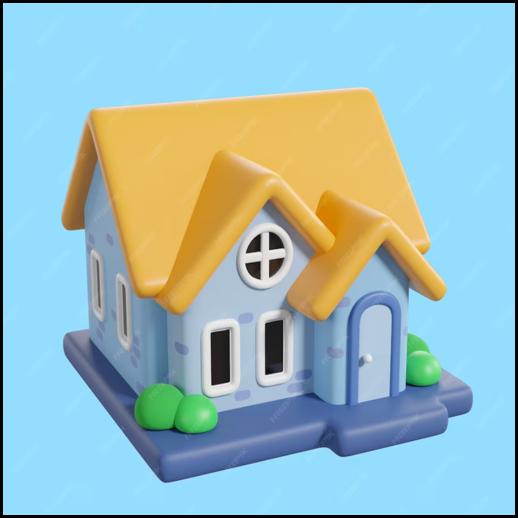}&
           \includegraphics[scale=0.19]{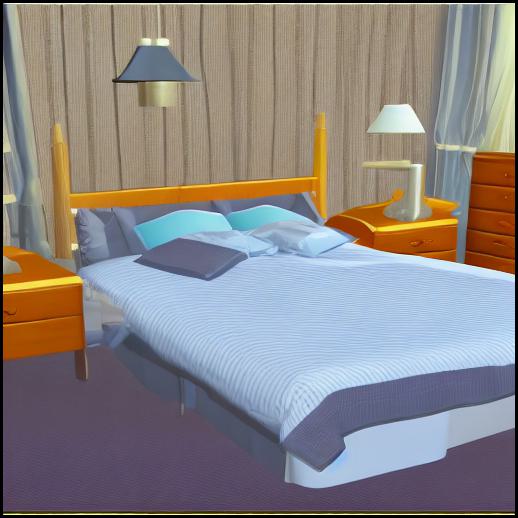}&
         \includegraphics[scale=0.19]{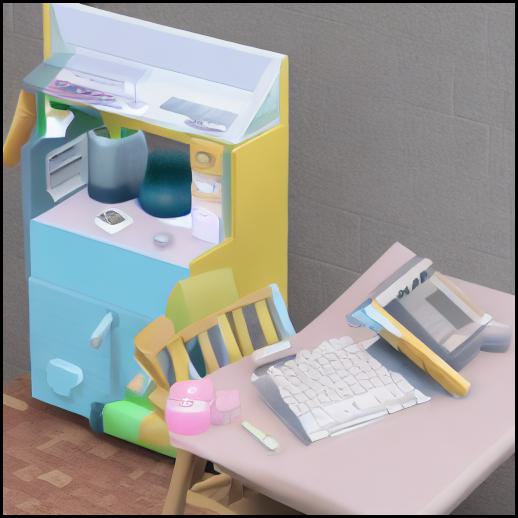} &
         \includegraphics[scale=0.19]{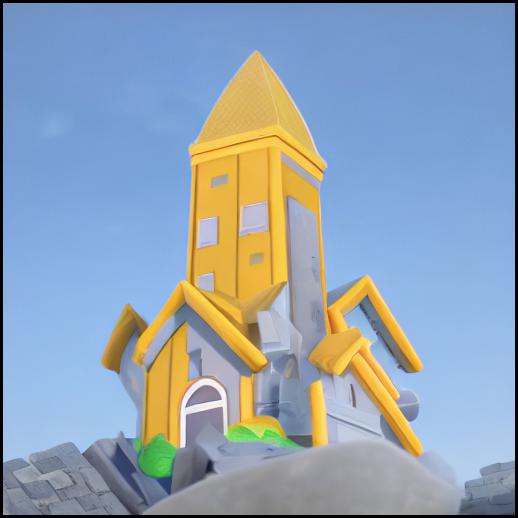}
         \\
           & & \textit{whale} & \textit{garden} & \textit{cat} \\ 
         \includegraphics[scale=0.19]{figs/supp_text_style/input1.jpg}&
          \includegraphics[scale=0.19]{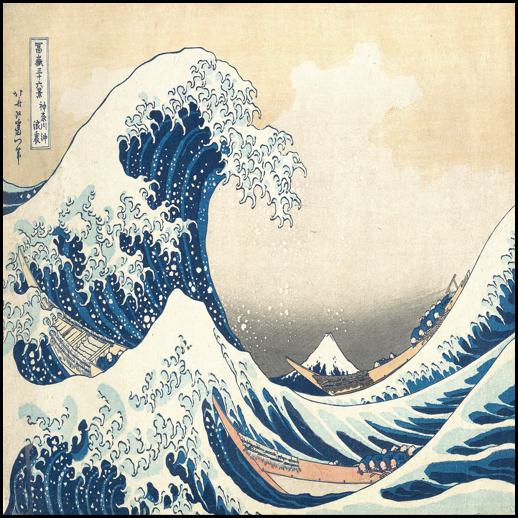}&
           \includegraphics[scale=0.19]{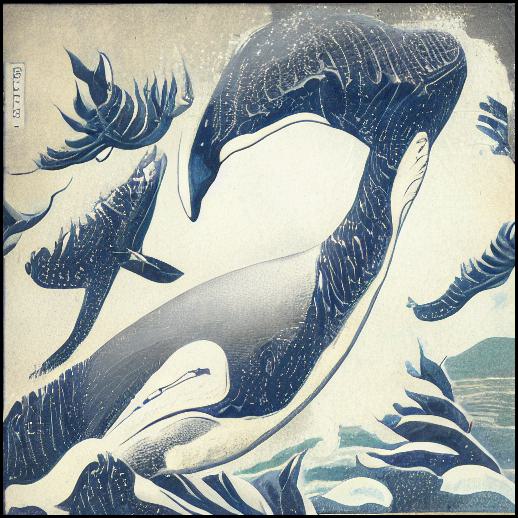}&
         \includegraphics[scale=0.19]{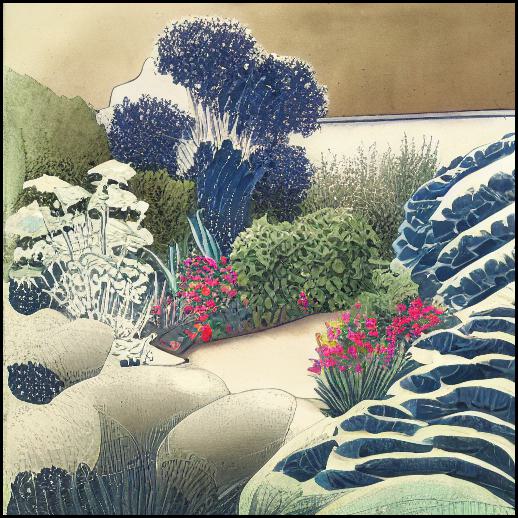} &
         \includegraphics[scale=0.19]{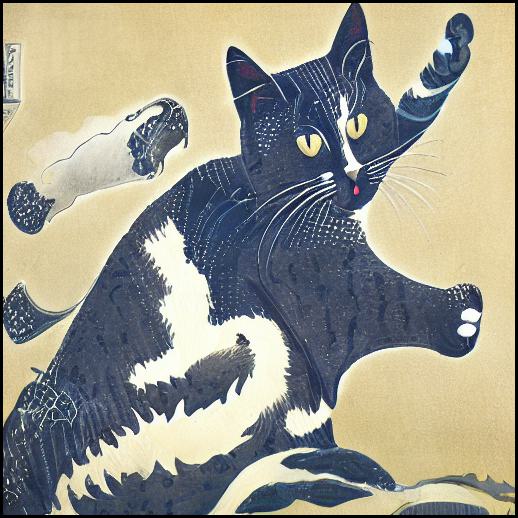}
         \\
           & & \textit{lotus} & \textit{bird} & \textit{teddy bear} \\ 
         \includegraphics[scale=0.19]{figs/supp_text_style/input1.jpg}&
          \includegraphics[scale=0.19]{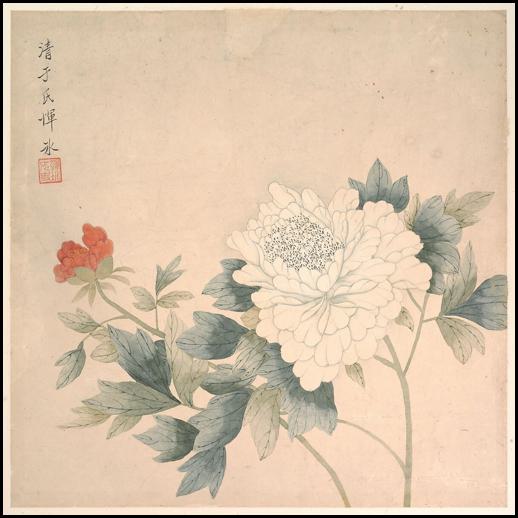}&
           \includegraphics[scale=0.19]{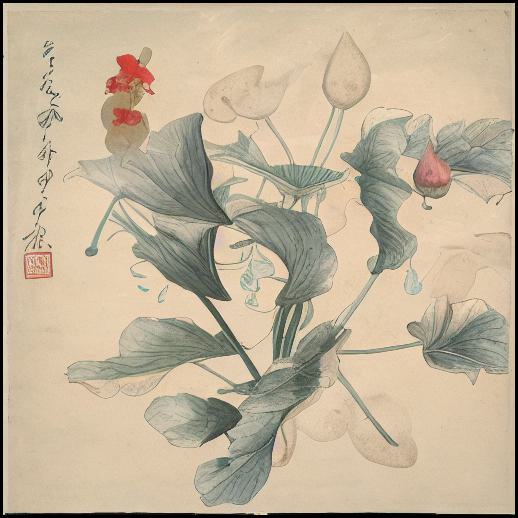}&
         \includegraphics[scale=0.19]{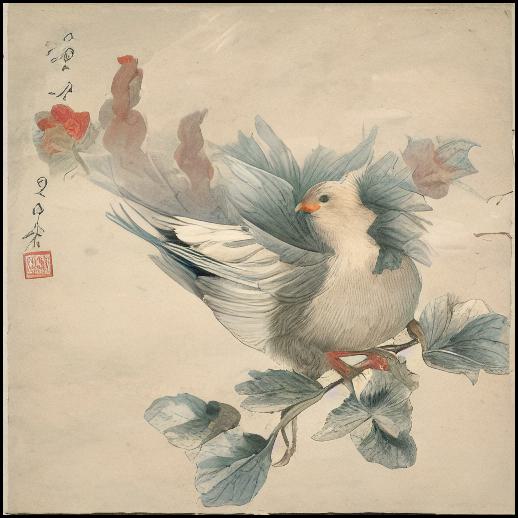} &
         \includegraphics[scale=0.19]{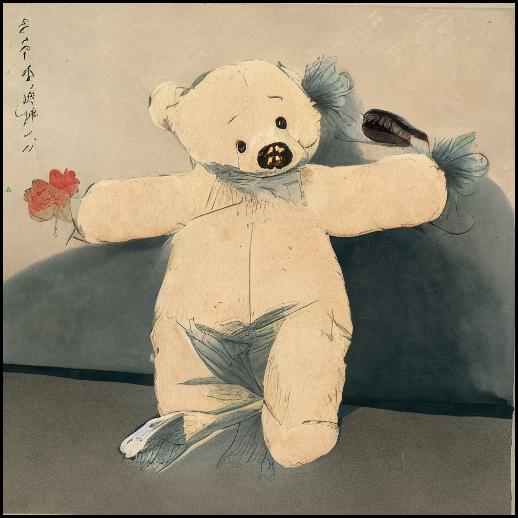}
         \\ \bottomrule
    \end{tabular}
    }
    \vspace{-2mm}
    \caption{Text-guided subject-driven image inpainting results for style. Note that the strong inpainting baselines~\citep{rombach2022high,yang2023paint} does not support both text and image guidance.}
    \label{fig:supp_text_style2}
\end{figure*}

In this section, we present more examples of the text-guided subject driven image inpainting results. As shown in Fig. \ref{fig:supp_text}, \ref{fig:supp_text_style1} and \ref{fig:supp_text_style2},, our method is able to inpaint the missing region with the reference object while following the given text prompt, demonstrating many interesting applications of our proposed text-guided subject-driven image inpainting task. Firstly, the first three rows in Fig. \ref{fig:supp_text} show that we are able to generate the reference dog (cat)  
while changing their attributes. For example, our method changes the facial expression of the corgi in the first row and generates dog and sunglasses in the second row. Importantly, the identity of the dog and cat are preserved. The fourth and fifth row demonstrate that our method is able to change the shape of the reference object with the text prompt, such as the heart-shaped ballon. The final row in Fig. \ref{fig:supp_text} show that the model possess the ability to change the property of the reference object while preserving the main unique part of the reference object, e.g., the models learns to generate boots in the image while preserving the color and style of the reference shoes. 

Besides the amazing attribute editing ability, our method is also equipped with inpainting with style in reference image. As shown in Fig. \ref{fig:supp_text_style1}, we can fill the missing region in the input image by using the same image as the reference image. Thanks to our proposed discriminative subject feature method, our method learns to extract the essential information of the input imagen and enables the model to generate objects in the region while following the same style of the input image, which avoids the potential inconsistency between the generated content and the given background. Our method handles various styles, including the pencil drawing, children's illustration and oil paintings. 

Another interesting application of our method is the image generation with style. The whole input image is masked and we are provided with the reference and text prompt. In other words, the model needs to learn to utilize the text and reference image together to  generate the image, otherwise it will fails. Fortunately, our proposed decoupling regularization allows us to achieve this challenging task easily as demonstrated in Fig. \ref{fig:supp_text_style2}. 

\section{Subject-Driven Image Inpainting}
\label{sec:supp_image}
\begin{figure*}
    \centering
    \setlength{\tabcolsep}{2pt}
    \scalebox{0.9}{
    \begin{tabular}{ccccc}
    Masked Input & Reference & Stable Inpaint \citep{rombach2022high} & PBE \citep{yang2023paint} & Ours\\
        \includegraphics[scale=0.18]{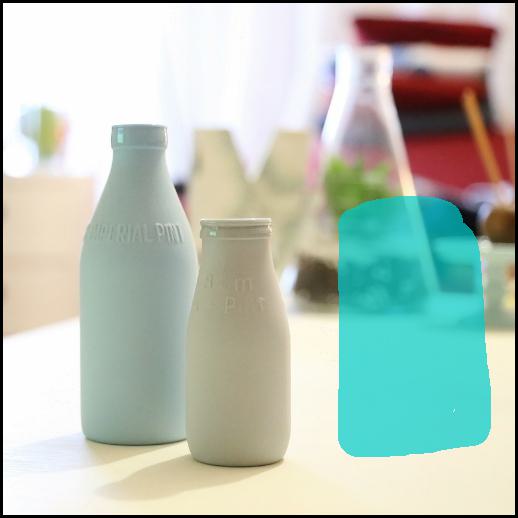} &
         \includegraphics[scale=0.18]{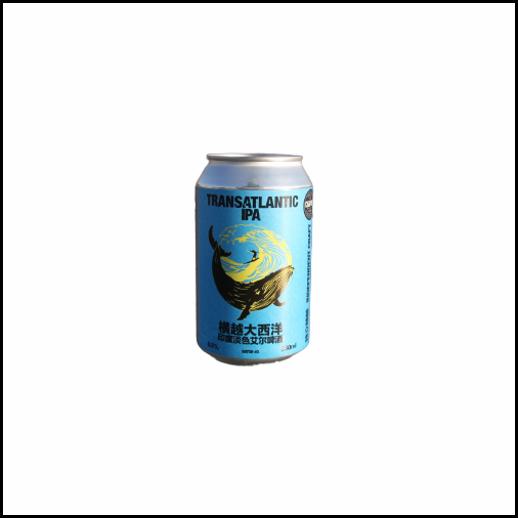} &
          \includegraphics[scale=0.18]{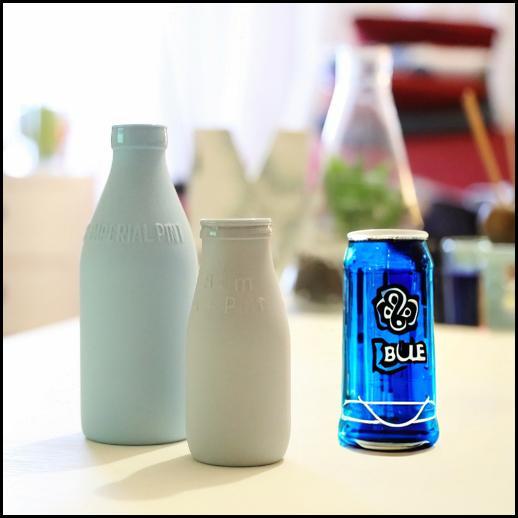} &
           \includegraphics[scale=0.18]{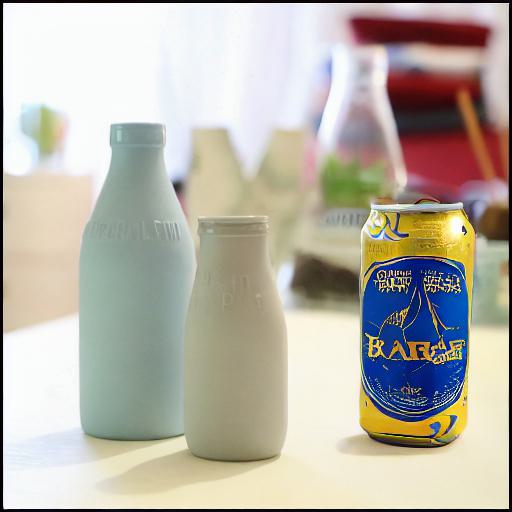} &
            \includegraphics[scale=0.18]{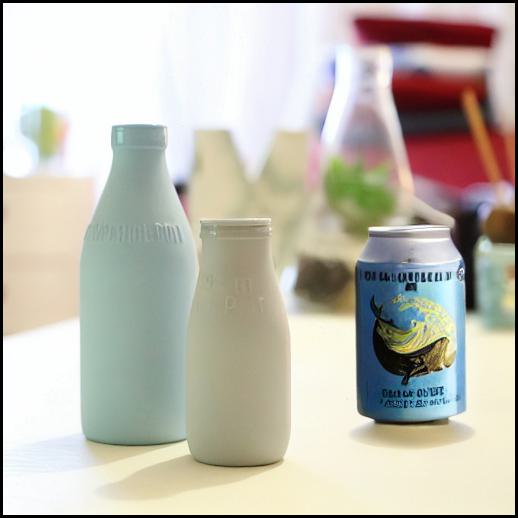} 
        \\
           \includegraphics[scale=0.18]{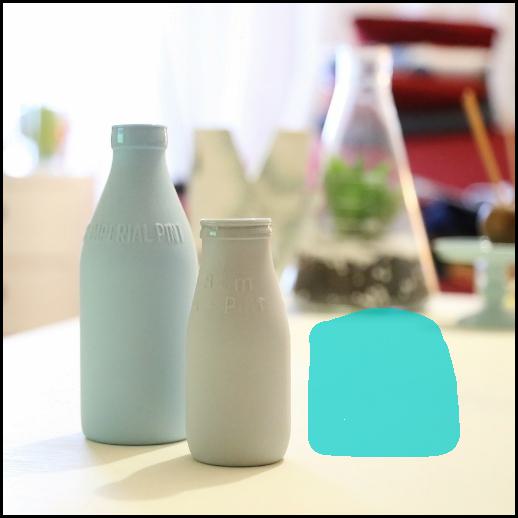} &
         \includegraphics[scale=0.18]{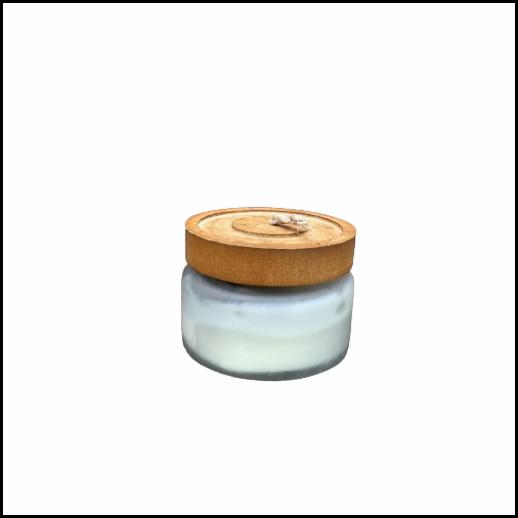} &
          \includegraphics[scale=0.18]{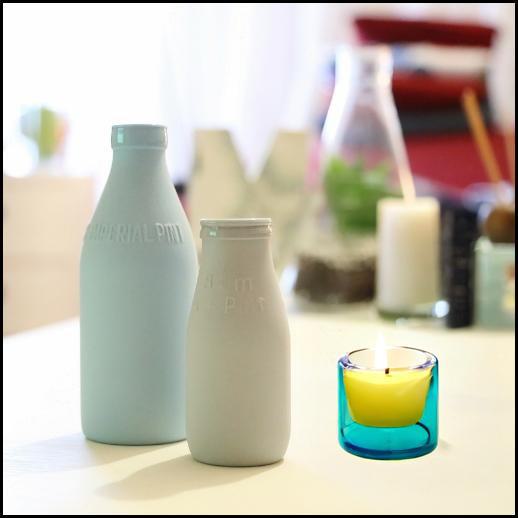} &
           \includegraphics[scale=0.18]{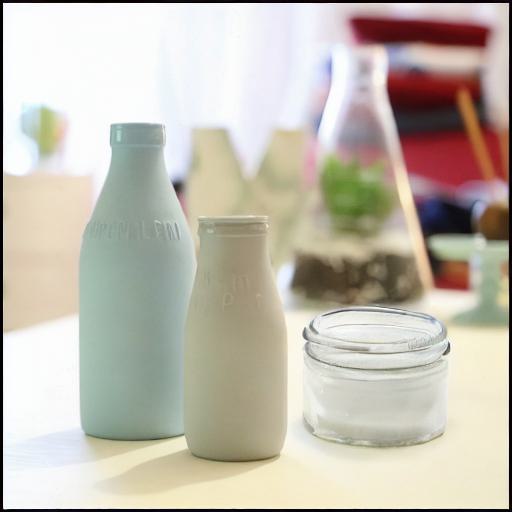} &
            \includegraphics[scale=0.18]{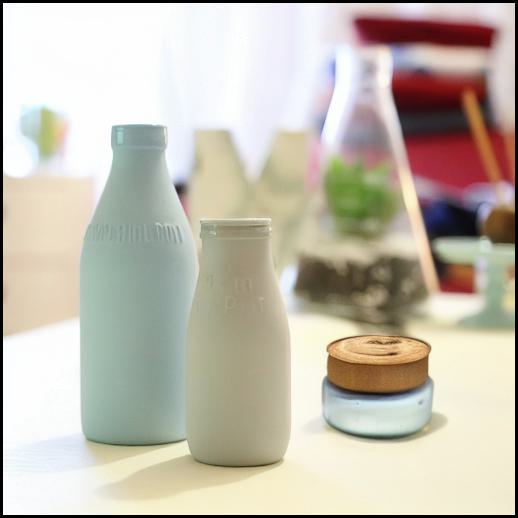} 
        \\
        \includegraphics[scale=0.18]{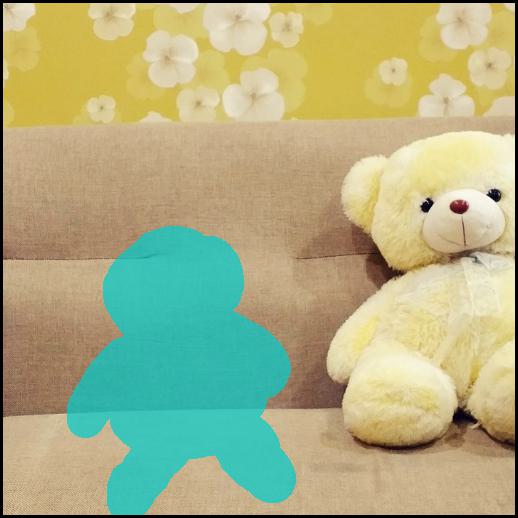} &
         \includegraphics[scale=0.18]{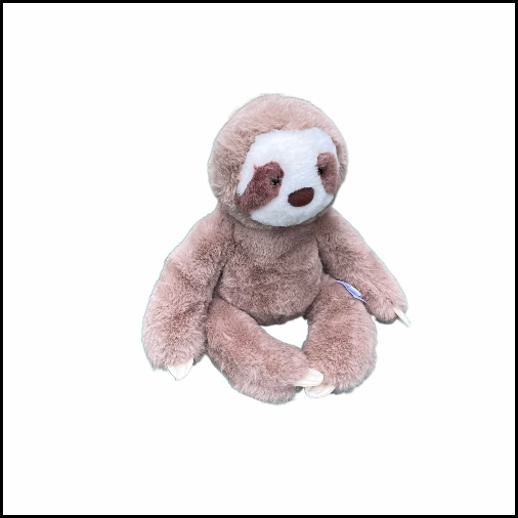} &
          \includegraphics[scale=0.18]{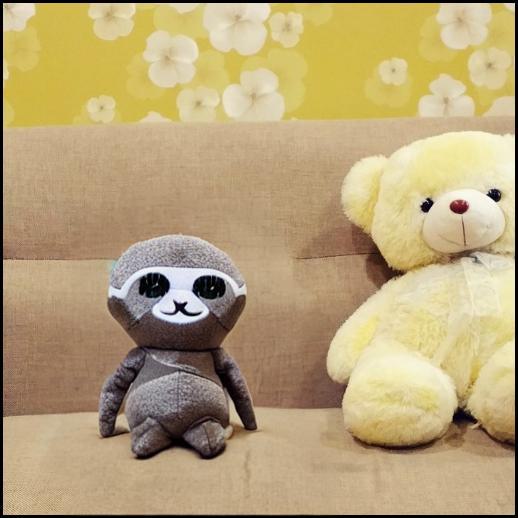} &
           \includegraphics[scale=0.18]{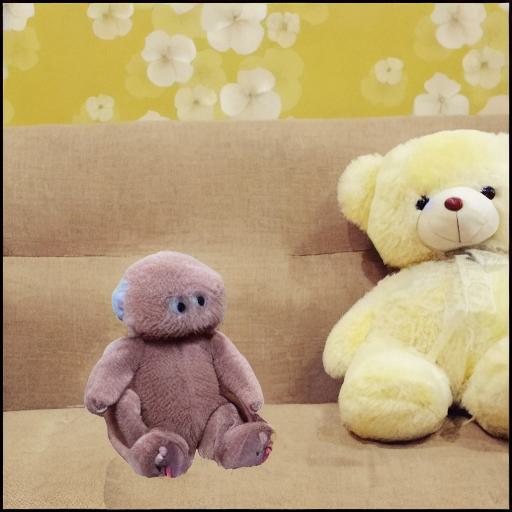} &
            \includegraphics[scale=0.18]{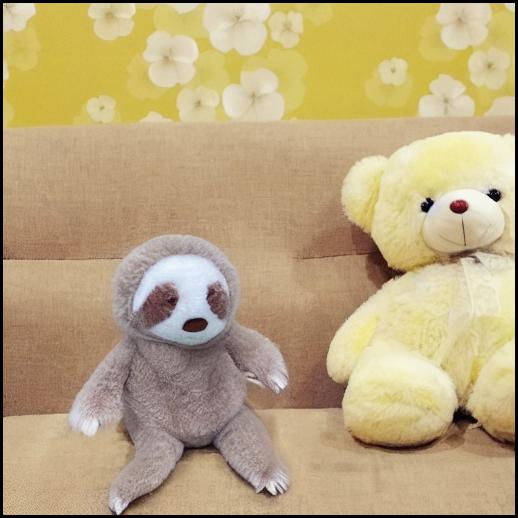} 
        \\
        \includegraphics[scale=0.18]{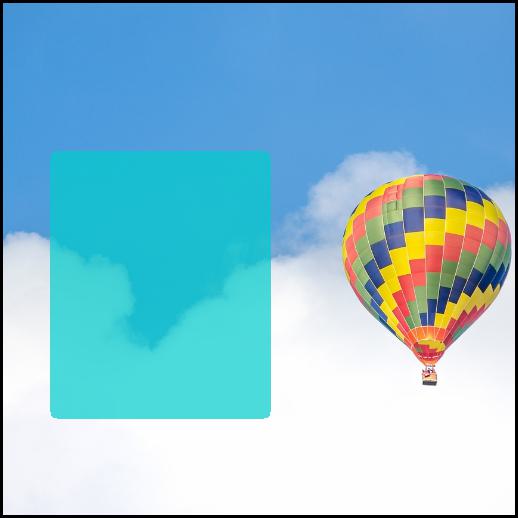} &
         \includegraphics[scale=0.18]{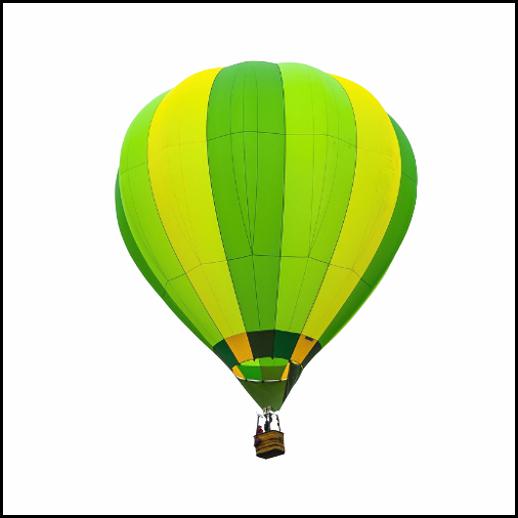} &
          \includegraphics[scale=0.18]{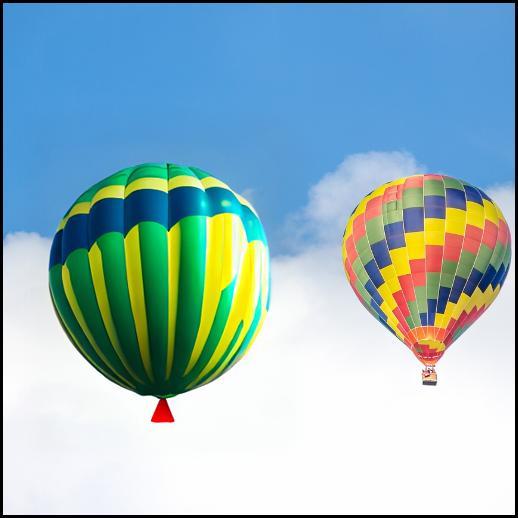} &
           \includegraphics[scale=0.18]{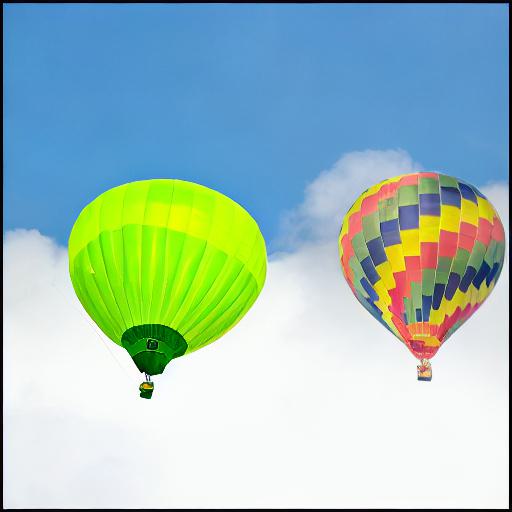} &
            \includegraphics[scale=0.18]{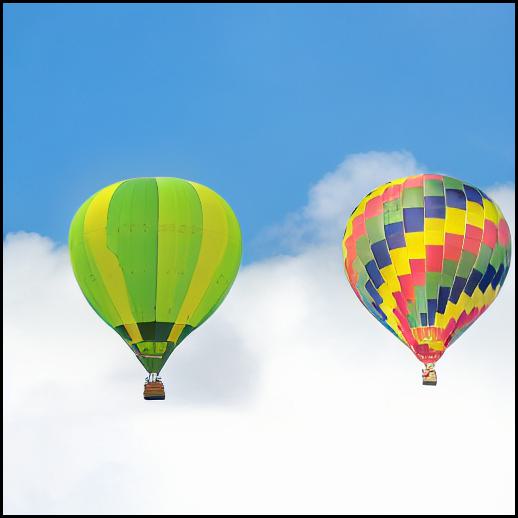} 
        \\
       \includegraphics[scale=0.18]{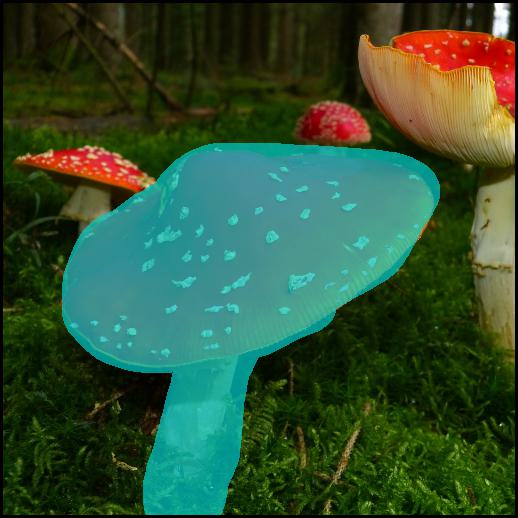} &
         \includegraphics[scale=0.18]{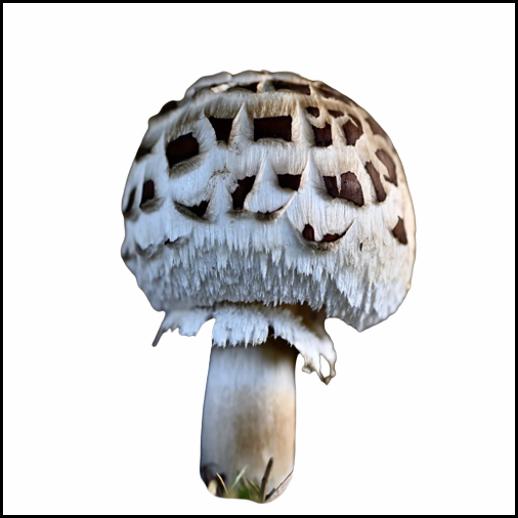} &
          \includegraphics[scale=0.18]{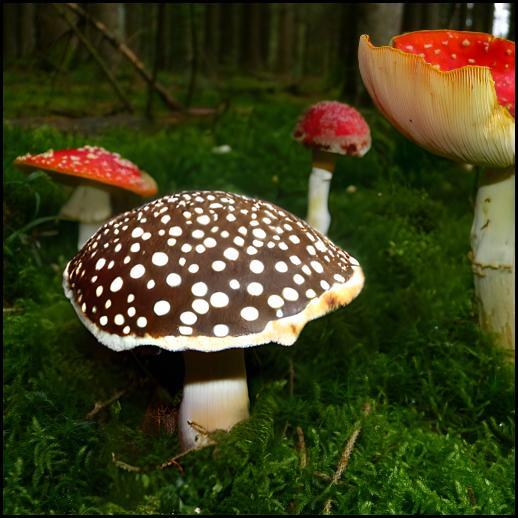} &
           \includegraphics[scale=0.18]{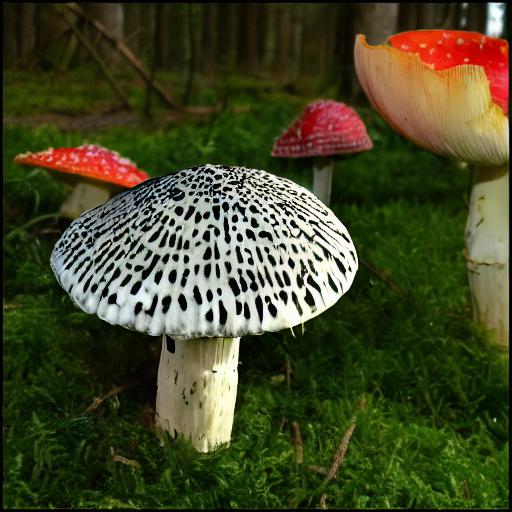} &
            \includegraphics[scale=0.18]{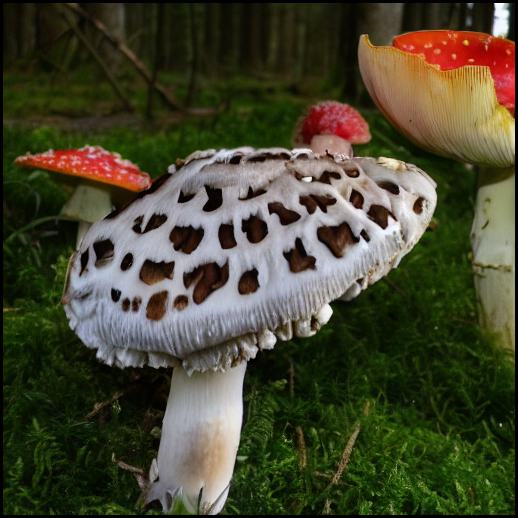} 
        \\
         \includegraphics[scale=0.18]{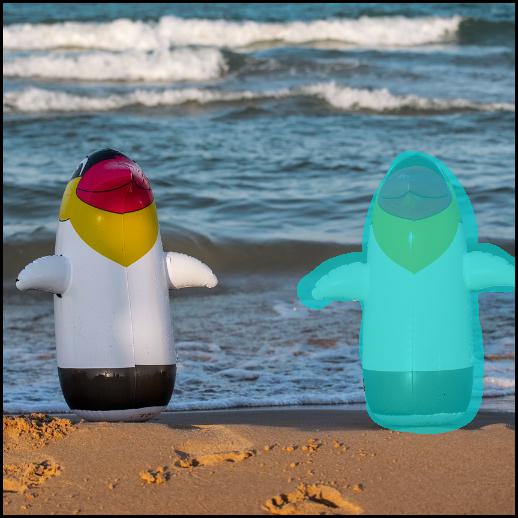} &
         \includegraphics[scale=0.18]{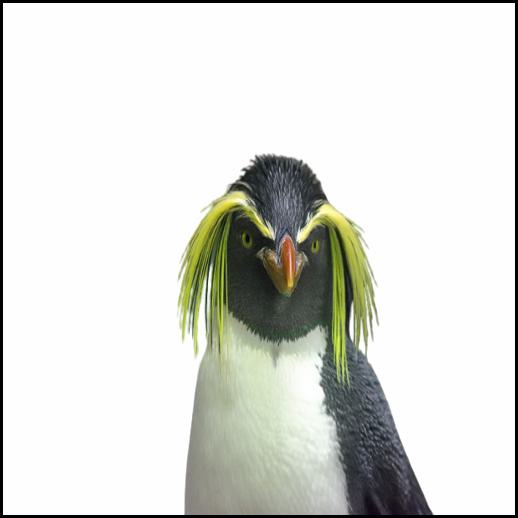} &
          \includegraphics[scale=0.18]{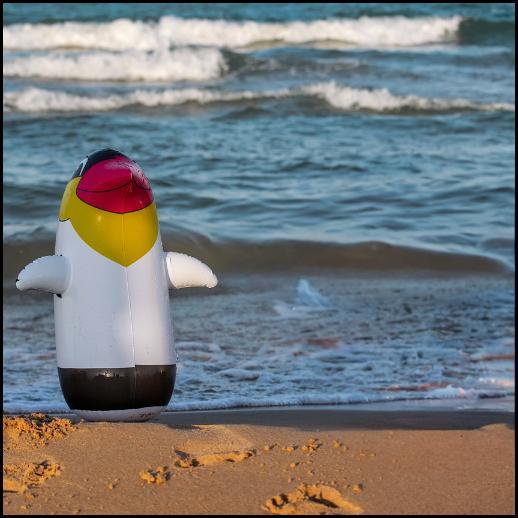} &
           \includegraphics[scale=0.18]{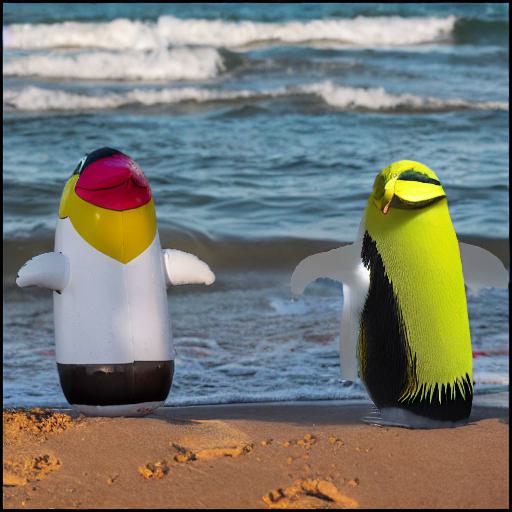} &
            \includegraphics[scale=0.18]{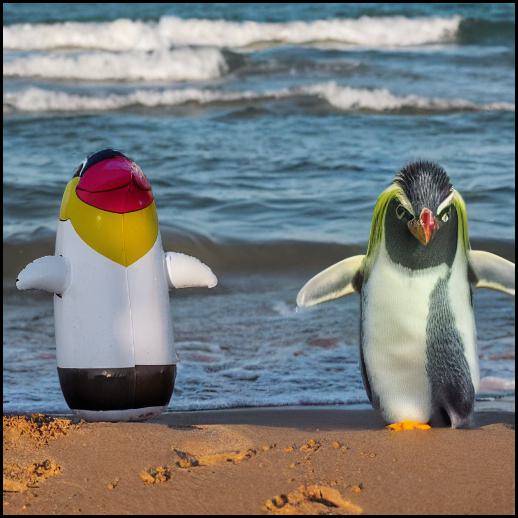} 
        \\
           \includegraphics[scale=0.18]{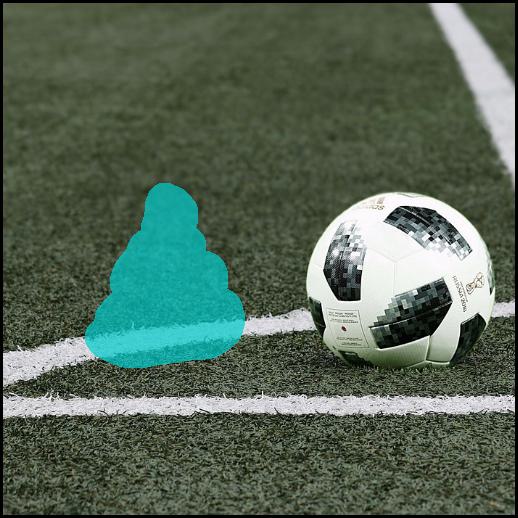} &
         \includegraphics[scale=0.18]{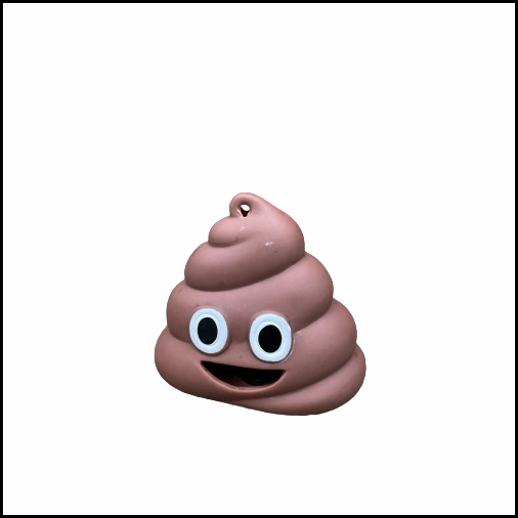} &
          \includegraphics[scale=0.18]{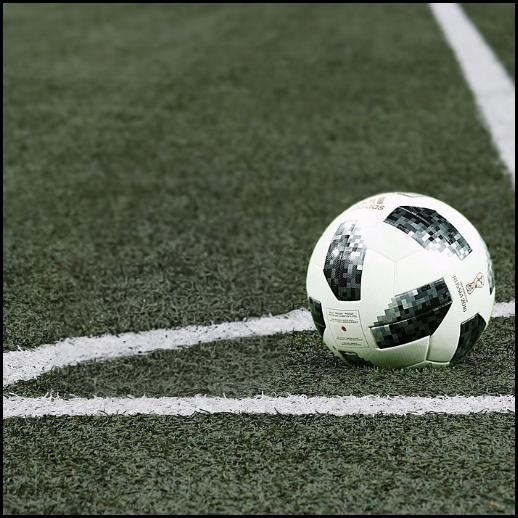} &
           \includegraphics[scale=0.18]{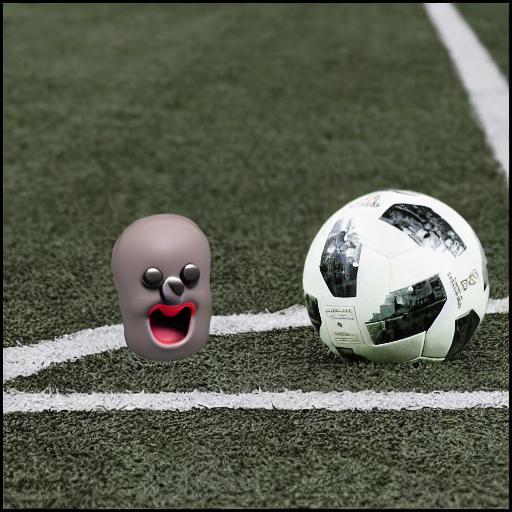} &
            \includegraphics[scale=0.18]{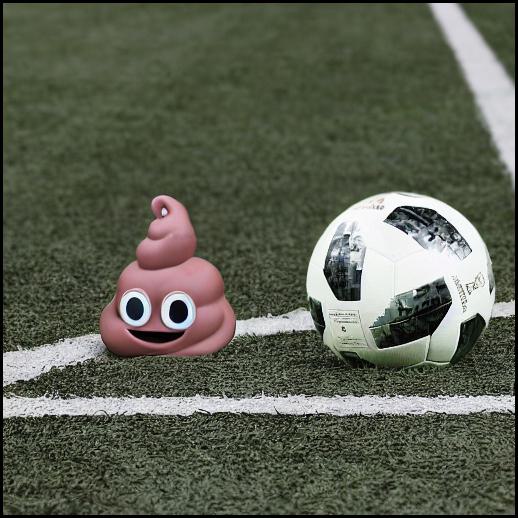} 
        \\
    \end{tabular}
    }
    \caption{Image-guided inpainting results \textbf{Part 1}.}
    \label{fig:supp_image1}
\end{figure*}

\begin{figure*}
    \centering
    \setlength{\tabcolsep}{2pt}
    \scalebox{0.9}{
    \begin{tabular}{ccccc}
    Masked Input & Reference & Stable Inpaint \citep{rombach2022high} & PBE \citep{yang2023paint} & Ours\\
        \includegraphics[scale=0.18]{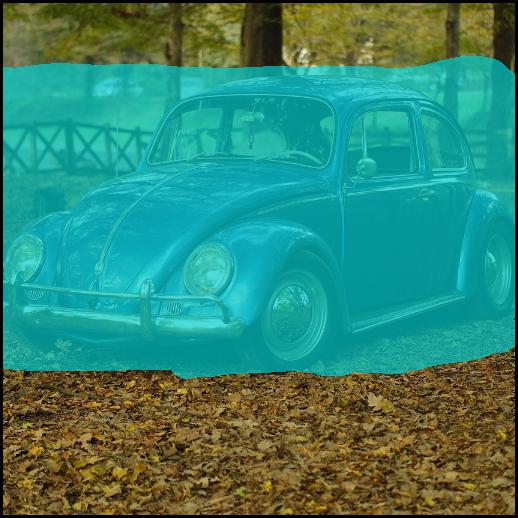} &
         \includegraphics[scale=0.18]{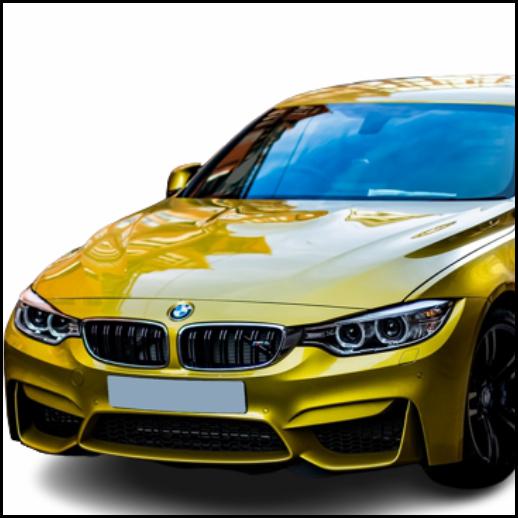} &
          \includegraphics[scale=0.18]{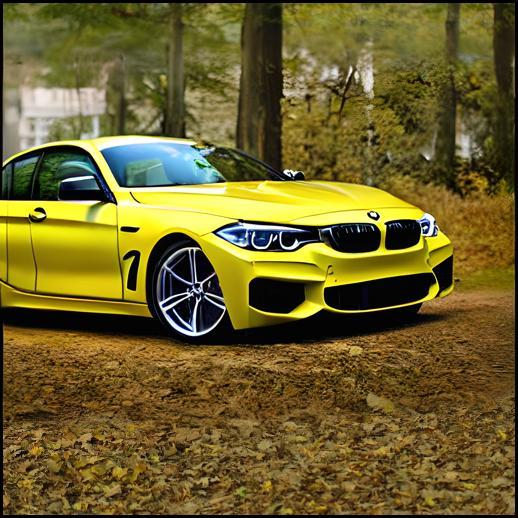} &
           \includegraphics[scale=0.18]{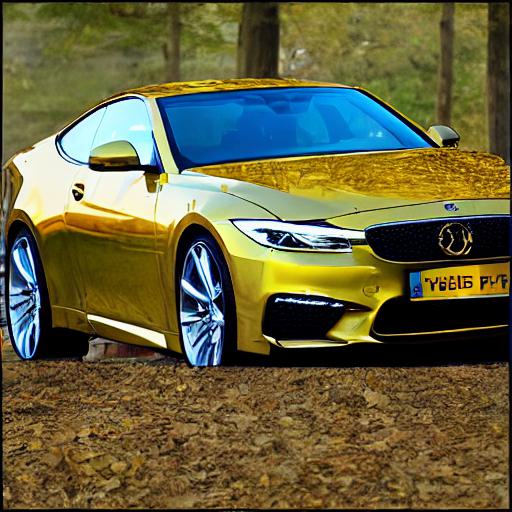} &
            \includegraphics[scale=0.18]{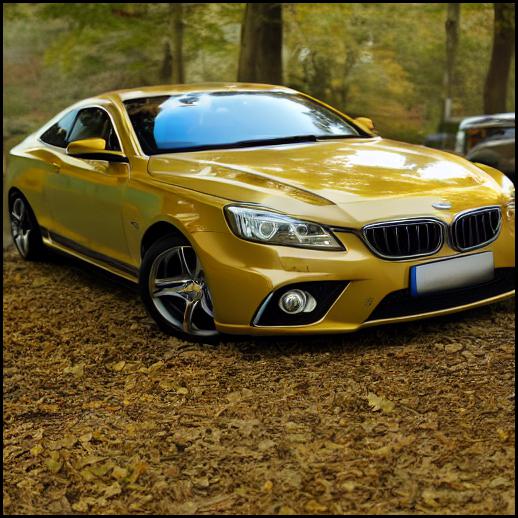} \\
             \includegraphics[scale=0.18]{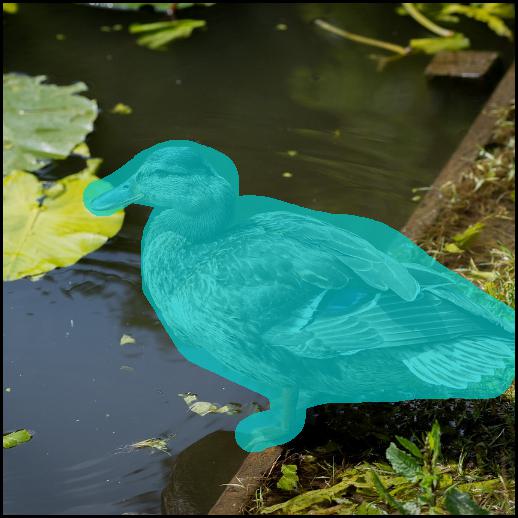} &
         \includegraphics[scale=0.18]{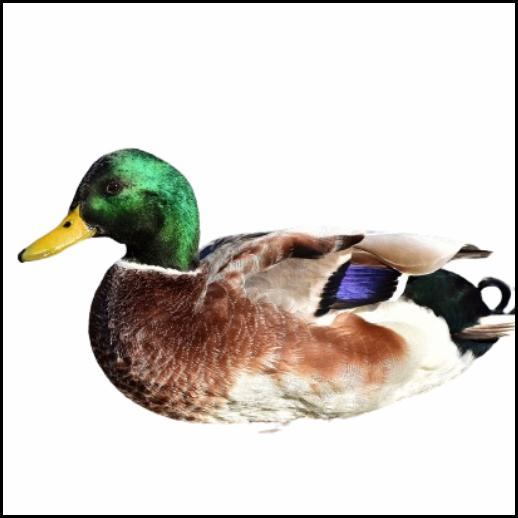} &
          \includegraphics[scale=0.18]{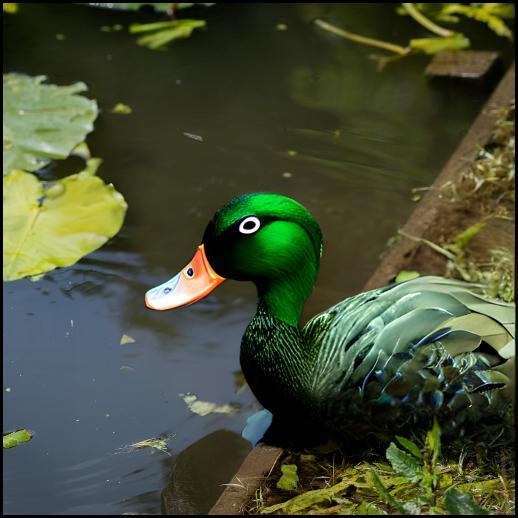} &
           \includegraphics[scale=0.18]{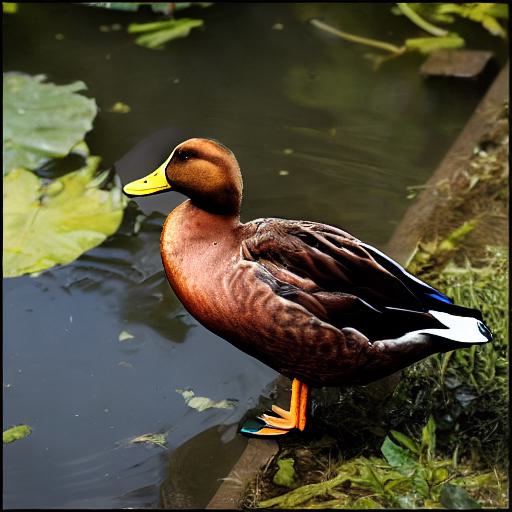} &
            \includegraphics[scale=0.18]{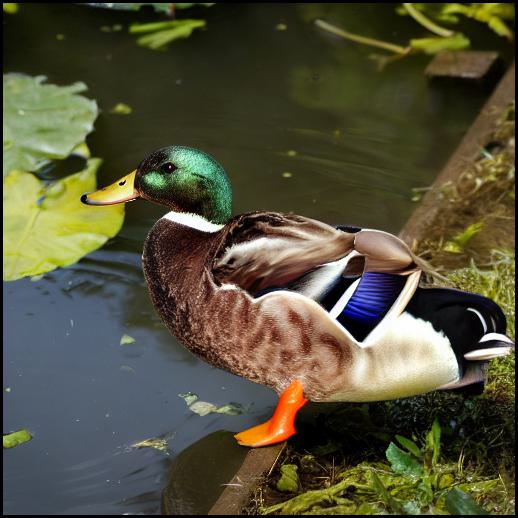} \\
                \includegraphics[scale=0.18]{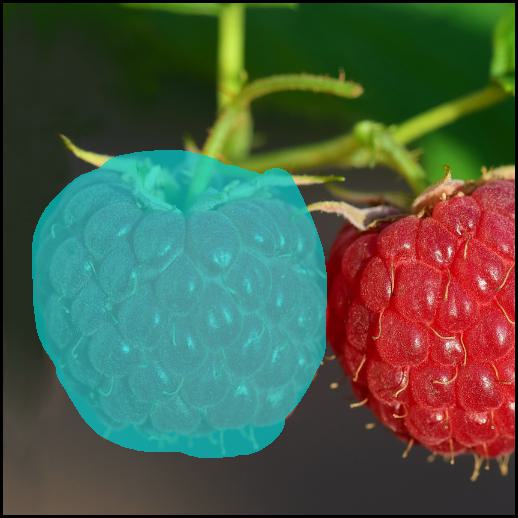} &
         \includegraphics[scale=0.18]{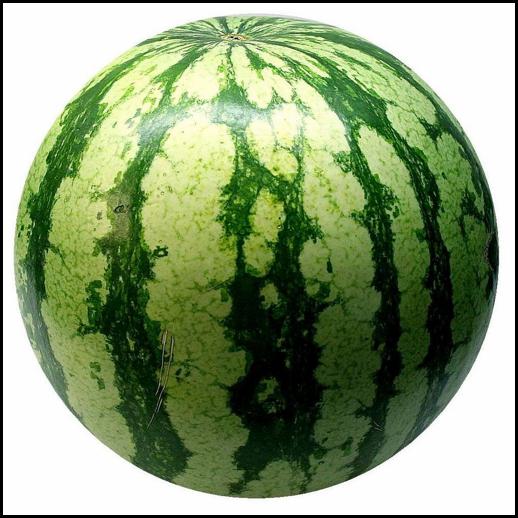} &
          \includegraphics[scale=0.18]{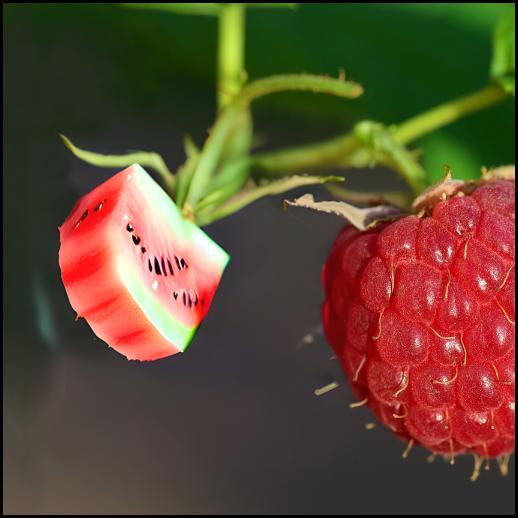} &
           \includegraphics[scale=0.18]{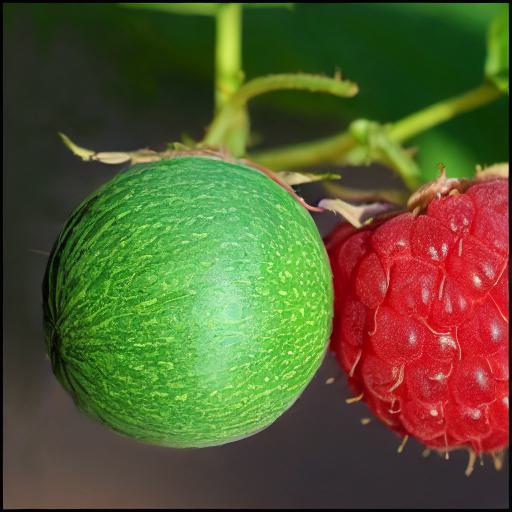} &
            \includegraphics[scale=0.18]{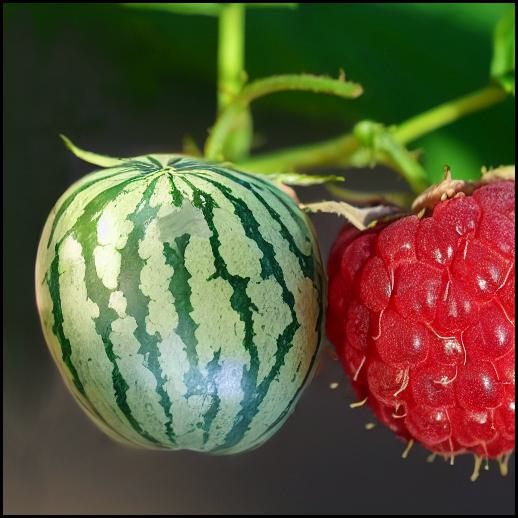} \\
             \includegraphics[scale=0.18]{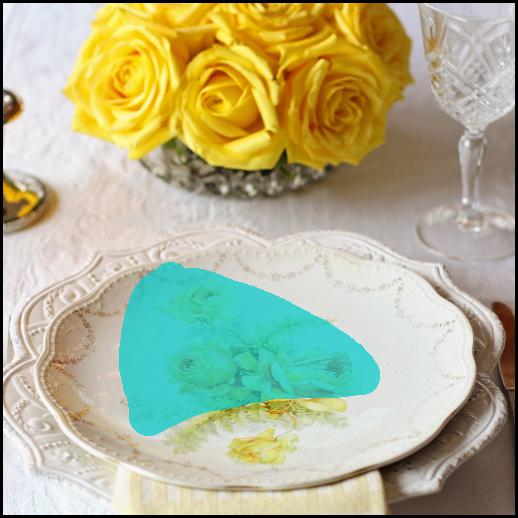} &
         \includegraphics[scale=0.18]{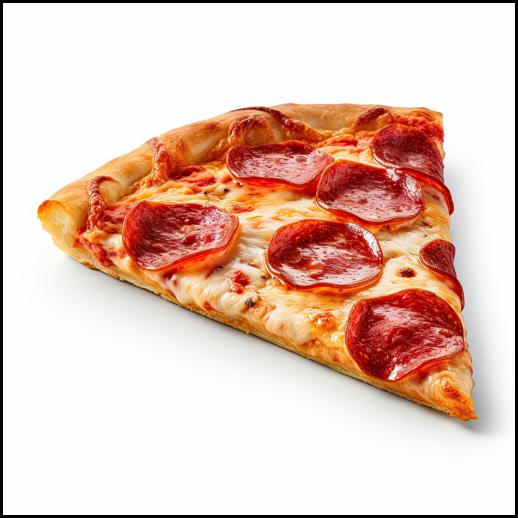} &
          \includegraphics[scale=0.18]{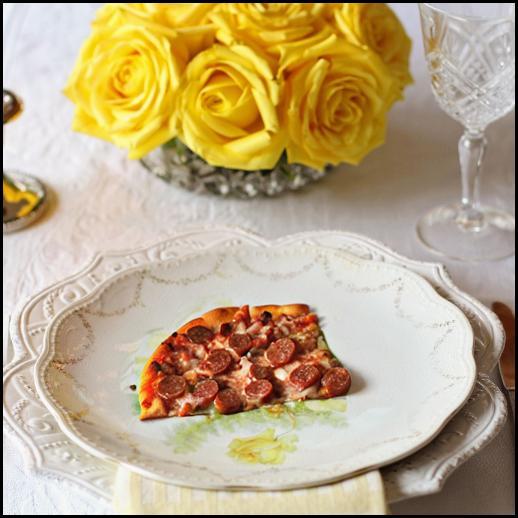} &
           \includegraphics[scale=0.18]{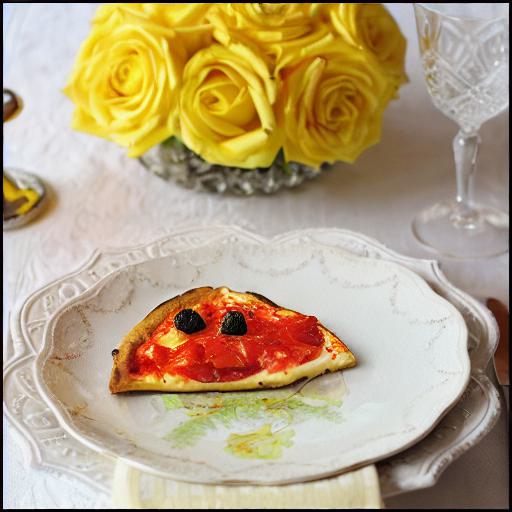} &
            \includegraphics[scale=0.18]{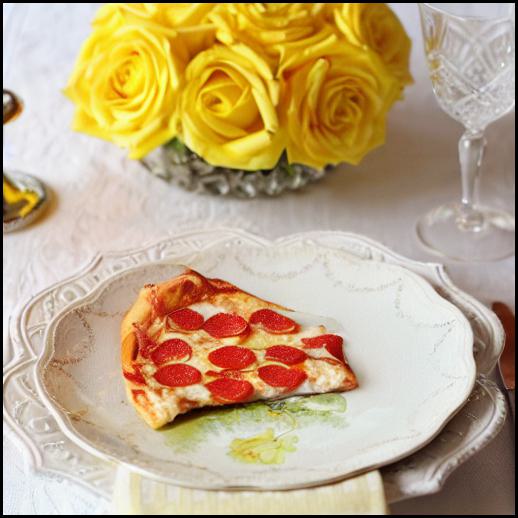}  \\
             \includegraphics[scale=0.18]{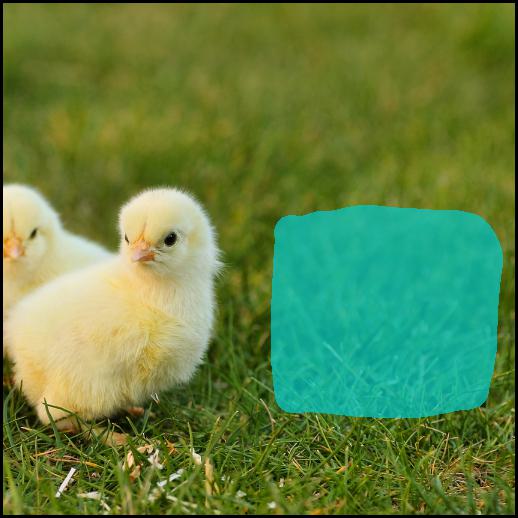} &
         \includegraphics[scale=0.18]{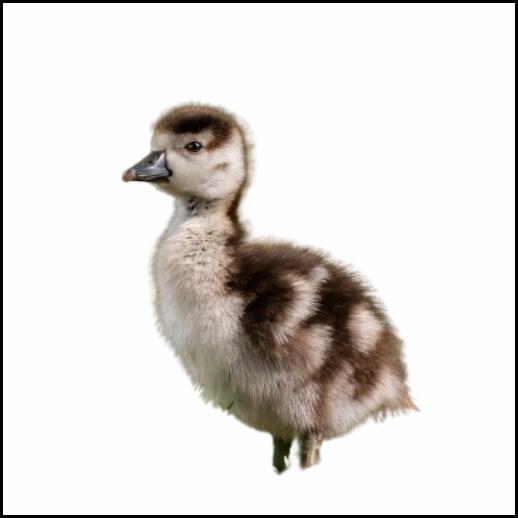} &
          \includegraphics[scale=0.18]{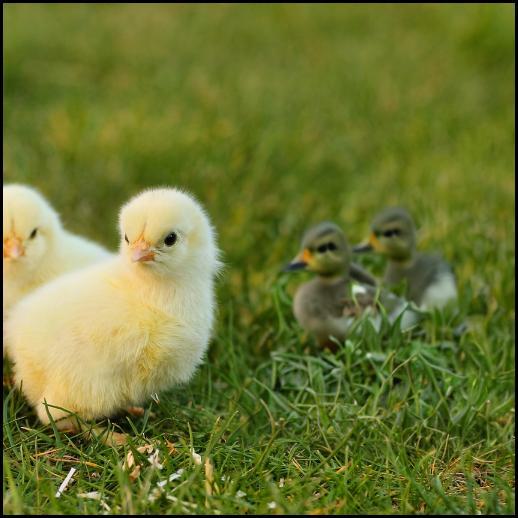} &
           \includegraphics[scale=0.18]{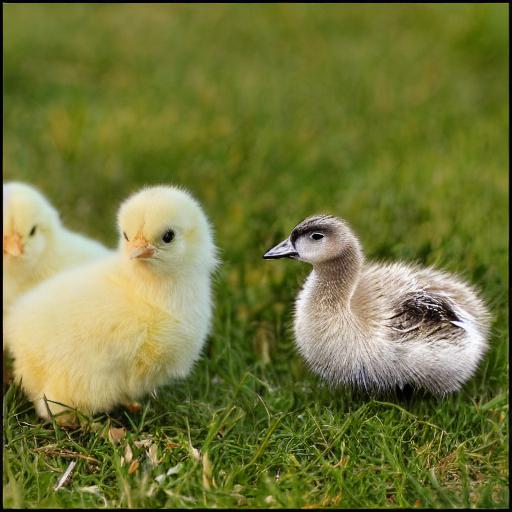} &
            \includegraphics[scale=0.18]{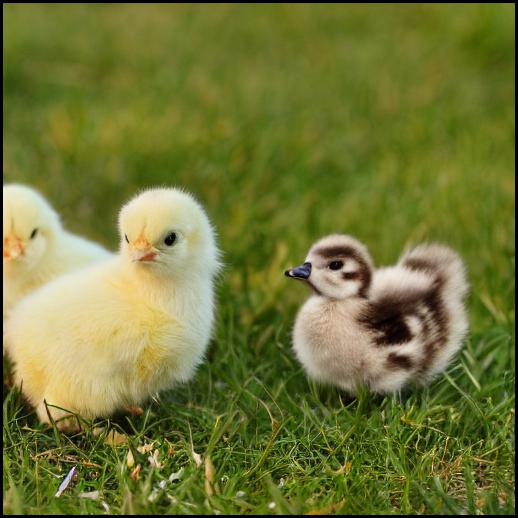} \\
             \includegraphics[scale=0.18]{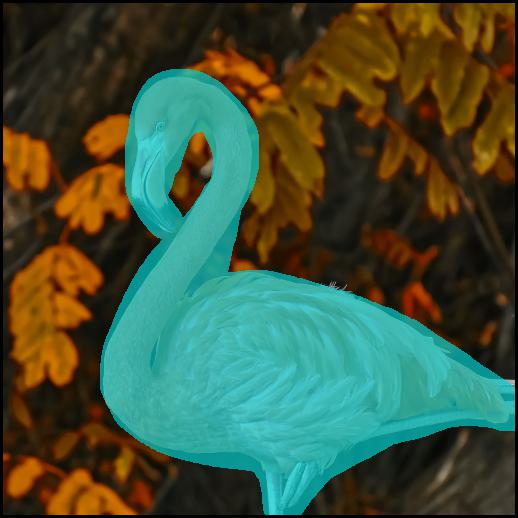} &
         \includegraphics[scale=0.18]{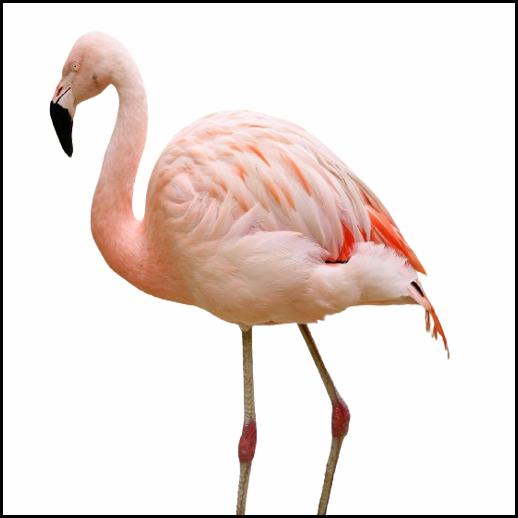} &
          \includegraphics[scale=0.18]{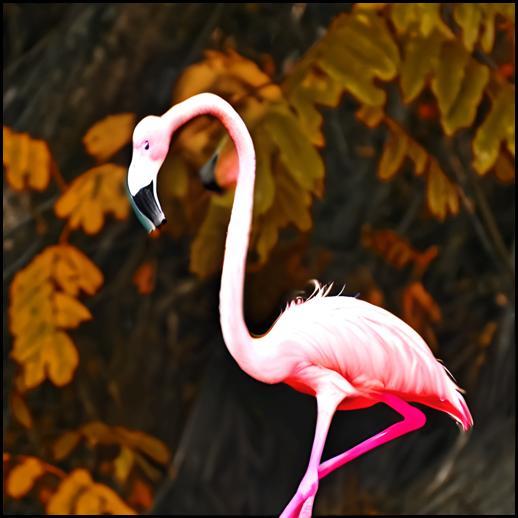} &
           \includegraphics[scale=0.18]{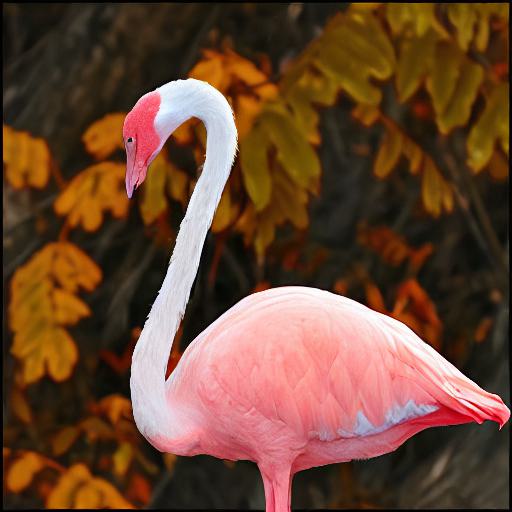} &
            \includegraphics[scale=0.18]{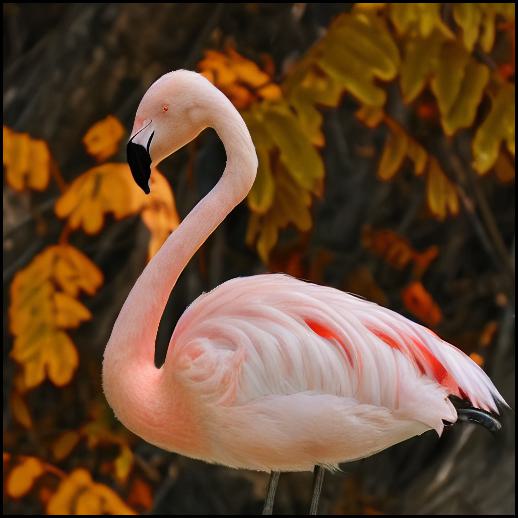} \\
              \includegraphics[scale=0.18]{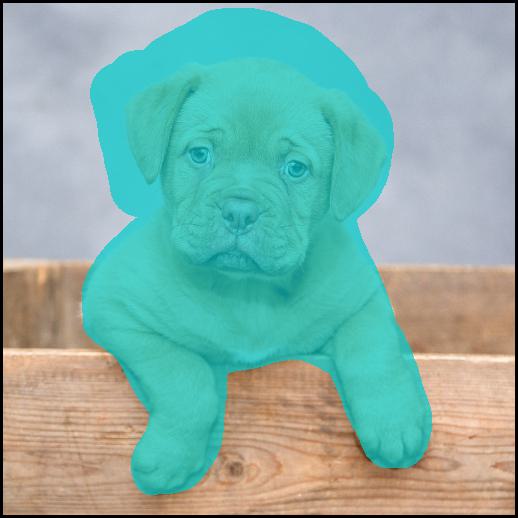} &
         \includegraphics[scale=0.18]{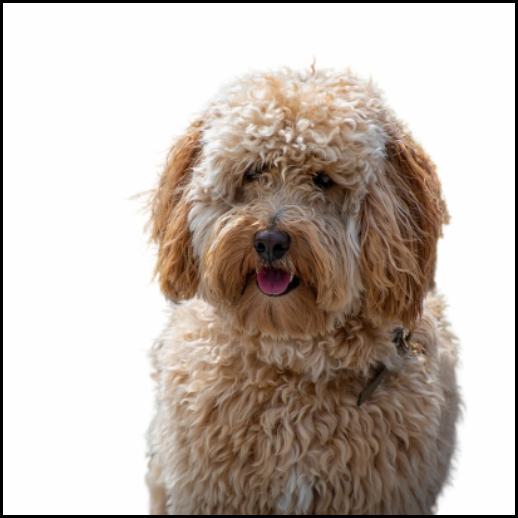} &
          \includegraphics[scale=0.18]{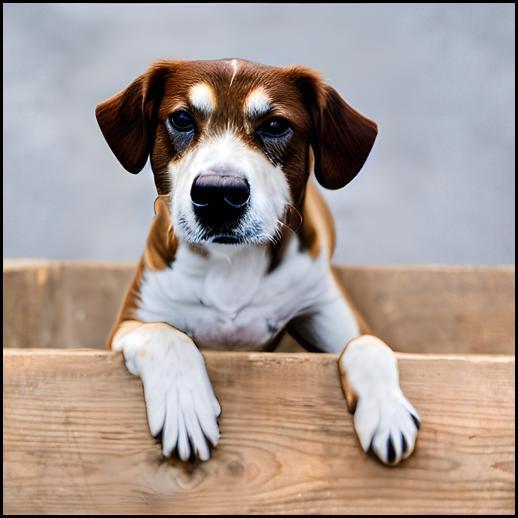} &
           \includegraphics[scale=0.18]{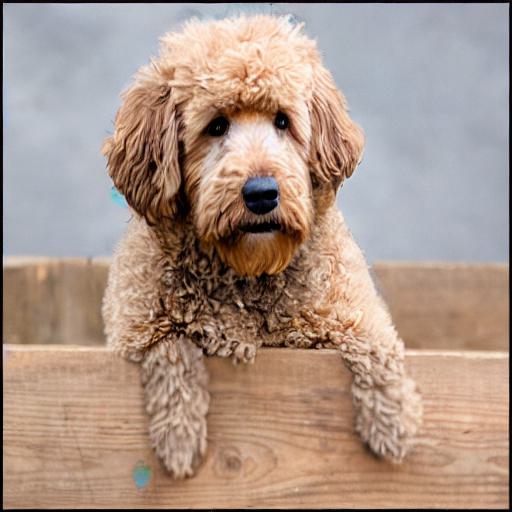} &
            \includegraphics[scale=0.18]{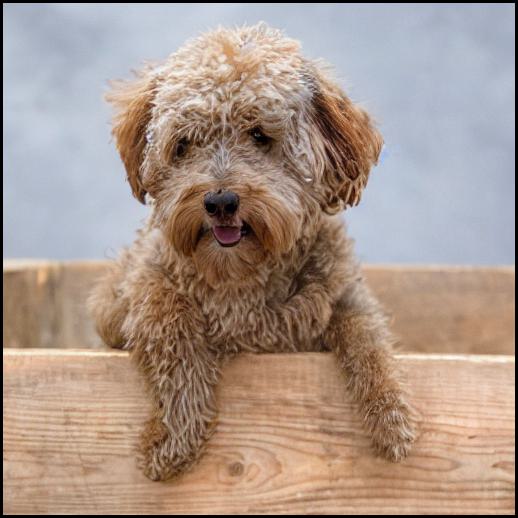} \\
    \end{tabular}
    }
    \caption{Image-guided inpainting \textbf{Part 2}.}
    \label{fig:supp_image2}
\end{figure*}

In this section, we provide more samples of the subject-driven image inpainting results in Fig. \ref{fig:supp_image1} and \ref{fig:supp_image2}. We are required to fill the missing region with the reference object while following the mask shape and preserving the identity of the reference object. We observe that our method can generates the reference object in the missing region while the strong baseline method PBE \citep{yang2023paint} fails to preserve the identity when the reference objects are complex, e.g., the beer and the candle in the first two rows in Fig. \ref{fig:supp_image1}. We argue that the identity lost could be caused by using only the CLIP class token in PBE. The CLIP class token has dimension 1024 and can only preserve the main information for classification and drops the details of the reference object, which is important for image-guided inpainting.  It is also worth noting that our method does not only learn to copy the reference object to the missing region. The third, fifth and sixth row that our method captures the high-level semantic information of the reference object and learns to change the pose (the sloth plushie), structure (the mushroom) and adding necessary content to make the output image more reaslistic (the penguine) thanks to our proposed token selection module. Otherwise, the model may only learns copy and paste, which needs to be avoided. The examples in Fig. \ref{fig:supp_image2} further demonstrate that our method is enpowered with such ability.

\section{Ablation}
\label{sec:supp_ablation}
\begin{figure*}
    \centering
    \setlength{\tabcolsep}{2pt}
    \begin{tabular}{cccccc}
         Reference & Token1 & Token2 &  Reference & Token1 & Token2\\
         \includegraphics[scale=0.15]{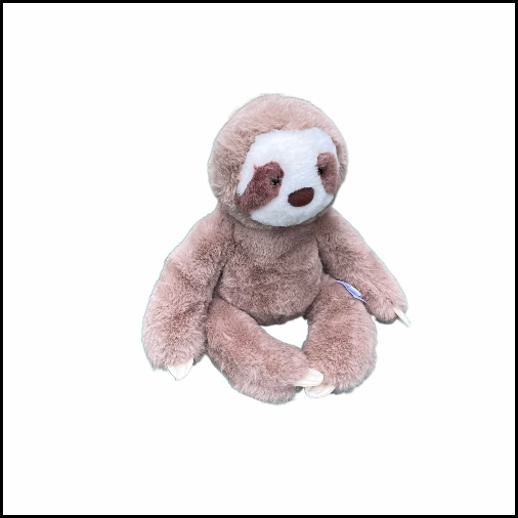}&
         \includegraphics[scale=0.15]{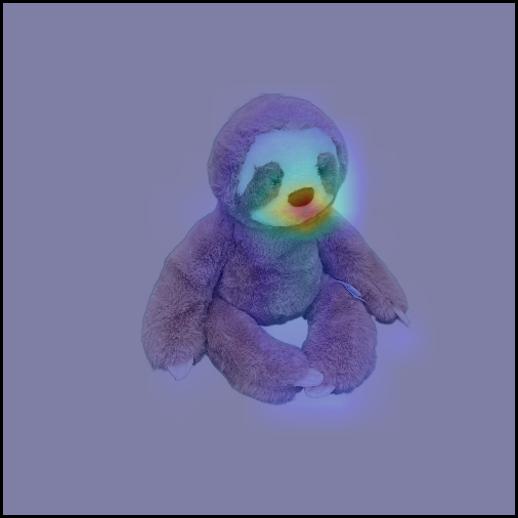}&
          \includegraphics[scale=0.15]{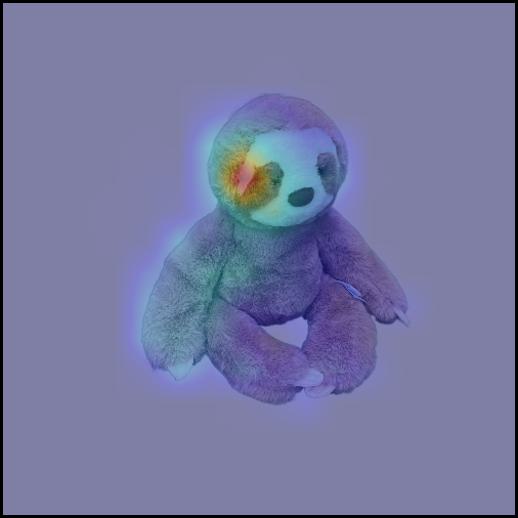}&
             \includegraphics[scale=0.15]{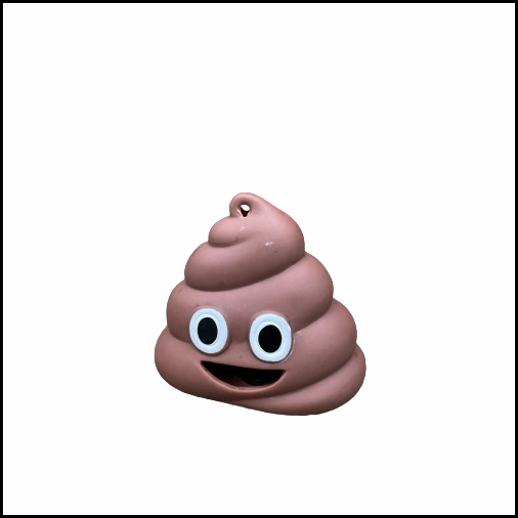}&
         \includegraphics[scale=0.15]{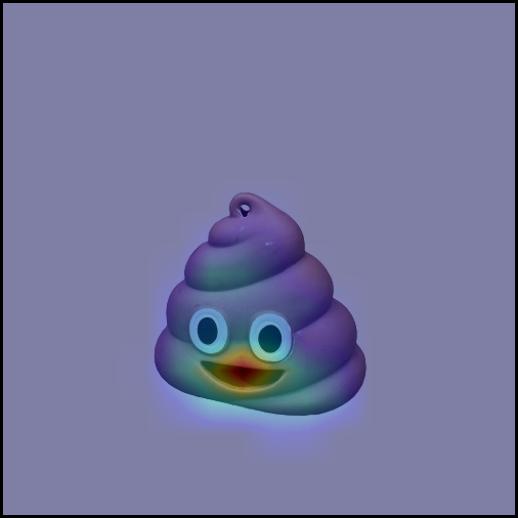}&
          \includegraphics[scale=0.15]{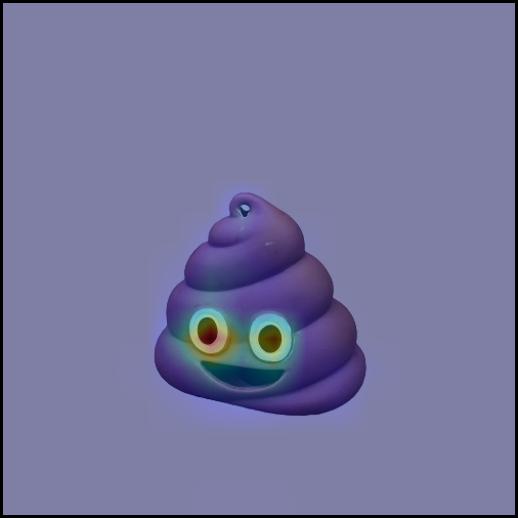}
         \\
           \includegraphics[scale=0.15]{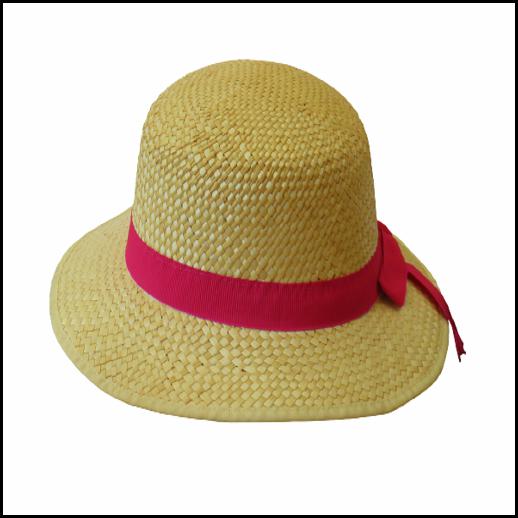}&
         \includegraphics[scale=0.15]{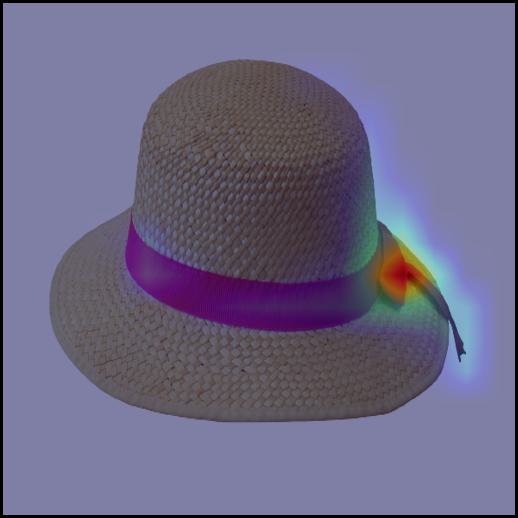}&
          \includegraphics[scale=0.15]{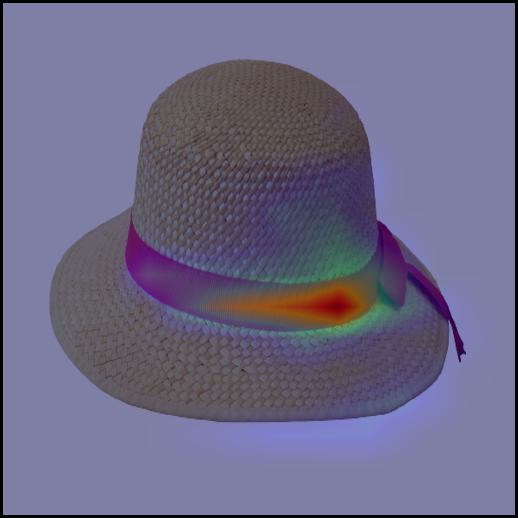}&
             \includegraphics[scale=0.15]{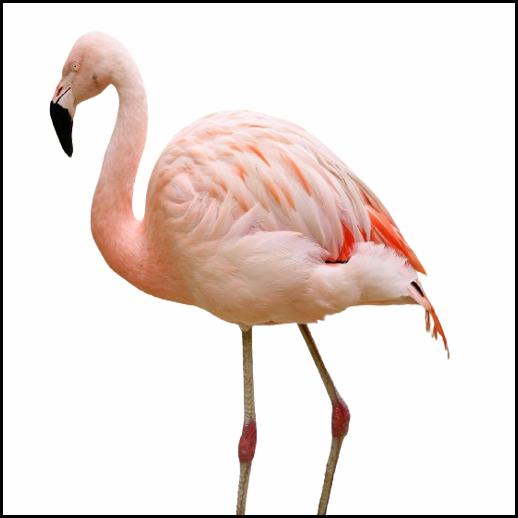}&
         \includegraphics[scale=0.15]{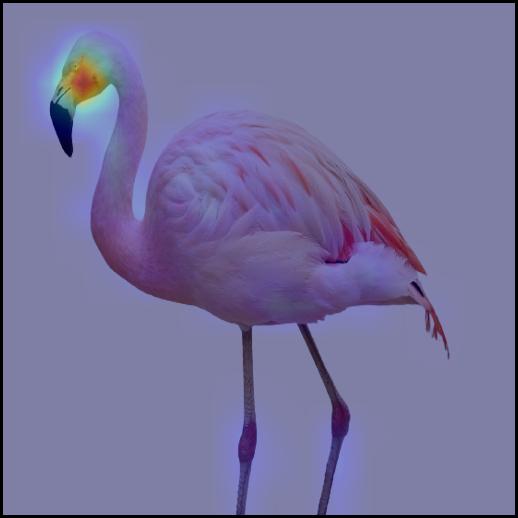}&
          \includegraphics[scale=0.15]{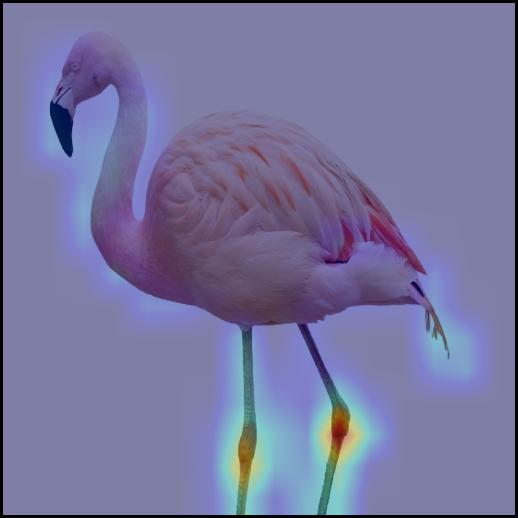}
         \\
            \includegraphics[scale=0.15]{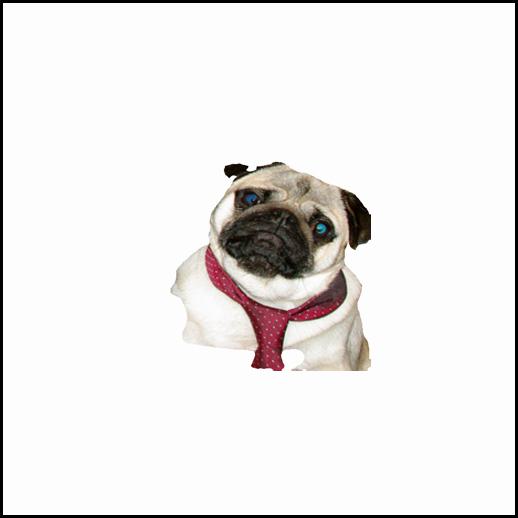}&
         \includegraphics[scale=0.15]{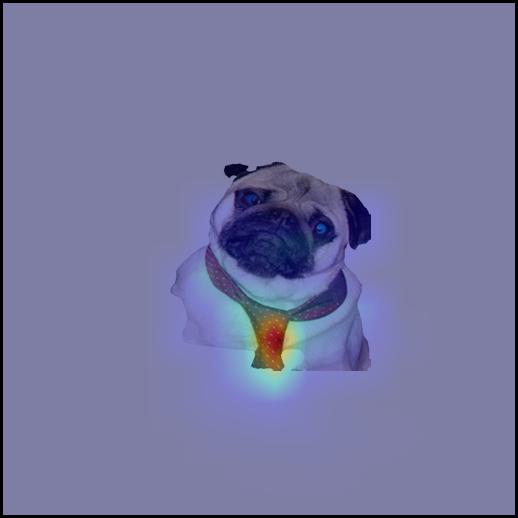}&
          \includegraphics[scale=0.15]{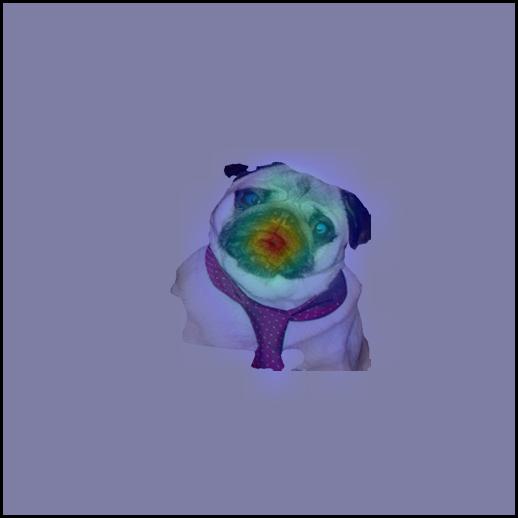}&
             \includegraphics[scale=0.15]{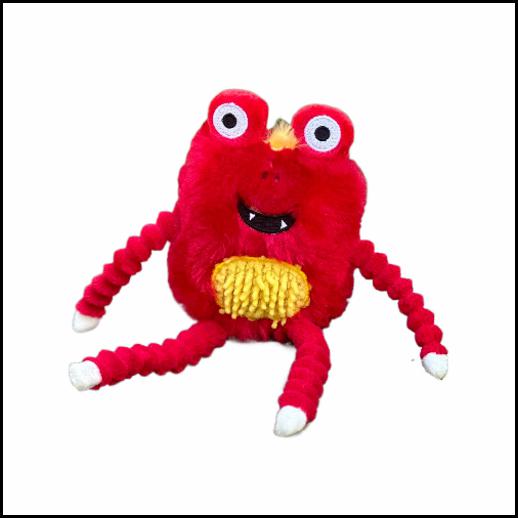}&
         \includegraphics[scale=0.15]{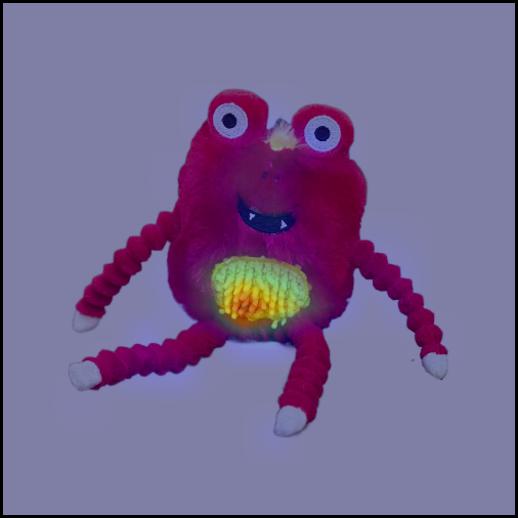}&
          \includegraphics[scale=0.15]{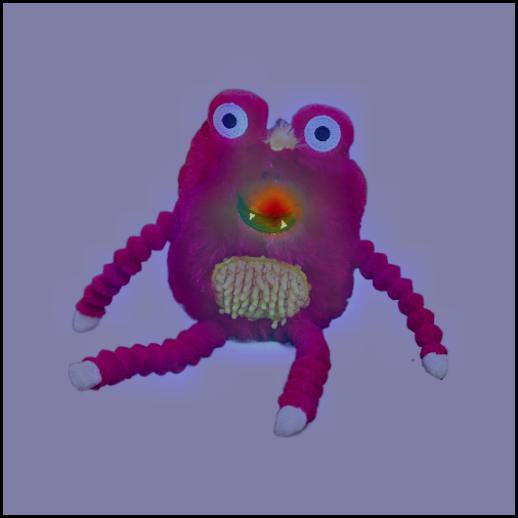}
         \\
          \includegraphics[scale=0.15]{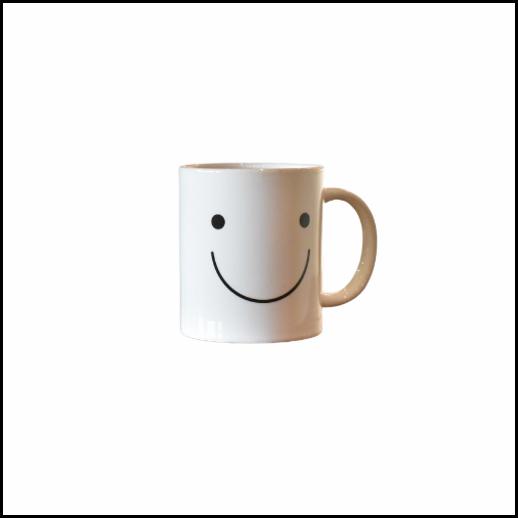}&
         \includegraphics[scale=0.15]{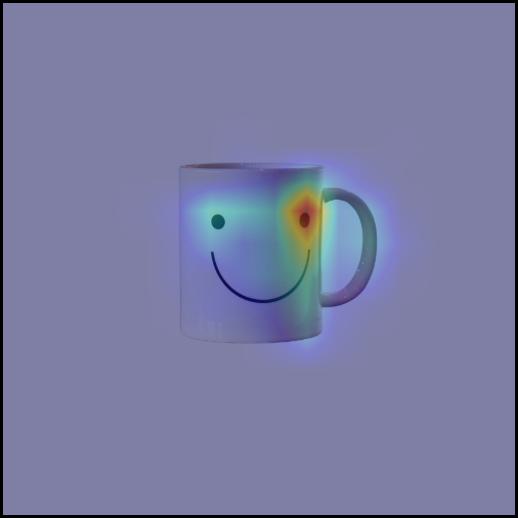}&
          \includegraphics[scale=0.15]{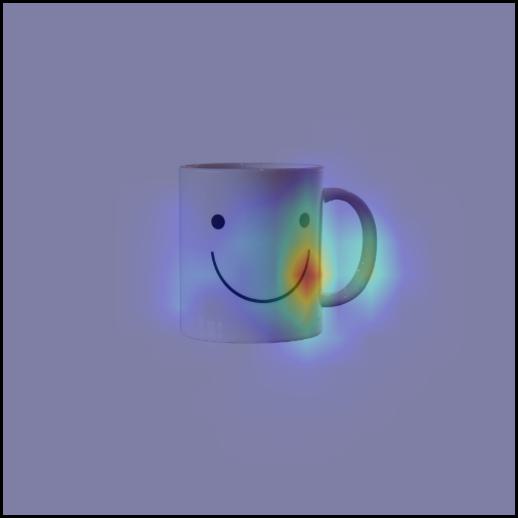}&
             \includegraphics[scale=0.15]{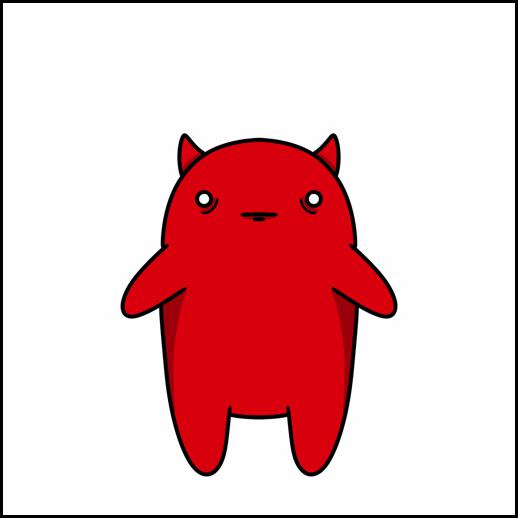}&
         \includegraphics[scale=0.15]{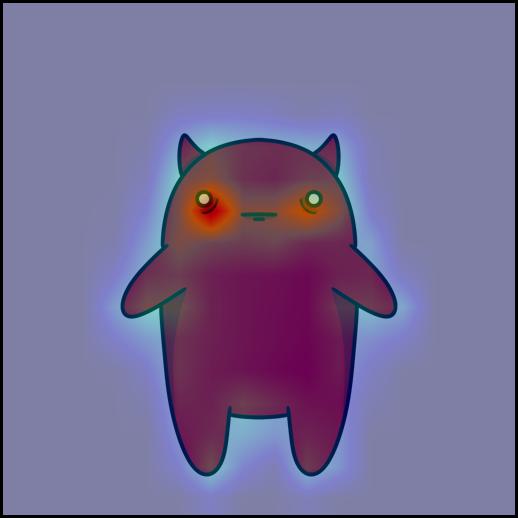}&
          \includegraphics[scale=0.15]{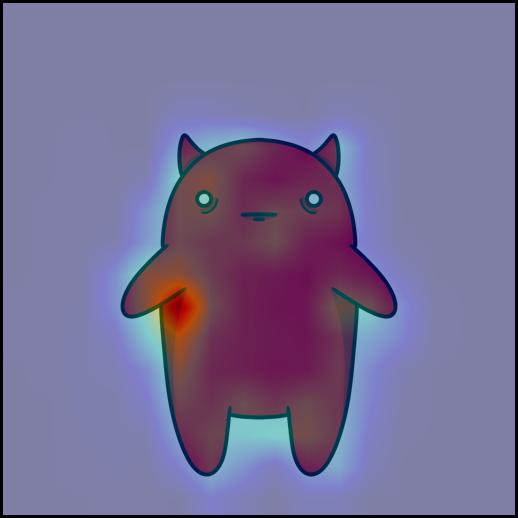}
         \\
          \includegraphics[scale=0.15]{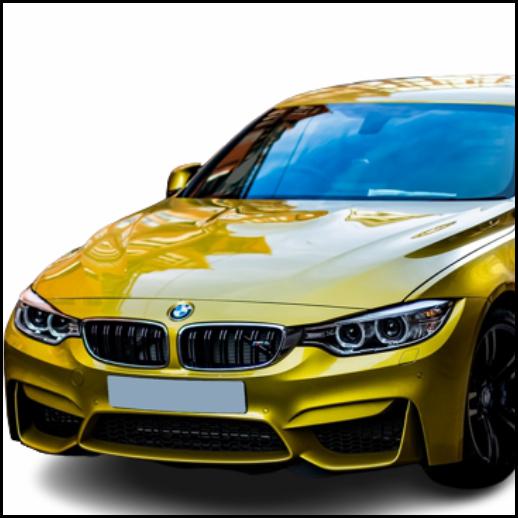}&
         \includegraphics[scale=0.15]{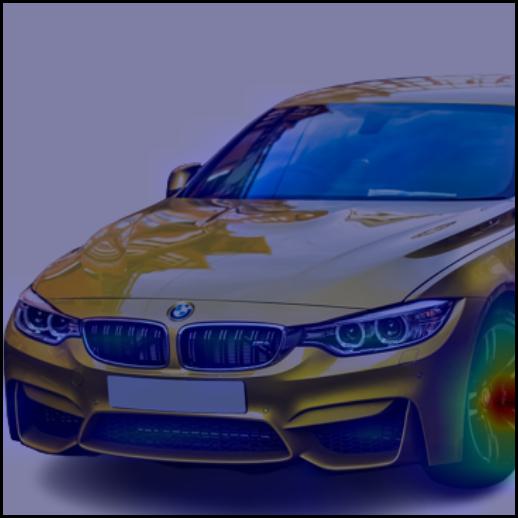}&
          \includegraphics[scale=0.15]{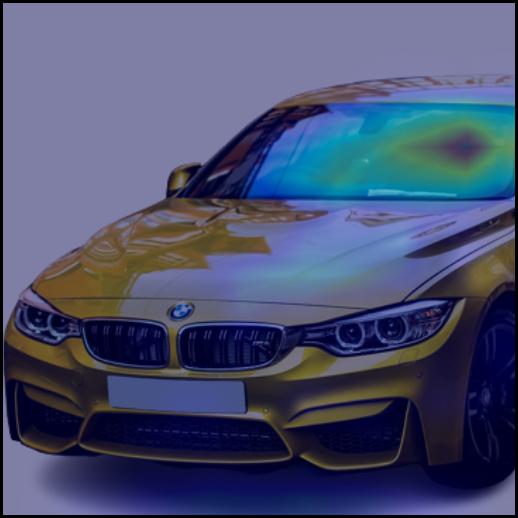}&
             \includegraphics[scale=0.15]{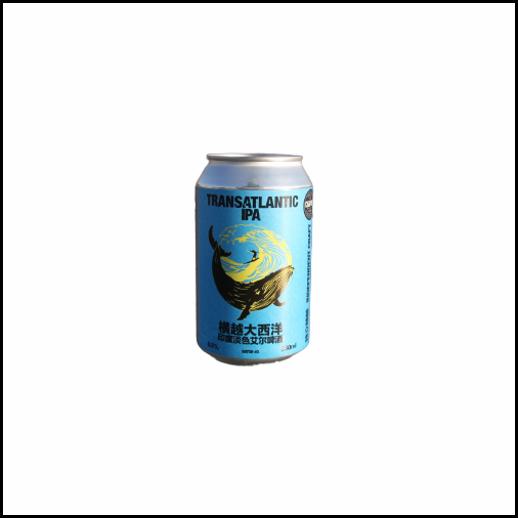}&
         \includegraphics[scale=0.15]{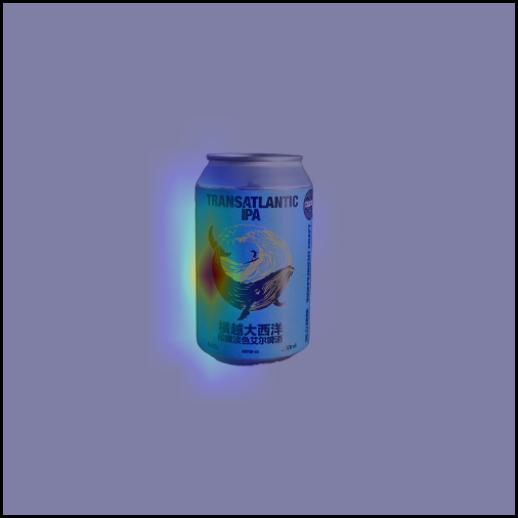}&
          \includegraphics[scale=0.15]{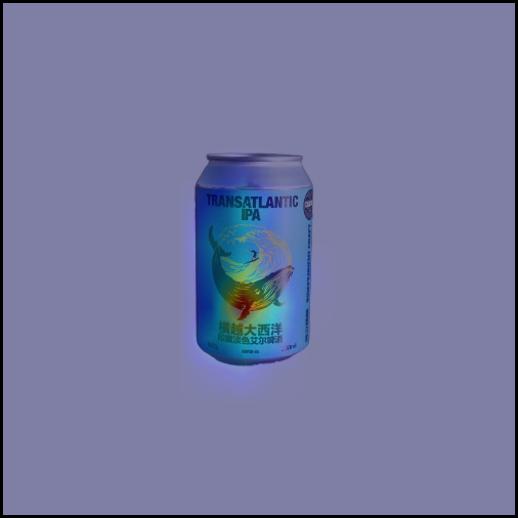}
         \\
          \includegraphics[scale=0.15]{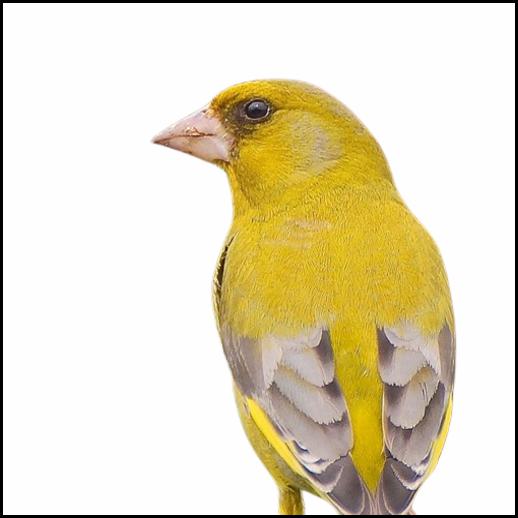}&
         \includegraphics[scale=0.15]{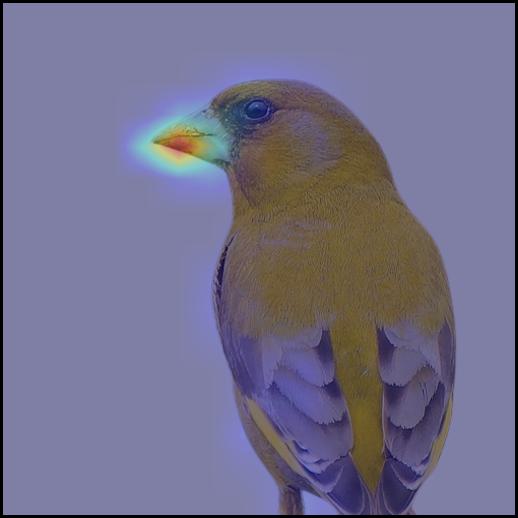}&
          \includegraphics[scale=0.15]{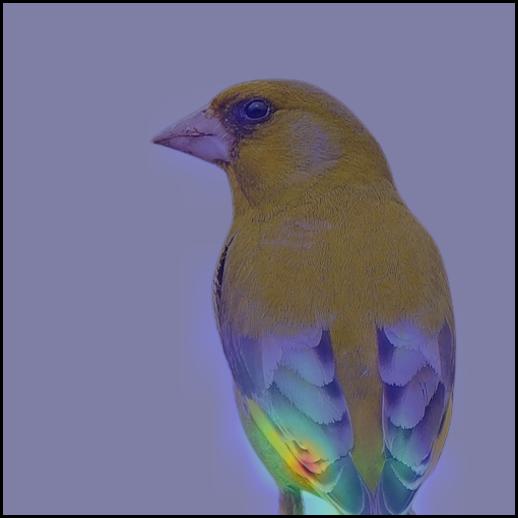}&
             \includegraphics[scale=0.15]{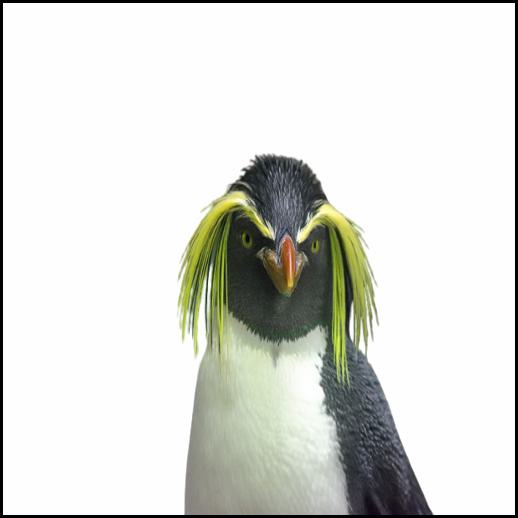}&
         \includegraphics[scale=0.15]{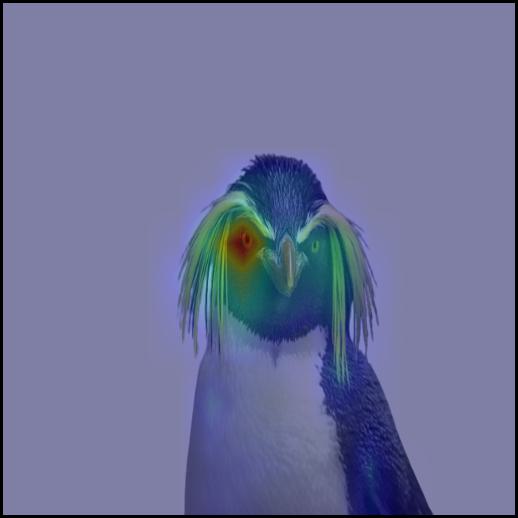}&
          \includegraphics[scale=0.15]{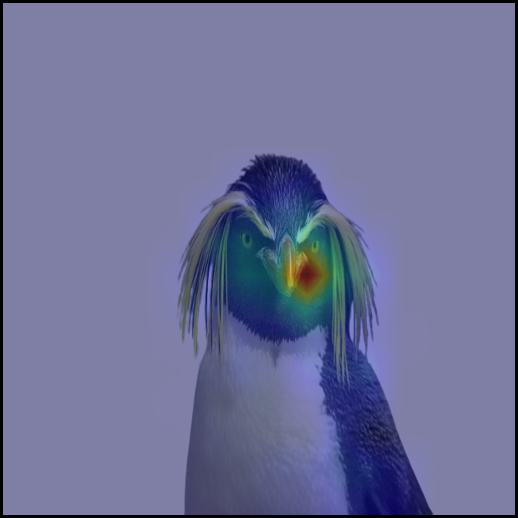}
         \\
          \includegraphics[scale=0.15]{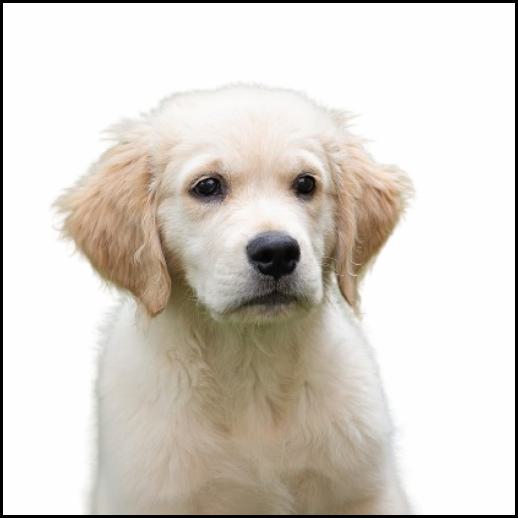}&
         \includegraphics[scale=0.15]{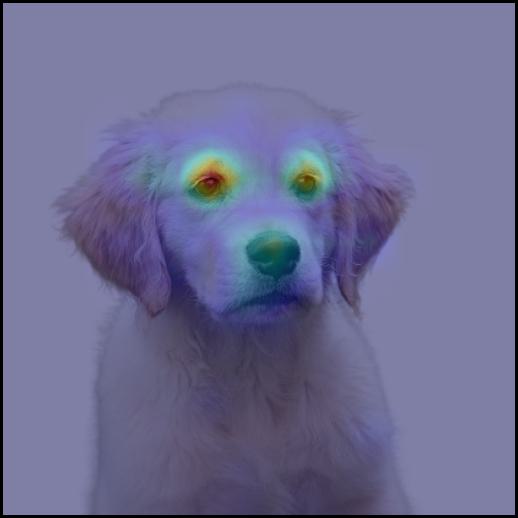}&
          \includegraphics[scale=0.15]{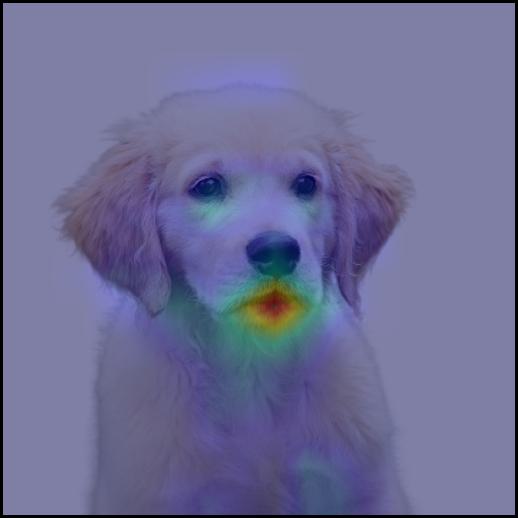}&
             \includegraphics[scale=0.15]{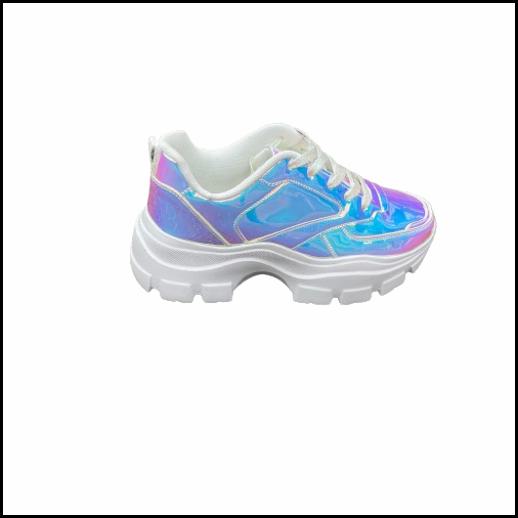}&
         \includegraphics[scale=0.15]{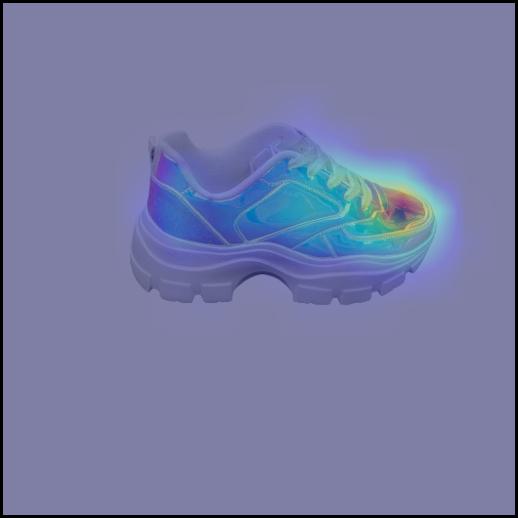}&
          \includegraphics[scale=0.15]{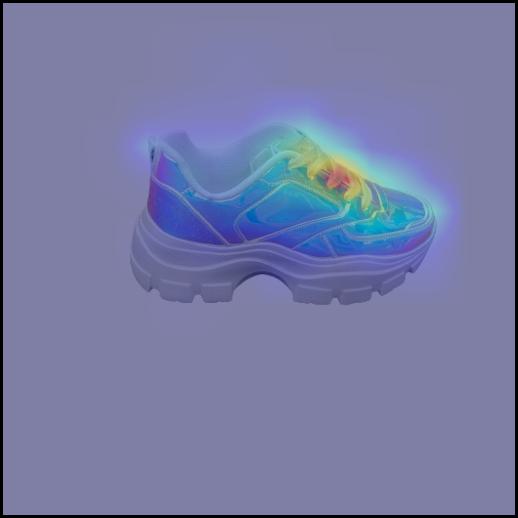}
         \\
    \end{tabular}
    \caption{Visualization of the selected tokens}
    \label{fig:supp_vis_token}
\end{figure*}

In this section, we provide the qualitative ablations of the proposed two modules: the token selection module and the decoupling regularization. We firstly show more visualization of the selected tokens. As indicated by Fig. \ref{fig:supp_vis_token}, our method extracts the most discriminative regions of the reference object, such the face of the sloth plushie and eyes of the poop emoji. This is essential and mainly explains the success of our subject-driven image inpainting method: we only need to preserve those discriminative tokens to preserve identity and avoids the copy-paste artifacts by using the high-dimensional redundant features. 

Then we present more examples of the generations versus different number of selected tokens. As mentioned in the main paper, too few tokens will lead to identity lost while too many tokens tends to copy the reference object. The examples in Fig. \ref{fig:supp_topk} well supports our argument. For simple objects such as the smile ballon and the teapot, we are able to get satisfactory results with $K\geq 8$. However, we may lose details for more complex objects, such as the crab in the last row by setting $K=8$ or 16. However, if we set $K$ too high, it tends to utilize some unnecessary information, causing the output image unrealistic. For instance, when we set $K=48$, the generated sloth plushie in the fourth row copies the sitting pose of the reference object and fails to align with the input mask. Generally, our mthod is quite robust versus with the values of $K$ but $K=24$ works best across different scenarios.

Finally, we show examples of the proposed decoupling regularization. Without the proposed regularization, the model ignores the text prompt sometimes, e.g., the generations are almost identical for the three different text prompts containing in the second row. In some other cases, the model may generates the content according to the text prompt but loses more identity information about the reference object. For instance, the generated cube ballon containing blue grids in the fourth row, which makes it very different from the reference ballon. In the shoes example, the model may generate slight changes according to the text prompt but fails to exactly follow the text prompt such as the boots example. These evidences indicate that the proposed decoupling regularization is essential for the text-guided subject-driven image inpainting results.

\section{Limitation of our method}
\label{sec:supp_limit}

\begin{figure*}
    \centering
     \setlength{\tabcolsep}{2pt}
     \scalebox{0.925}{
    \begin{tabular}{cccccccc}
    Input & Reference & K=1 & K=4 & K=8 & K=16 & K=24 & K=48 \\
       \includegraphics[scale=0.12]{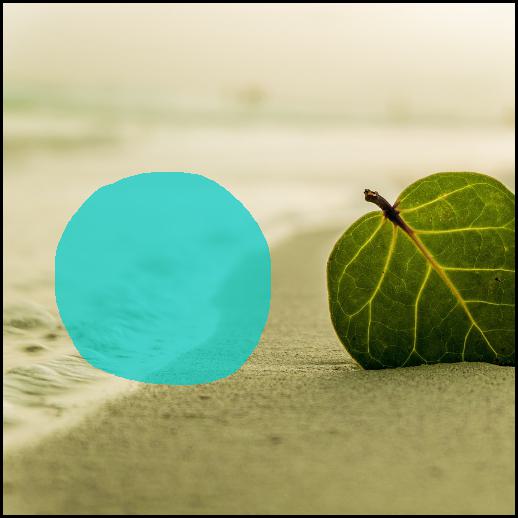} & 
       \includegraphics[scale=0.12]{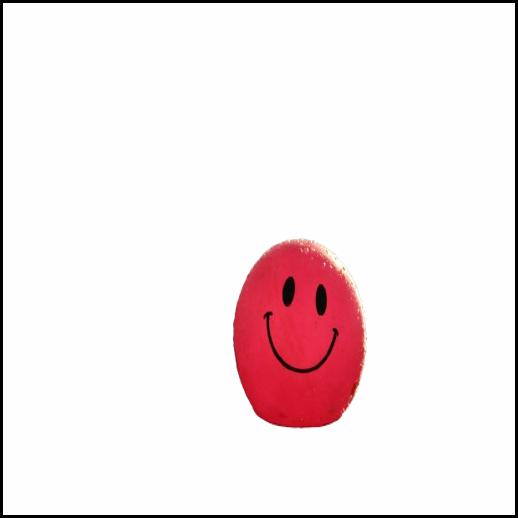} & 
       \includegraphics[scale=0.12]{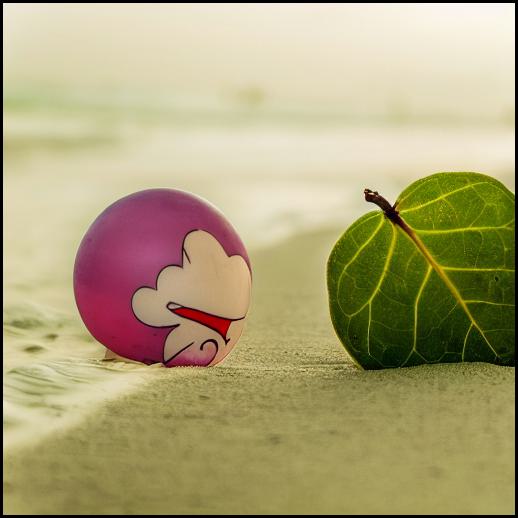} & 
       \includegraphics[scale=0.12]{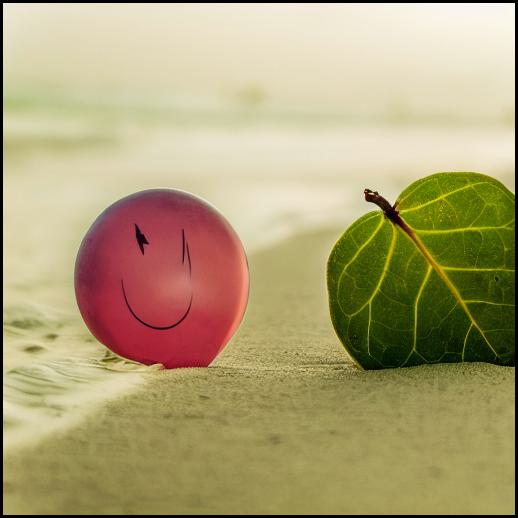} & 
       \includegraphics[scale=0.12]{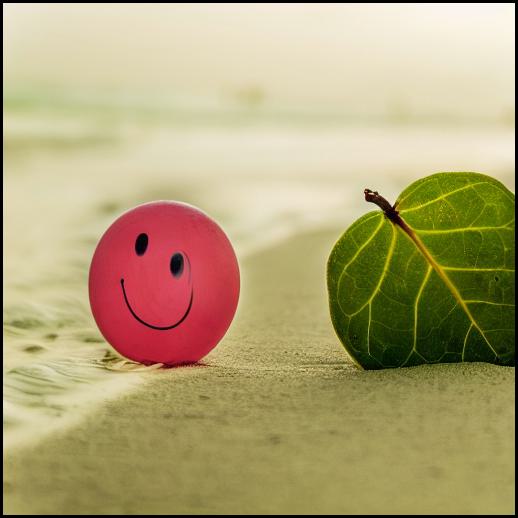} & 
       \includegraphics[scale=0.12]{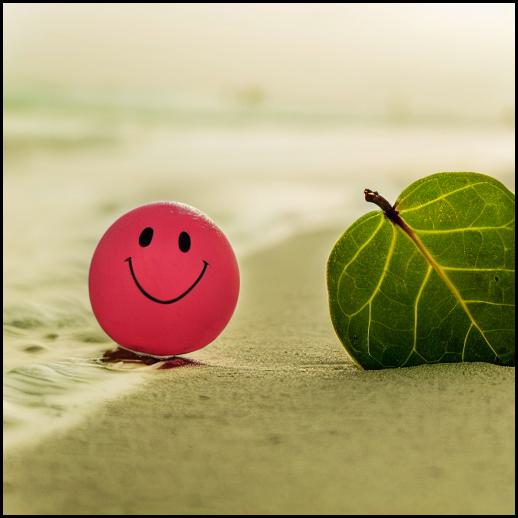} & 
       \includegraphics[scale=0.12]{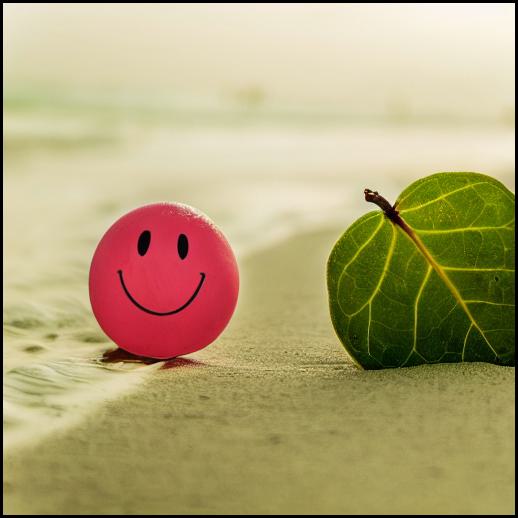} & 
       \includegraphics[scale=0.12]{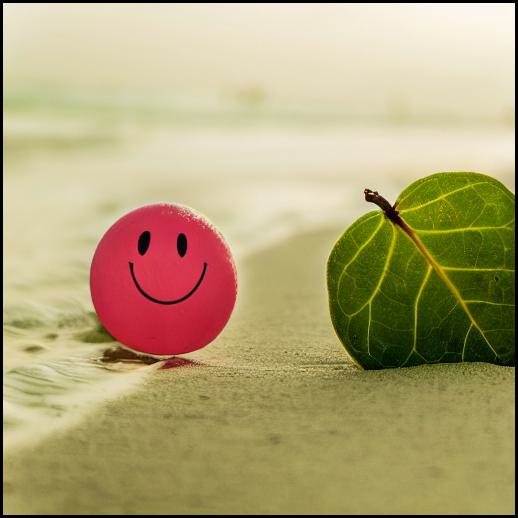} \\
         \includegraphics[scale=0.12]{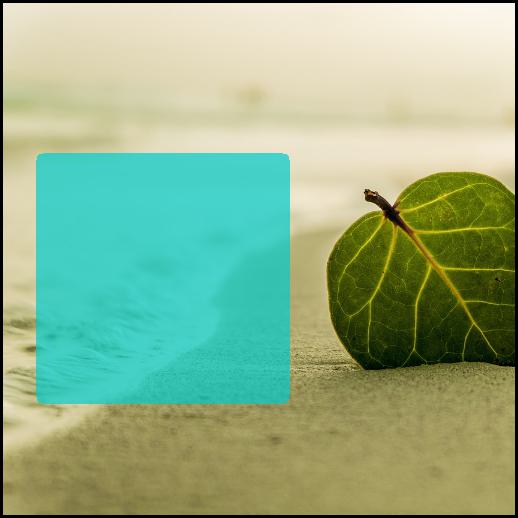} & 
       \includegraphics[scale=0.12]{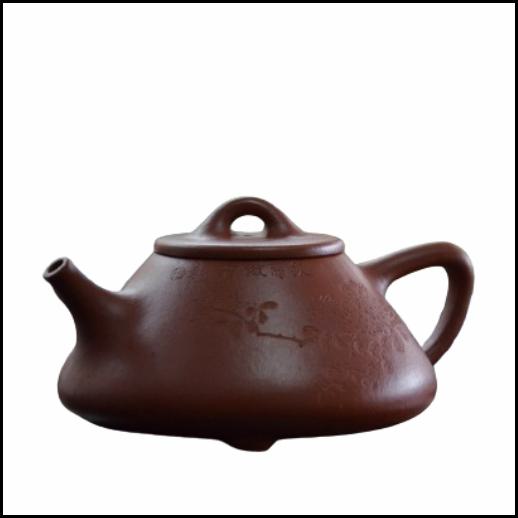} & 
       \includegraphics[scale=0.12]{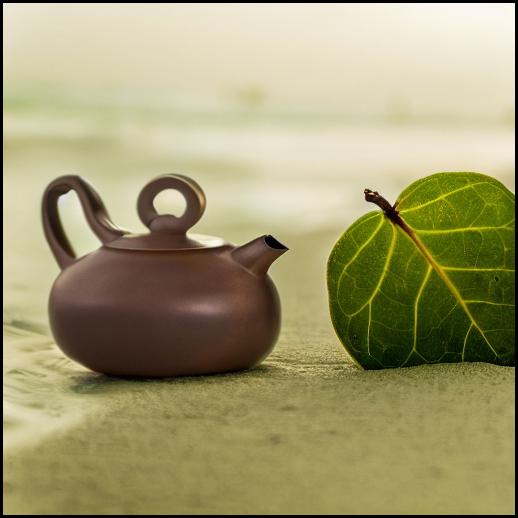} & 
       \includegraphics[scale=0.12]{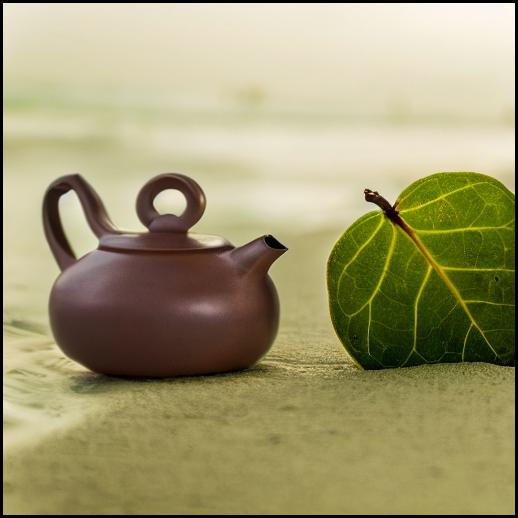} & 
       \includegraphics[scale=0.12]{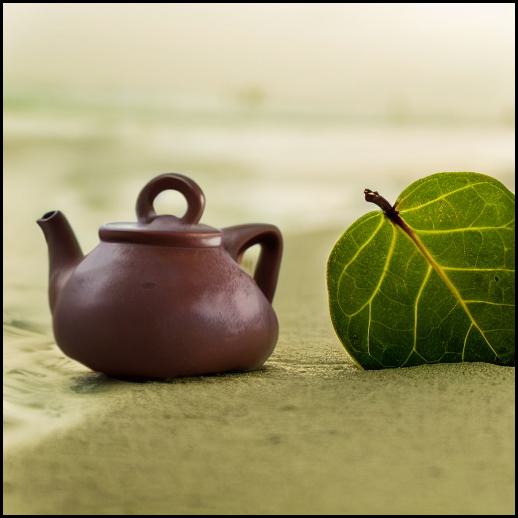} & 
       \includegraphics[scale=0.12]{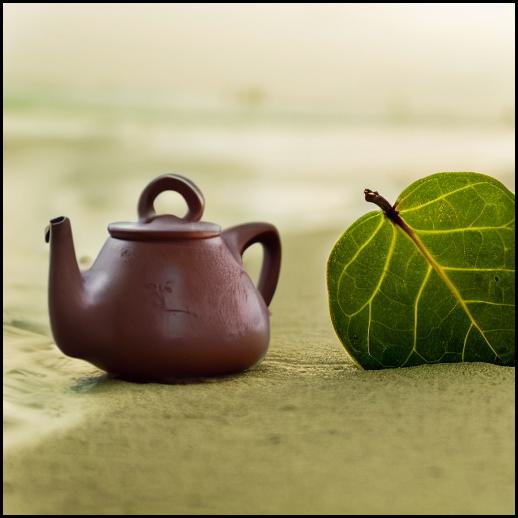} & 
       \includegraphics[scale=0.12]{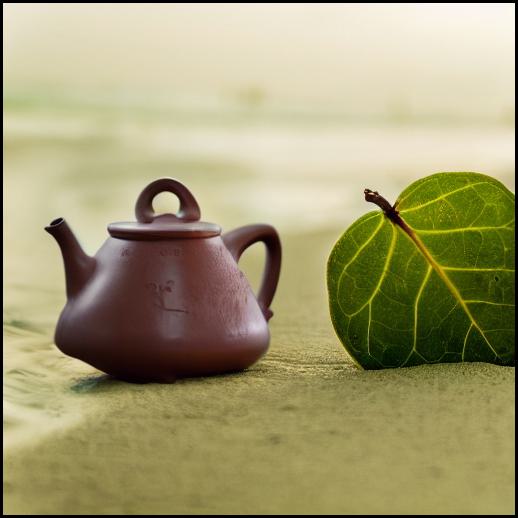} & 
       \includegraphics[scale=0.12]{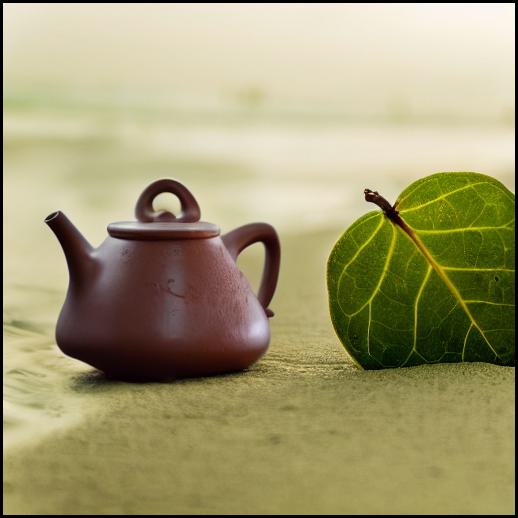} \\
           \includegraphics[scale=0.12]{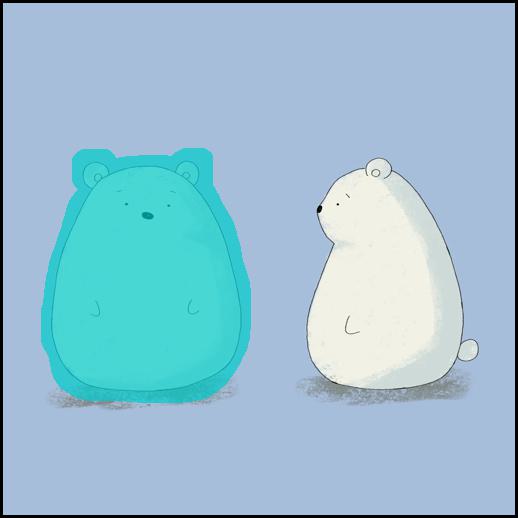} & 
       \includegraphics[scale=0.12]{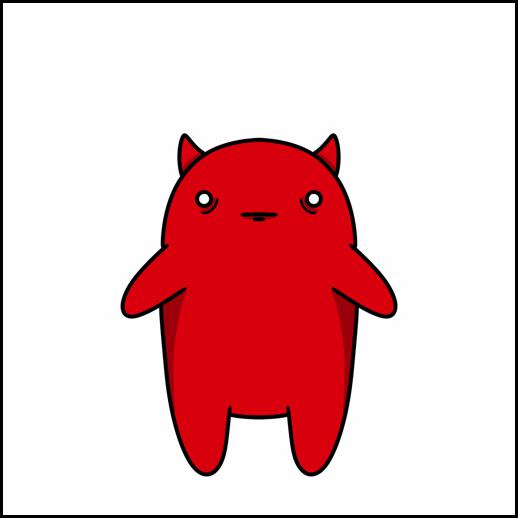} & 
       \includegraphics[scale=0.12]{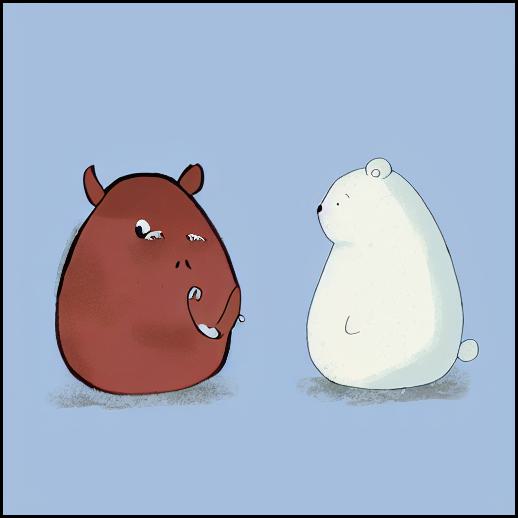} & 
       \includegraphics[scale=0.12]{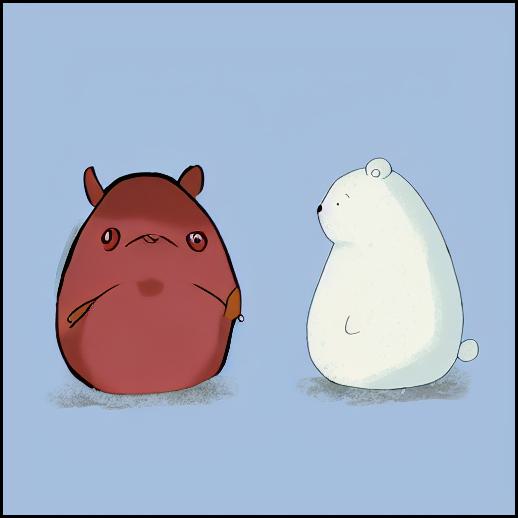} & 
       \includegraphics[scale=0.12]{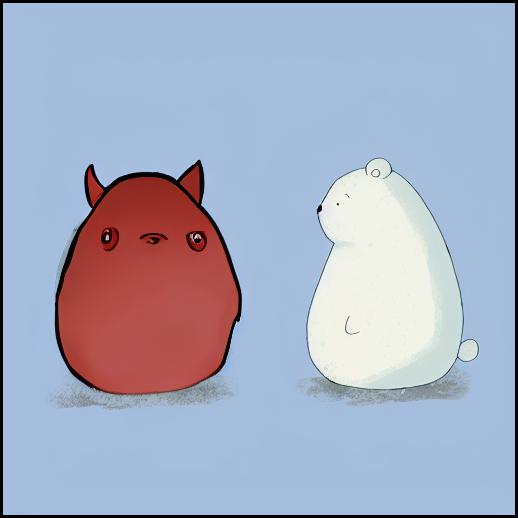} & 
       \includegraphics[scale=0.12]{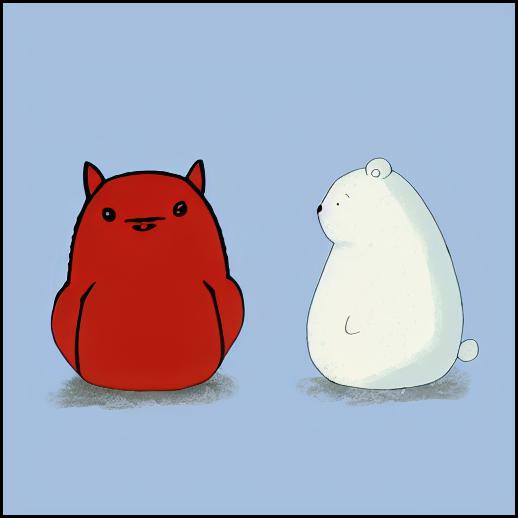} & 
       \includegraphics[scale=0.12]{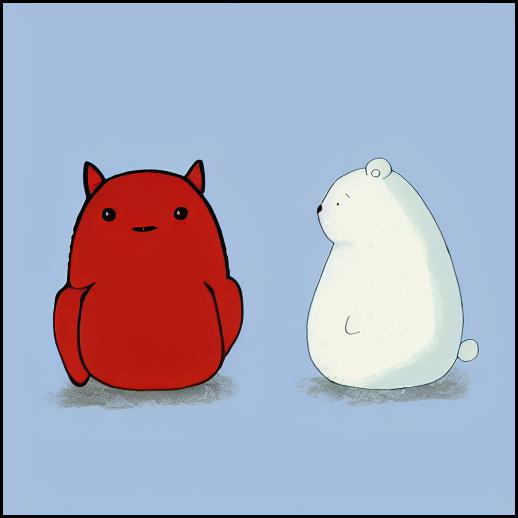} & 
       \includegraphics[scale=0.12]{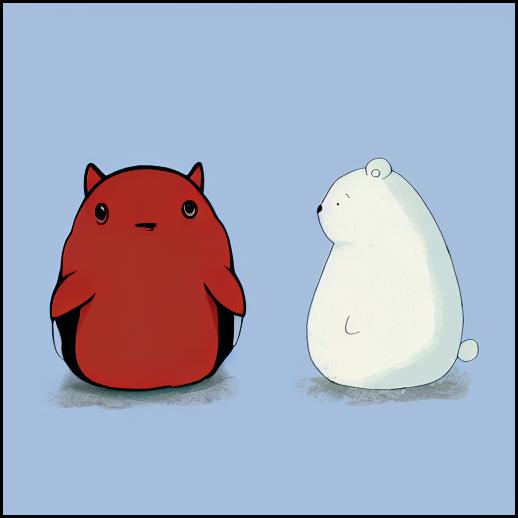} \\
            \includegraphics[scale=0.12]{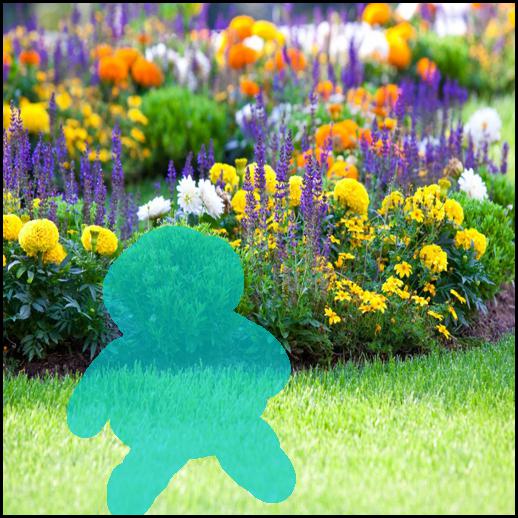} & 
       \includegraphics[scale=0.12]{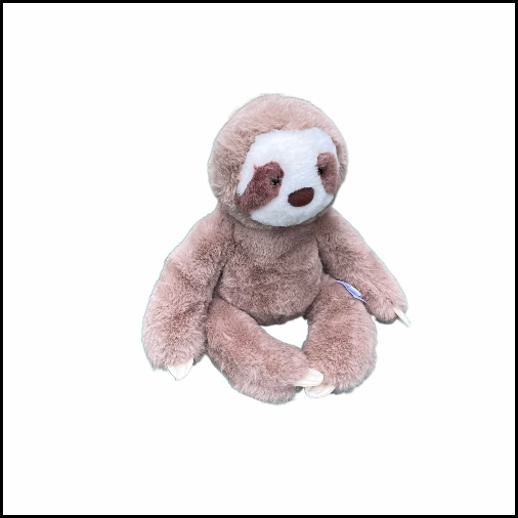} & 
       \includegraphics[scale=0.12]{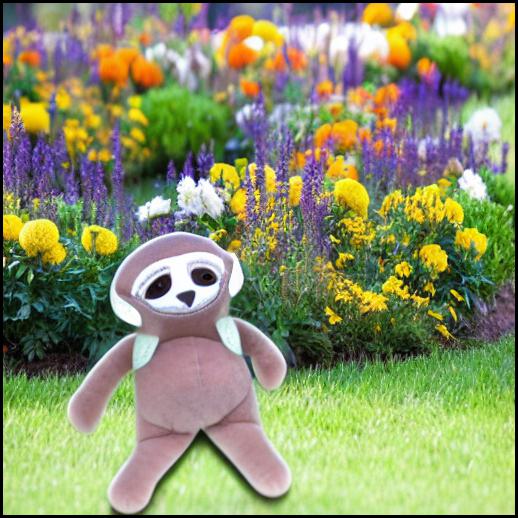} & 
       \includegraphics[scale=0.12]{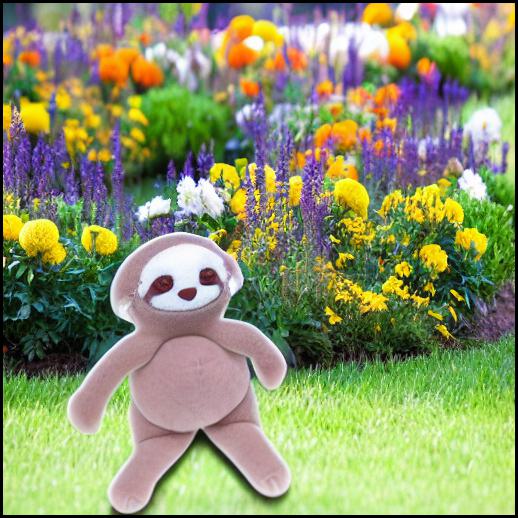} & 
       \includegraphics[scale=0.12]{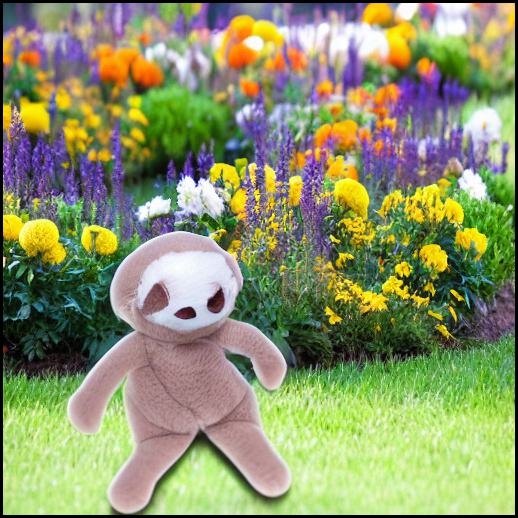} & 
       \includegraphics[scale=0.12]{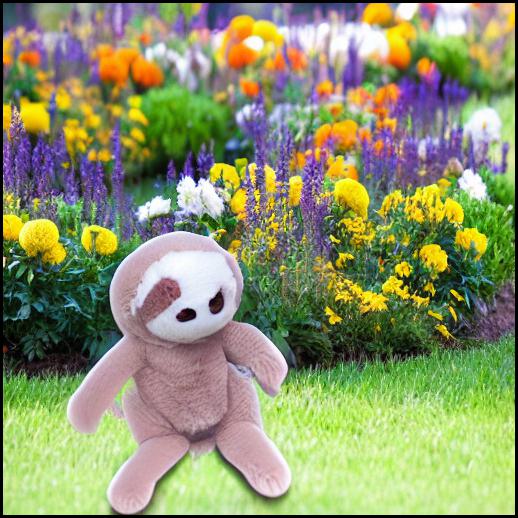} & 
       \includegraphics[scale=0.12]{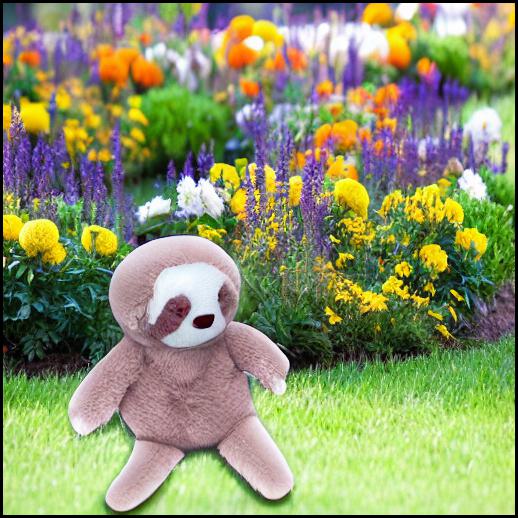} & 
       \includegraphics[scale=0.12]{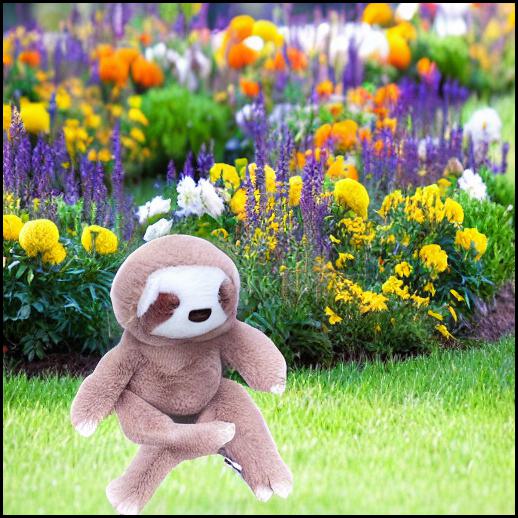} \\
         \includegraphics[scale=0.12]{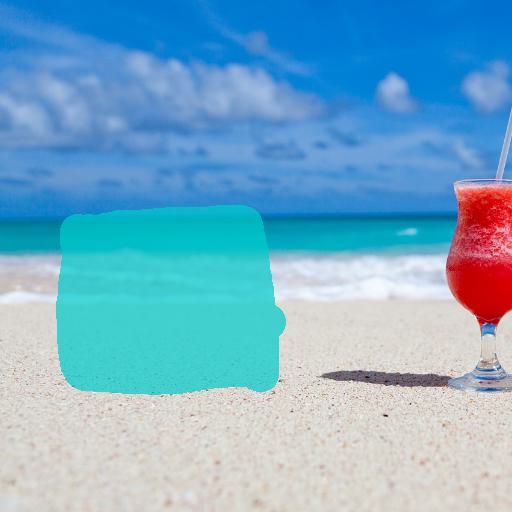} & 
       \includegraphics[scale=0.12]{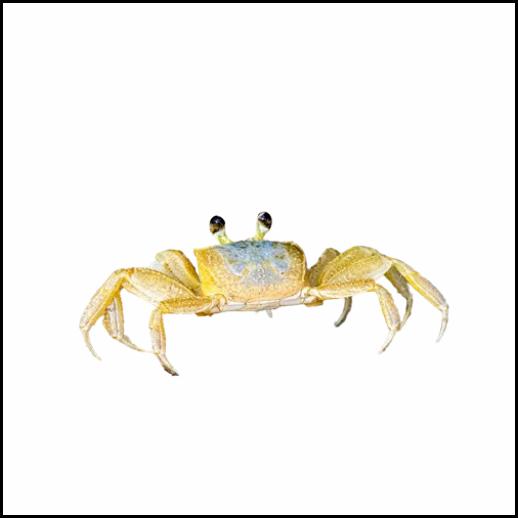} & 
       \includegraphics[scale=0.12]{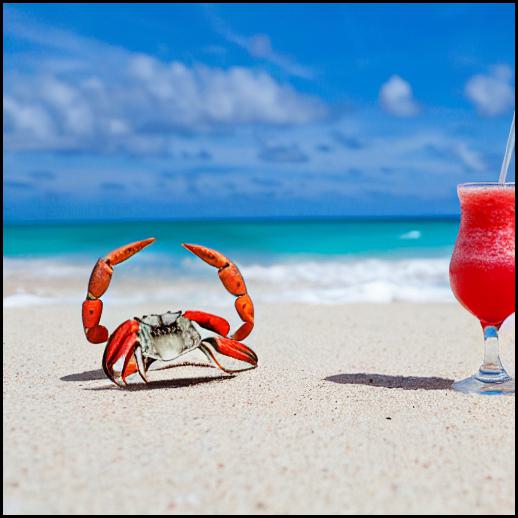} & 
       \includegraphics[scale=0.12]{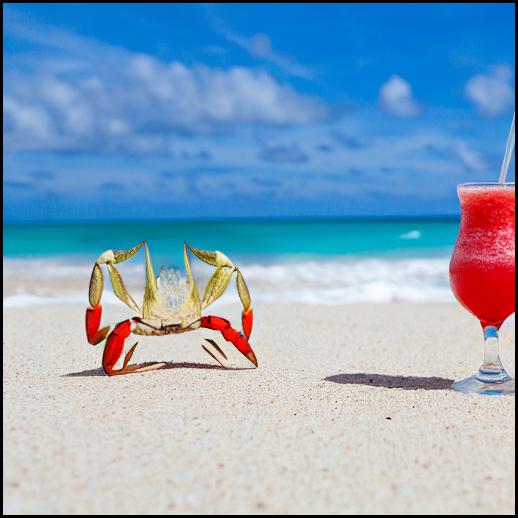} & 
       \includegraphics[scale=0.12]{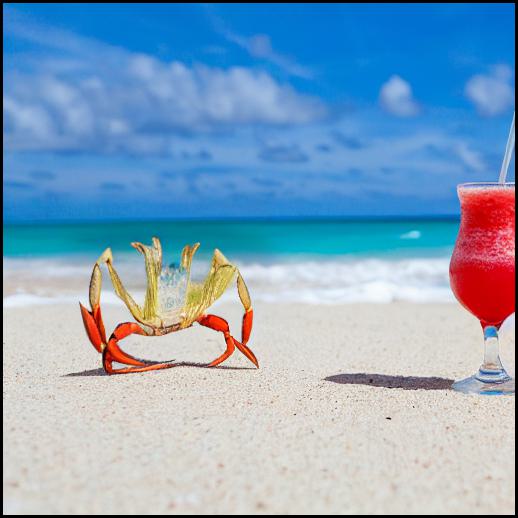} & 
       \includegraphics[scale=0.12]{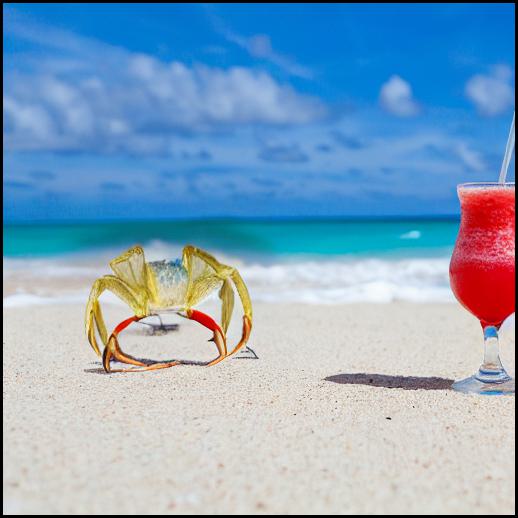} & 
       \includegraphics[scale=0.12]{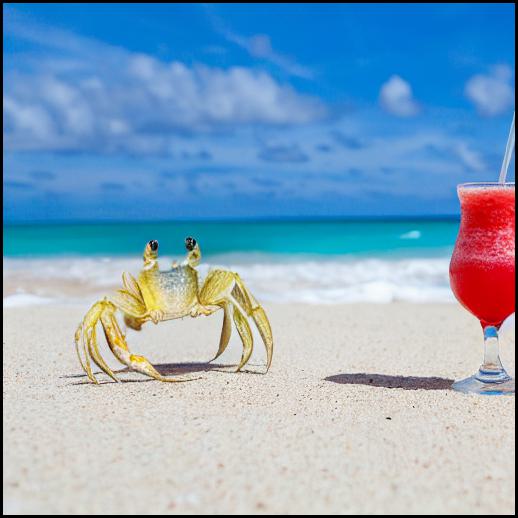} & 
       \includegraphics[scale=0.12]{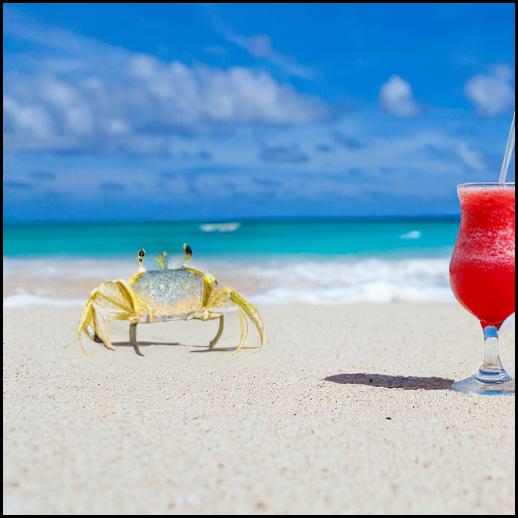} \\
    \end{tabular}
    }
    \caption{The inpaintingr results versus different numbers of selected tokens.}
    \label{fig:supp_topk}
\end{figure*}

\begin{figure*}
    \centering
    \begin{tabular}{cccc}
    Input & Ref & PBE \citep{yang2023paint} & Ours \\
        \includegraphics[scale=0.17]{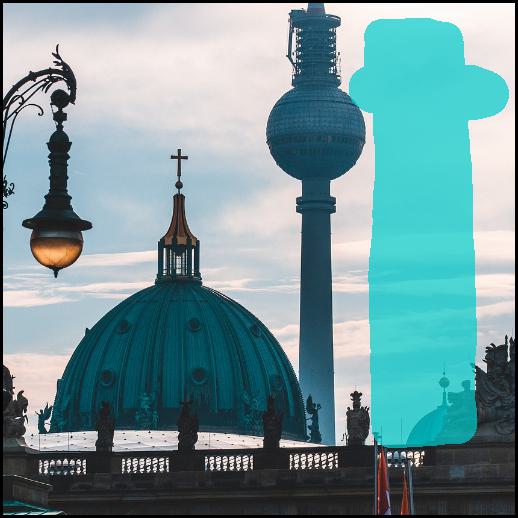} & 
        \includegraphics[scale=0.17]{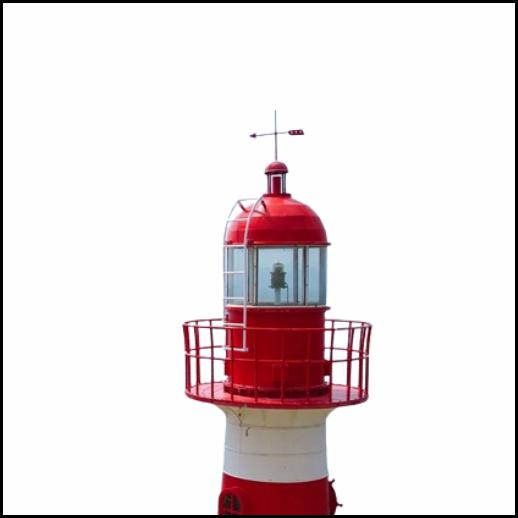} & 
        \includegraphics[scale=0.17]{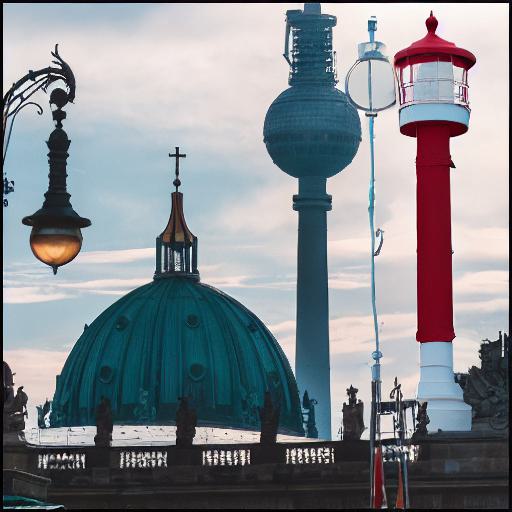} & 
        \includegraphics[scale=0.17]{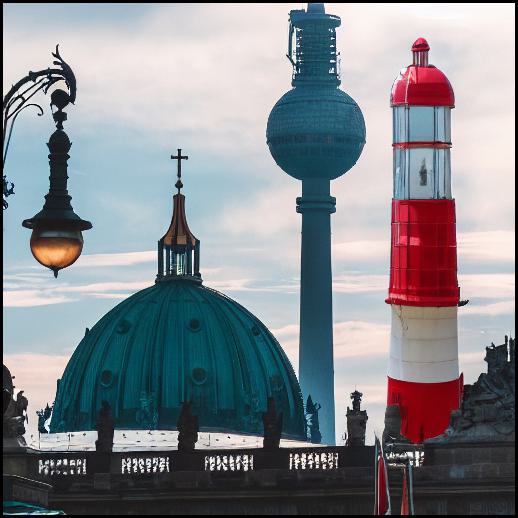} 
        \\
         \includegraphics[scale=0.17]{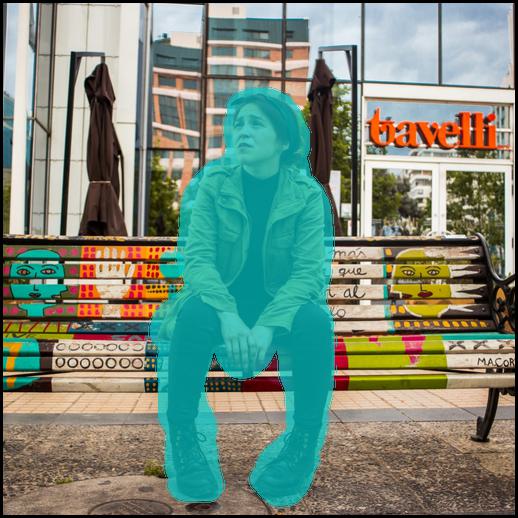} & 
        \includegraphics[scale=0.17]{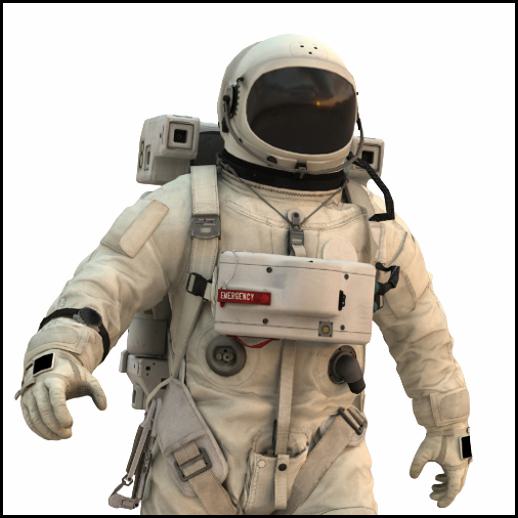} & 
        \includegraphics[scale=0.17]{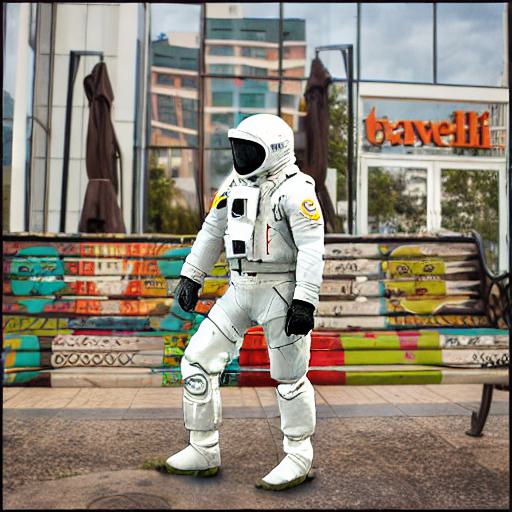} & 
        \includegraphics[scale=0.17]{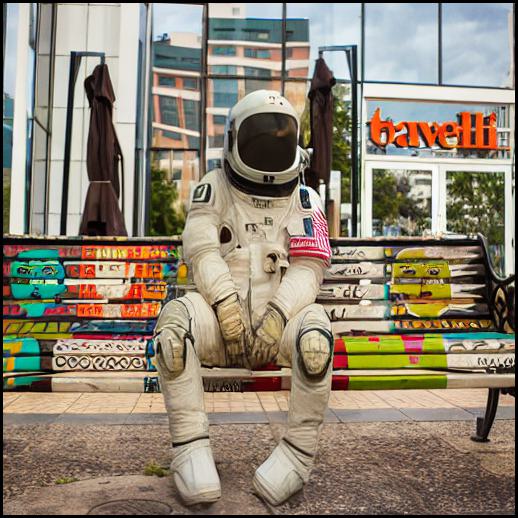} 
        \\
         \includegraphics[scale=0.17]{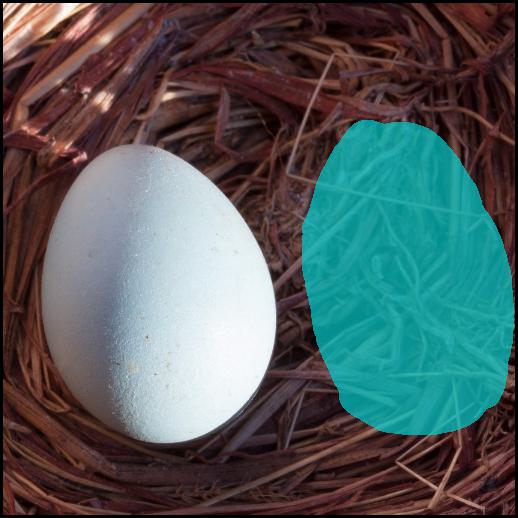} & 
        \includegraphics[scale=0.17]{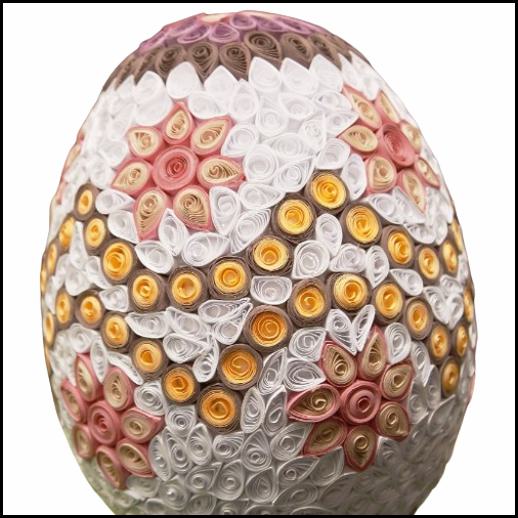} & 
        \includegraphics[scale=0.17]{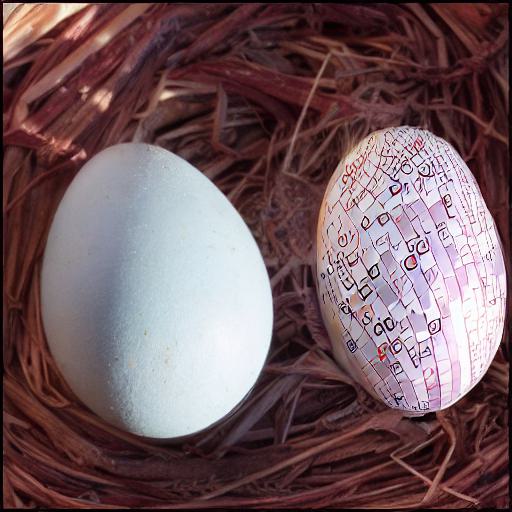} & 
        \includegraphics[scale=0.17]{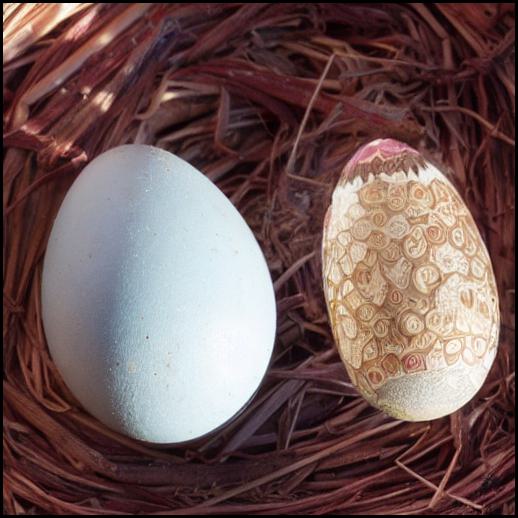} 
        \\
         & 
    \end{tabular}
    \caption{Limitation of our method.}
    \label{fig:supp_limit}
\end{figure*}

As mentioned in the last section of the main paper, our method still suffers the identity lost problem when all details of the reference object are need. As shown in Fig. \ref{fig:supp_limit}, our method learns to generate the tower, astronaut, and easter egg in the missing region. Although our method preserves better identity than the strong baseline \citep{yang2023paint}, it still fails to preserve the details, such as the box in front of the astronaut and the curve patterns of the easter egg. Using more tokens can address this identity issue since more tokens contain more information. However, more tokens may lead to copy-paste artifacts as demonstrated in Fig. \ref{fig:supp_topk}. A potential solution may be using different number of tokens during training. Then the method learns to extract different volume of information from the reference image. However, it can be difficult to learn to decide the number of tokens for each reference object since no ground truth exists and it can be challenging to decide which reference object needs more detail and we leave it as our future work.


\newpage

\begin{figure*}
    \centering
    \setlength{\tabcolsep}{2pt}
    \scalebox{0.95}{
    \begin{tabular}{cc|ccc}
    \toprule

         Input & Reference Subject & \multicolumn{3}{c}{Text-Guided Subject-Driven Inpainting} \\ \midrule

           & & \textit{dog} & \textit{dog wearing sunglasses} & \textit{dog wearing hat} \\ 
           \includegraphics[scale=0.19]{figs/supp_text/input2.jpg}&
          \includegraphics[scale=0.19]{figs/supp_text/ref2.jpg}&
           \includegraphics[scale=0.19]{figs/supp_text/ours2_1.jpg}&
         \includegraphics[scale=0.19]{figs/supp_text/ours2_2.jpg} &
         \includegraphics[scale=0.19]{figs/supp_text/ours2_3.jpg}
         \\ 
          &&
           \includegraphics[scale=0.19]{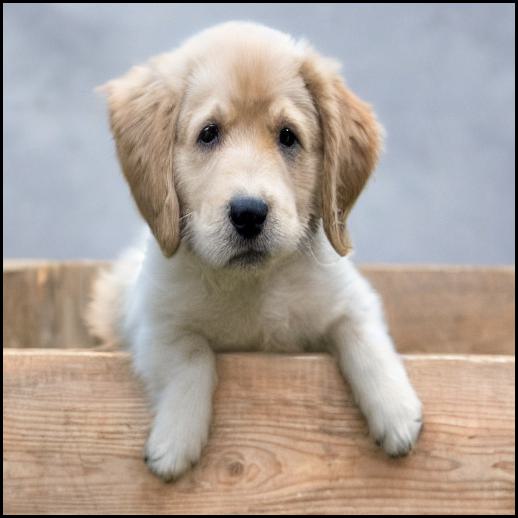}&
         \includegraphics[scale=0.19]{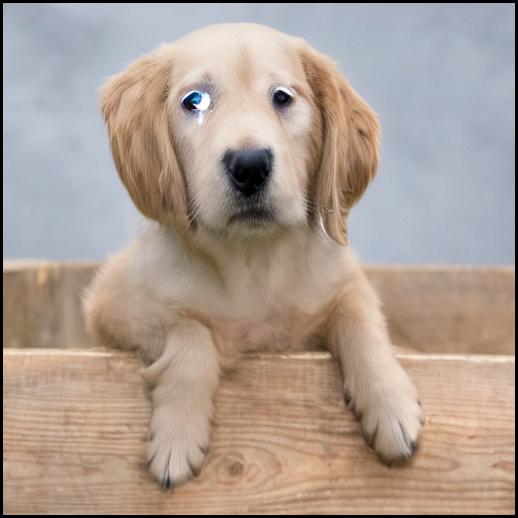} &
         \includegraphics[scale=0.19]{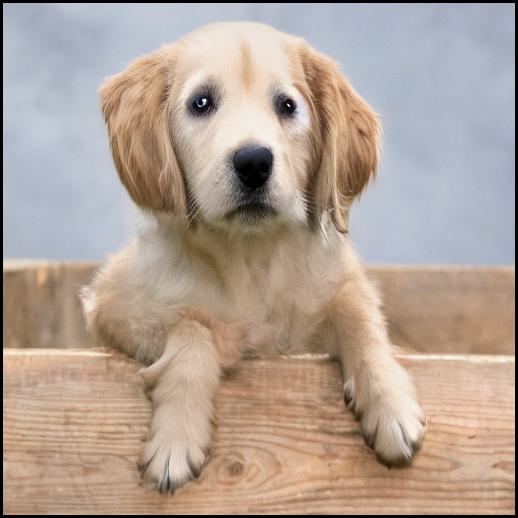}
         \\ 
           & & \textit{ballon} & \textit{heart-shaped ballon} & \textit{cube ballon} \\ 
           \includegraphics[scale=0.19]{figs/supp_text/input4.jpg}&
          \includegraphics[scale=0.19]{figs/supp_text/ref4.jpg}&
           \includegraphics[scale=0.19]{figs/supp_text/ours4_1.jpg}&
         \includegraphics[scale=0.19]{figs/supp_text/ours4_2.jpg} &
         \includegraphics[scale=0.19]{figs/supp_text/ours4_3_.jpg}
         \\ 
           &&
           \includegraphics[scale=0.19]{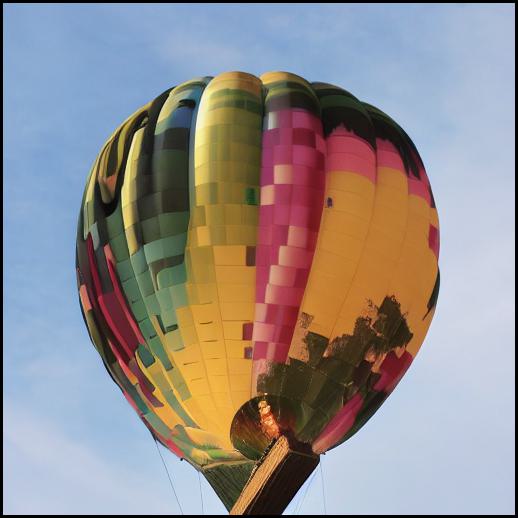}&
         \includegraphics[scale=0.19]{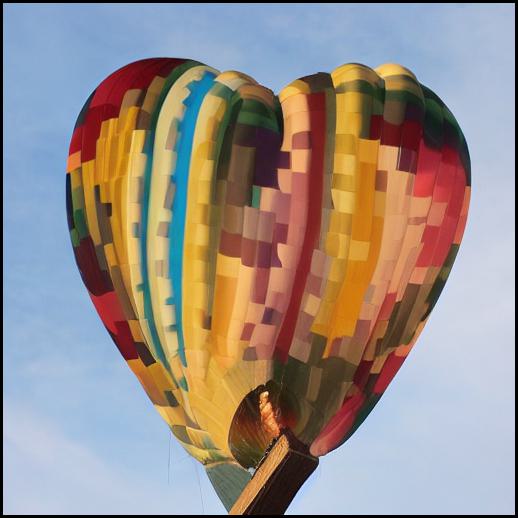} &
         \includegraphics[scale=0.19]{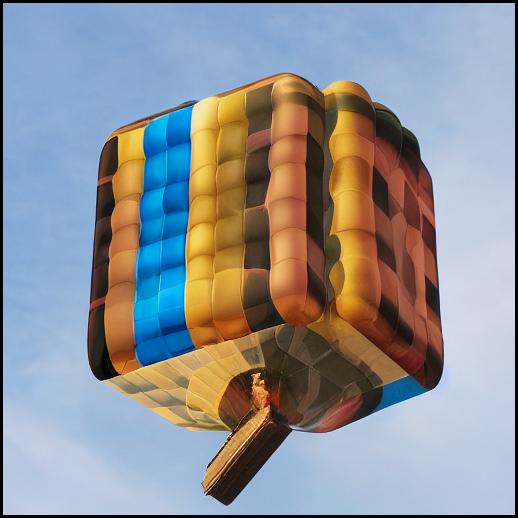}
         \\ 
            & & \textit{shoe} & \textit{boots} & \textit{vans} \\ 
           \includegraphics[scale=0.19]{figs/supp_text/input5.jpg}&
          \includegraphics[scale=0.19]{figs/supp_text/ref5.jpg}&
           \includegraphics[scale=0.19]{figs/supp_text/ours5_1.jpg}&
         \includegraphics[scale=0.19]{figs/supp_text/ours5_2.jpg} &
         \includegraphics[scale=0.19]{figs/supp_text/ours5_3.jpg}
         \\ 
          &&
           \includegraphics[scale=0.19]{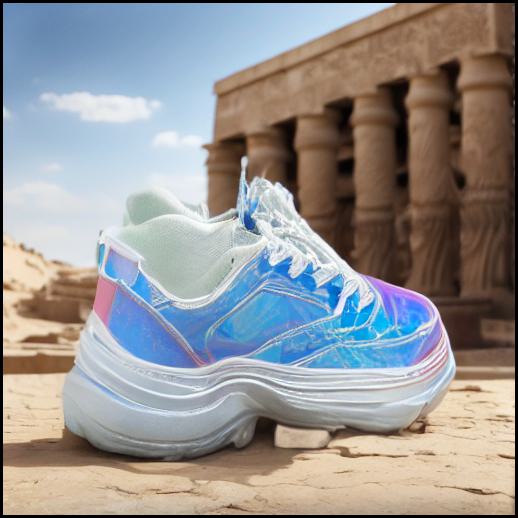}&
         \includegraphics[scale=0.19]{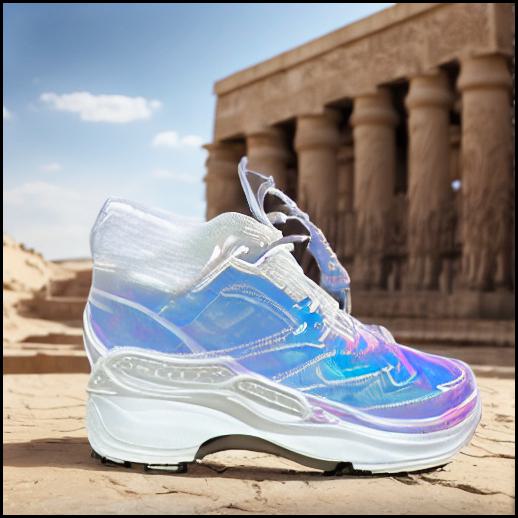} &
         \includegraphics[scale=0.19]{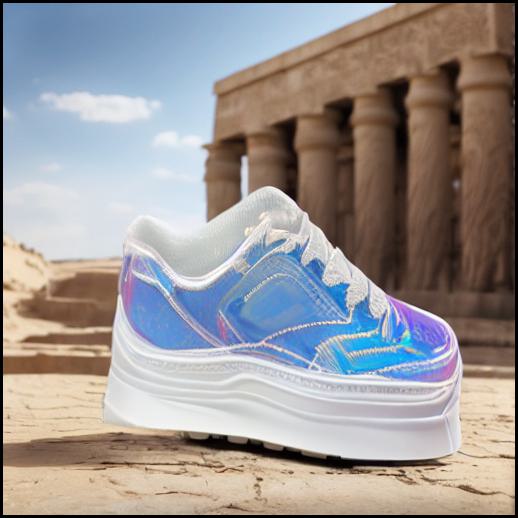}
         \\ 
         \bottomrule
    \end{tabular}
    }
    \vspace{-2mm}
    \caption{Ablation of the proposed decoupling regularization.}
    \label{fig:supp_ablation_decouple}
\end{figure*}

\end{document}